\documentclass[sigconf,nonacm,screen]{acmart}

\newcommand{\methodname}{MemTrial\xspace}
\newcommand{\methodexpansion}{memory on trial}

\usepackage{xspace}
\usepackage{amsmath}
\usepackage{booktabs}
\usepackage{multirow}
\usepackage{makecell}
\usepackage{tabularx}
\usepackage{array}
\usepackage{xcolor}
\usepackage{tikz}
\usetikzlibrary{arrows.meta,calc,positioning}
\usepackage{algorithm}
\usepackage[noend]{algpseudocode}

\DeclareMathOperator*{\argmax}{arg\,max}
\DeclareMathOperator{\corr}{corr}
\DeclareMathOperator{\diag}{diag}
\DeclareMathOperator{\mean}{mean}
\newcommand{\design}{\mathcal{D}}
\newcommand{\xref}{x^{\mathrm{ref}}}
\newcommand{\KL}{\mathrm{KL}}
\newcommand{\R}{\mathbb{R}}
\newcommand{\Normal}{\mathcal{N}}

\newcommand{\adapted}{\textsuperscript{\dag}}
\newcommand{\signif}{\textsuperscript{*}}
\newcommand{\pmsd}[2]{#1\,{\scriptsize$\pm$\,#2}}
\newcommand{\best}[1]{\textbf{#1}}                   %
\newcommand{\second}[1]{\underline{#1}}             %
\newcommand{\ours}{\textbf{\methodname{} (ours)}}   %
\definecolor{posgreen}{RGB}{0,128,55}
\definecolor{negred}{RGB}{200,30,30}
\newcommand{\improvpos}[1]{\textcolor{posgreen}{#1}}   %
\newcommand{\improvneg}[1]{\textcolor{negred}{#1}}     %

\algrenewcommand\algorithmicrequire{\textbf{Input:}}
\algrenewcommand\algorithmicensure{\textbf{Output:}}
\algrenewcommand{\algorithmiccomment}[1]{\hfill{\footnotesize\textcolor{gray}{$\triangleright$ #1}}}

\newcolumntype{L}[1]{>{\raggedright\arraybackslash}p{#1}}

\renewcommand{\arraystretch}{1.08}

\usepackage{colortbl}
\definecolor{oursrow}{RGB}{232,238,248}

\providecolor{posgreen}{RGB}{0,128,55}
\providecolor{negred}{RGB}{200,30,30}
\renewcommand{\best}[1]{\textbf{#1}}
\renewcommand{\second}[1]{\underline{#1}}
\providecommand{\improvpos}{}\renewcommand{\improvpos}[1]{\textcolor{posgreen}{#1}}
\providecommand{\improvneg}{}\renewcommand{\improvneg}[1]{\textcolor{negred}{#1}}

\usepackage{listings}
\lstdefinestyle{prompt}{basicstyle=\ttfamily\fontsize{6.4}{7.6}\selectfont, breaklines=true, columns=fullflexible,
  keepspaces=true, frame=single, rulecolor=\color{black!35}, framerule=0.3pt, framesep=2pt, xleftmargin=2.5pt,
  xrightmargin=2.5pt, aboveskip=2pt, belowskip=7pt, upquote=true, showstringspaces=false, breakindent=0pt,
  escapeinside={(*@}{@*)}}

\usepackage{dblfloatfix}
\usepackage{etoolbox}

\makeatletter
\gdef\lst@frameInit{%
    \ifx\lst@framelshape\@empty \let\lst@frameL\@empty \fi
    \ifx\lst@framershape\@empty \let\lst@frameR\@empty \fi
    \def\lst@framevrule{\vrule\@width\lst@framerulewidth\relax}%
    \lst@ifframeround
        \lst@frameCalcDimA\z@ \@getcirc\@tempdima
        \@tempdimb\@tempdima \divide\@tempdimb\tw@
        \advance\@tempdimb -\@wholewidth
        \edef\lst@frametextsep{\the\@tempdimb}%
        \edef\lst@framerulewidth{\the\@wholewidth}%
        \lst@frameCalcDimA\@ne \@getcirc\@tempdima
        \@tempdimb\@tempdima \divide\@tempdimb\tw@
        \advance\@tempdimb -\tw@\@wholewidth
        \advance\@tempdimb -\lst@frametextsep
        \edef\lst@rulesep{\the\@tempdimb}%
    \fi
    \lst@frameMakeBoxV\lst@framebox{\ht\strutbox}{\dp\strutbox}%
    \def\lst@framelr{\copy\lst@framebox}%
    \ifx\lst@frametshape\@empty\else
        \lst@frameH T\lst@frametshape
        \ifvoid\z@\else
            \par\lst@parshape
            \@tempdima-\baselineskip \advance\@tempdima\ht\z@
            \ifdim\prevdepth<\@cclvi\p@\else
                \advance\@tempdima\prevdepth
            \fi
            \ifdim\@tempdima<\z@
                \vskip\@tempdima\vskip\lineskip
            \fi
            \noindent\box\z@\par\nobreak
            \lineskiplimit\maxdimen \lineskip\z@
        \fi
        \lst@frameSpreadV\lst@framextopmargin
    \fi}
\gdef\lst@frameExit{%
    \lst@frameSpreadV\lst@framexbottommargin
    \lst@frameH B\lst@framebshape
    \ifvoid\z@\else
        \everypar{}\par\nobreak\lst@parshape\nointerlineskip\noindent\box\z@
    \fi}

\patchcmd{\lst@Init}{\par\penalty-50\relax}{\par\lst@prepenalty\relax}{}{\PackageWarningNoLine{supp}{listings patch 1 failed}}
\patchcmd{\lst@MakeCaption}{\allowbreak}{\lst@capbreak}{}{\PackageWarningNoLine{supp}{listings patch 2 failed}}
\patchcmd{\lst@HRefStepCounter}{\hyper@refstepcounter{#1}}{\hyper@refstepcounter{#1}\lst@anchorbreak}{}{\PackageWarningNoLine{supp}{listings patch 3 failed}}
\def\lst@prepenalty{\penalty-50}\def\lst@capbreak{\allowbreak}\let\lst@anchorbreak\relax
\newcommand{\promptlistingstart}{\par\nopagebreak\let\lst@prepenalty\nobreak\let\lst@capbreak\nobreak\let\lst@anchorbreak\nobreak}
\lstnewenvironment{promptlisting}{\promptlistingstart\lstset{style=prompt,nolol}}{}
\makeatother
\makeatletter
\long\def\arxiv@comma\par{, }
\long\def\arxiv@addauthor#1{\g@addto@macro\@currentauthors{\arxiv@comma#1}}
\patchcmd{\@mkauthors@iii}{\g@addto@macro\@currentauthors}{\arxiv@addauthor}{}{\PackageError{main_arxiv}{Author line patch failed}{}}
\makeatother
\newcommand{\codelink}{ Our code is available \href{https://github.com/moderncavemann/MemTrial}{here}.}

\begin{document}

\title[MemTrial: Learning When to Trust Memory]{\methodname: Learning When to Trust Memory \\in LLM Portfolio Agents}

\author{Guanghao Wu}
\authornote{Both authors contributed equally to this research.}
\author{Zhuo Cai}
\authornotemark[1]
\author{Shoujin Wang}
\authornote{Corresponding author.}
\affiliation[obeypunctuation=true]{\institution{University of Technology Sydney}\country{}}

\renewcommand{\shortauthors}{Wu, Cai and Wang}

\begin{abstract}
Large language model (LLM) agents for portfolio management learn from experience: they credit each experience in their memory with the outcome of the decisions that used it. In financial markets, however, this outcome mostly reflects the market move shared by all decisions on that date, so the credit tracks the market rather than the experience, and these agents often do worse than simply holding the equal-weight (1/$N$) portfolio. We ask how an agent can measure what an experience changes, and answer by putting memory on trial: drafts of the same decision with and without an experience face the same market, so the outcome they share cancels in their difference. Our agent, \methodname, generates eight drafts of each decision with subsets of its retrieved experiences chosen by a fractional factorial design, and estimates each experience's contribution as its Banzhaf value, the average of these differences. As each date occurs once and each draft is a noisy LLM sample, these contributions are noisy and may not hold on unseen dates. \methodname therefore pools them into values across dates and similar experiences with a hierarchical Bayesian model, acts on the values only after they have predicted unseen dates, and otherwise stays anchored at a conservative reference such as 1/$N$. On four benchmarks, it not only benefits from experiences that matter (the best of 15 methods on a semi-synthetic benchmark with known experience quality) but also, through its anchor, limits its losses when its values do not hold (at most 2.2\% below 1/$N$ on PortBench and InvestorBench, against 15--38\% for the best experience-learning agent). Averaged over five benchmark settings, its utility is 21.2\% higher than that of the best experience-learning agent, and with eight LLMs it beats every LLM-based baseline on InvestorBench.\codelink
\end{abstract}


\maketitle

\section{Introduction}\label{sec:intro}

LLM agents are increasingly used to manage investment portfolios~\cite{yu2025finmem,zhang2024multimodal,li2025investorbench}. To improve with use, they learn from experience: when the outcome of a decision is known, the agent writes an experience, such as a short lesson, into its memory, and at later dates it retrieves the most relevant experiences and conditions its decision on them~\cite{shinn2023reflexion,zhao2024expel}. Which experiences it comes to rely on is decided by \emph{outcome credit}, in one of two ways: an experience used in a decision that turned out well gains importance or value~\cite{yu2025finmem,zhang2026memrl}, or lessons are distilled from episodes judged by their outcomes to be successes or failures~\cite{shinn2023reflexion,zhao2024expel}.

\begin{figure}[!htb]
  \centering
  \includegraphics[width=0.96\columnwidth]{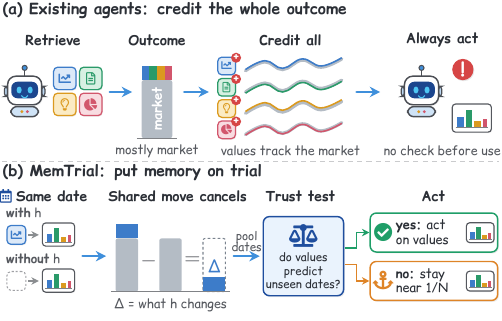}
  \caption{\small(a) Existing agents credit experiences with market-driven outcomes and always use these credits. (b) \methodname measures what each experience changes on the same date and acts on its values only after they have predicted unseen dates; otherwise it stays near 1/$N$.}
  \Description{Two rows of icons. Row (a), existing agents: a robot retrieves four experiences; the outcome is mostly the market; all experiences are credited, and the resulting values track the market; the agent always acts on them without a check, marked with a warning sign. Row (b), MemTrial: drafts with and without an experience h on the same date; the shared move cancels, and their difference is what h changes; differences are pooled across dates; a trust test asks whether the values predict unseen dates; if yes, the agent acts on the values; if no, it stays near 1/N, shown with an anchor.}
  \label{fig:overview}
\end{figure}

In portfolio management, this rule has a blind spot. The outcome of a decision is dominated by the market move over the holding period, and every decision made on that date shares this move. An experience retrieved before a rally is credited with the rally, and one retrieved before a sell-off is blamed for it, much as a treatment effect estimated without a control group mixes the treatment with everything else that happened (Figure~\ref{fig:overview}a). Table~\ref{tab:confound} measures this on three real benchmarks (\S\ref{sec:setup}), using the utility of the equal-weight (1/$N$) portfolio over the same holding period as a proxy for the market. On every date, we generate a draft with each subset of the four retrieved experiences and compare two measures for each experience: its outcome credit, the mean utility of the drafts that used it, and its counterfactual contribution, the mean utility of these drafts minus that of the drafts without it. Outcome credit rises and falls with the market almost in lockstep, but is nearly unrelated to the contribution of the experience, which itself is unrelated to the market. As each experience is in half of the drafts, its outcome credit is the date's mean utility plus half its contribution, and the first term dominates. Agents that accumulate outcome credit therefore rank experiences by the markets in which they were retrieved. Accordingly, every experience-learning agent we test underperforms 1/$N$ on PortBench and InvestorBench (Table~\ref{tab:main}).

Data valuation offers a remedy: it values a data point by its marginal contribution to a utility, averaged over subsets of the other points~\cite{ghorbani2019data,wang2023data}. Applying it to LLM portfolio agents raises two challenges (CHs).

\textbf{CH1: How can an agent measure what an experience changes when every measurement is a costly, noisy LLM call on a market date that never comes back?} Data valuation evaluates a utility many times on a fixed dataset. For an LLM agent, every evaluation is a paid call whose output varies from sample to sample, and every decision date is a single market episode. A contribution measured on one date is therefore mostly noise: on PortBench, the retrieved experiences change the agent's utility detectably on only 2--4 of 56 months, about as often as chance alone would produce (Table~\ref{tab:confound}).

\begin{table}[h]
  \caption{\small Outcome credit tracks the market, not the experience.}
  \label{tab:confound}
  \footnotesize
  \setlength{\tabcolsep}{2.5pt}
  \begin{tabular}{@{}lcccc@{}}
    \toprule
     & \multicolumn{3}{c}{Spearman correlation ($\rho$)} & \\
    \cmidrule(lr){2-4}
    Benchmark & \makecell{Outcome credit\\vs.\ 1/$N$} & \makecell{Contribution\\vs.\ 1/$N$} & \makecell{Outcome credit\\vs.\ contribution} & \makecell{Dates with a\\detectable effect} \\
    \midrule
    PortBench-Full & \pmsd{$+0.91$}{0.00} & \pmsd{$+0.04$}{0.03} & \pmsd{$+0.11$}{0.06} & 3--4 / 56 \\
    PortBench-Raw & \pmsd{$+0.92$}{0.00} & \pmsd{$+0.03$}{0.04} & \pmsd{$+0.04$}{0.03} & 2--4 / 56 \\
    InvestorBench & \pmsd{$+0.95$}{0.00} & \pmsd{$-0.07$}{0.02} & \pmsd{$-0.03$}{0.02} & 126--127 / 149 \\
    ClassAlloc & \pmsd{$+0.97$}{0.00} & \pmsd{$+0.02$}{0.07} & \pmsd{$+0.02$}{0.07} & 53--57 / 71 \\
    \bottomrule
  \end{tabular}
\end{table}

\textbf{CH2: How can an agent tell whether a learned value will hold on dates it has not seen?} Even where experiences clearly change the decision, as on about 85\% of InvestorBench days, the contributions learned from past days do not predict those of the following days (\S\ref{sec:main}), and selecting experiences by their average past contribution still trails 1/$N$ (Table~\ref{tab:main}). Existing agents always act on their learned scores: they have no notion of insufficient evidence and no fallback for when their scores do not hold.

\begin{figure*}[t]
  \centering
  \includegraphics[width=0.75\textwidth]{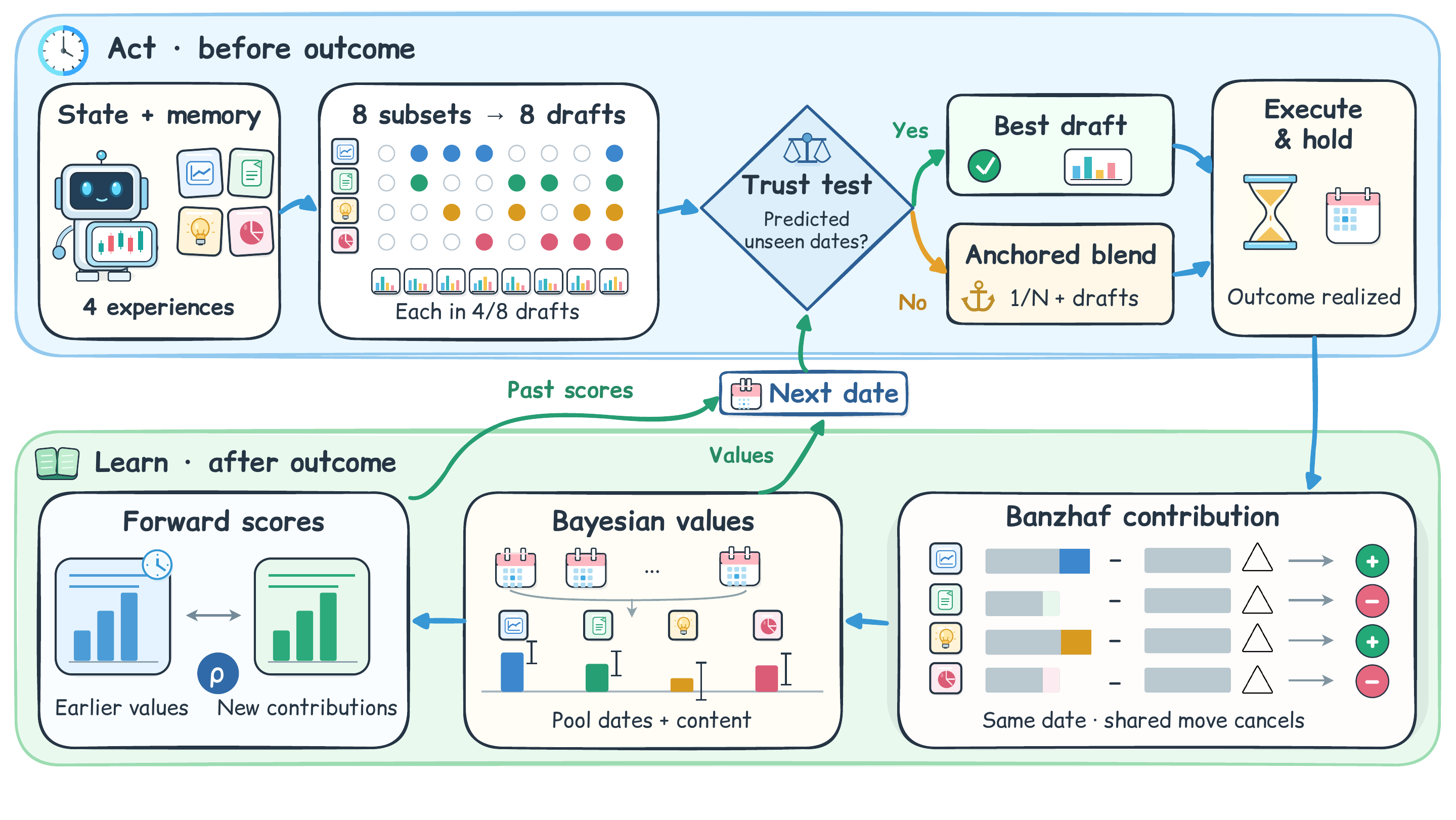}
  \caption{\small One decision date of \methodname. Before the outcome, it acts on its learned values only if they have predicted unseen dates; after the outcome, it updates the values and scores their predictions. Contributions compare drafts of one date, so their shared market move cancels.}
  \Description{Flow chart with two bands. Top band, act before the outcome: state and memory with four experiences; eight subsets give eight drafts; a trust test leads to the best draft if trusted or to an anchored blend of 1/N and the drafts if not; the deployed portfolio is held until the outcome is realized. Bottom band, learn after the outcome: Banzhaf contributions compare drafts with and without each experience on the same date, so the shared move cancels; Bayesian values per experience with error bars, pooled across dates and content; forward scores compare earlier values with the new contributions; values and past scores return to the trust test of the next date.}
  \label{fig:loop}
\end{figure*}

To address these challenges, we present \textbf{\methodname} (\methodexpansion), an experience-learning agent that learns when to trust its memory through a closed loop of four steps (Figures~\ref{fig:overview}b and~\ref{fig:loop}). (1)~\emph{Counterfactual attribution}: on every date, the agent generates eight drafts of its decision with combinations of its four retrieved experiences chosen by a fractional factorial design~\cite{box2008estadistica}, and estimates the contribution of each experience as its Banzhaf value~\cite{banzhaf1964weighted,wang2023data}, the average difference between the drafts with and without it. Because all drafts face the same market, the outcome they share cancels. (2)~\emph{Value estimation}: as in small-area estimation~\cite{fay1979estimates}, a hierarchical Bayesian model pools these noisy contributions into values across dates and across experiences with similar content. (3)~\emph{Trust test}: following prequential analysis~\cite{dawid1984present}, the agent scores its predicted values against the contributions measured once the outcome is known, and trusts its values only when these out-of-sample forward scores are significantly positive. (4)~\emph{Anchored action}: if the values are trusted, the agent deploys the draft with the highest predicted value; otherwise it deploys a blend of its drafts anchored at a conservative reference, such as 1/$N$, and learns online how far to move from it. Steps (1)--(2) address CH1, and steps (3)--(4) address CH2.

We evaluate \methodname on three real benchmarks, PortBench, InvestorBench and ClassAlloc, and on PlantedMem, a semi-synthetic environment with real prices and real LLM drafts in which the quality of every experience is known. \textbf{Our contributions are: (i)}~we show that outcome credit, the credit rule of existing experience-learning agents, tracks the market rather than what an experience changes; \textbf{(ii)}~we propose \methodname, a closed learning loop that learns what each experience is worth and when this knowledge can be trusted; \textbf{(iii)}~against 14 baselines on four benchmarks and with eight LLMs, \methodname is the best method when experiences matter and stays close to 1/$N$ when its values do not hold, where existing experience-learning agents fall behind. Appendices~\ref{app:related}--\ref{app:limitations} follow the references.

\section{Related Work}\label{sec:related}

\textbf{LLM agents for financial decision making.} LLM agents now trade single assets and allocate across many. FinMem~\cite{yu2025finmem} ranks memories by recency, relevance and an importance that trading feedback reinforces; FinCon~\cite{yu2024fincon} updates the investment beliefs of a team of agents by verbal reinforcement from realized profit and loss; FinAgent~\cite{zhang2024multimodal} and CryptoTrade~\cite{li2024cryptotrade} reflect on past trades; and TradingAgents~\cite{xiao2024tradingagents} simulates the roles of a trading firm. InvestorBench~\cite{li2025investorbench} and PortBench~\cite{zhao2026portbench} evaluate such agents on stocks, cryptocurrencies and multi-asset portfolios. These agents judge a memory, reflection or strategy by the outcome of the trades that used it, which the market move dominates, and none of them checks whether what it has learned holds on later dates.

\textbf{Experience learning and memory in LLM agents.} Outside finance, Reflexion~\cite{shinn2023reflexion} turns feedback on failed attempts into verbal reflections, ExpeL~\cite{zhao2024expel} distils insights from successful and failed trajectories, Voyager~\cite{wang2023voyager} keeps a library of skills that it reuses when they succeed, and Generative Agents~\cite{park2023generative}, MemoryBank~\cite{zhong2024memorybank} and MemGPT~\cite{packer2023memgpt} decide which memories to retrieve, keep or forget by recency, importance and relevance. MemRL~\cite{zhang2026memrl} learns a value for each memory from the rewards of the episodes that used it, and retrieval-augmented generation~\cite{lewis2020retrieval} retrieves by similarity alone. These methods credit outcomes to the memories involved or distil lessons from successes and failures, and never test whether what they learn holds later. This works when outcomes reflect the agent's own choices, as in games, coding or multi-agent environments, but not when a shock shared by all decisions dominates them.

\textbf{Counterfactual credit and data valuation.} Data Shapley~\cite{ghorbani2019data,jia2019towards}, Beta Shapley~\cite{kwon2021beta} and Data Banzhaf~\cite{wang2023data} value a training point by its averaged marginal contributions to a model's performance, and Data Banzhaf is the most robust of them when the utility is noisy. Influence functions~\cite{koh2017understanding} approximate the effect of removing a point. Uplift modelling~\cite{gutierrez2017causal} and counterfactual learning from logged decisions~\cite{bottou2013counterfactual,swaminathan2015counterfactual} estimate the effect of an action from observational data. These methods value data for a fixed model and test set, or estimate effects from logs in which, in our setting, the market and the choice of memory are confounded. None of them measures what a memory changes in an LLM's decisions, or asks whether values learned on past dates predict future ones.

\textbf{Safe deployment and portfolio choice under estimation error.} Safe policy improvement deploys a learned policy only when it beats a baseline with high confidence~\cite{thomas2015high,laroche2019safe}, and prequential analysis~\cite{dawid1984present} judges a model by its sequential out-of-sample predictions. In portfolio choice, estimation error makes optimized portfolios hard to trust: none of those evaluated by \citet{demiguel2009optimal} consistently beats 1/$N$ out of sample, which motivates shrinkage towards conservative targets~\cite{jorion1986bayes,ledoit2003honey,tu2011markowitz}. \methodname brings these ideas to the memory of LLM agents: a trust test on forward scores decides whether its learned values are trusted, and until then an anchored blend learns online how far to move from a conservative reference.

What is new in \methodname is not the Banzhaf value~\cite{banzhaf1964weighted} but how it is measured and used: on LLM decisions through a designed experiment on each date, pooled into a value model, and trusted only after this model has predicted unseen dates. Please refer to Appendix~\ref{app:related} for more related work.

\section{\methodname: Learning When to Trust Memory}\label{sec:method}

\subsection{Problem Formulation and Overview}\label{sec:problem}

\textbf{Portfolio decisions under an investor mandate.} An agent manages a portfolio of $N$ assets and cash for an investor. At each decision date $t=1,\dots,T$, it observes the market state $s_t$ and outputs portfolio weights $x_t$, which are held over a holding period. A projection $P_\theta$ enforces the investor's mandate, a cap $M$ on the share of risky assets and a floor $m$ on the share of defensive assets; the investor profile $\theta=(M,m,\gamma)$ also contains the risk aversion $\gamma$. A decision is scored by the investor utility of the executed portfolio, a certainty-equivalent (CEQ) return~\cite{demiguel2009optimal},
\begin{equation}\label{eq:utility}
  u_t(x) = r_t\big(P_\theta(x)\big) - \frac{\gamma}{2}\, V_t\big(P_\theta(x)\big),
\end{equation}
where $r_t$ is the realized return over the holding period net of proportional trading costs and $V_t$ is the variance of daily returns times the number of days in that period (for one day, the squared return).

\textbf{An LLM agent with an experience memory.} The agent is an LLM policy $\pi$ that maps the state and a set of experiences $S$ to weights, $x\sim\pi(\cdot\mid s_t,S)$. Its memory holds experiences written once the outcomes of earlier decisions were known: lessons, or records of earlier decisions and their outcomes. At date $t$, the retriever returns the $k=4$ experiences whose embeddings are most similar to that of the current state, $R_t=(h_{t,1},\dots,h_{t,k})$. Because the LLM samples at a positive temperature, the same prompt yields different weights; we call each sample a \emph{draft}. Experience learning selects which retrieved experiences enter the prompt, $S_t\subseteq R_t$, based on scores learned from earlier dates. Existing agents credit an experience used at date $t$ with $u_t(x_t)$, the utility of the whole decision~\cite{yu2025finmem,zhang2026memrl}, which Table~\ref{tab:confound} shows mostly measures the market.

\textbf{Our goal and the \methodname loop.} We aim to maximize the investor's cumulative utility $\sum_t u_t(x_t)$ by learning what each experience is worth, from what it changes in decisions, and when this knowledge can be trusted. Figure~\ref{fig:loop} shows the loop (Algorithm~\ref{alg:loop} in Appendix~\ref{app:algorithms} gives its pseudocode). Before the outcome of date $t$ is known, \methodname generates drafts with designed subsets of $R_t$ and acts (\S\ref{sec:anchor}) according to the trust test (\S\ref{sec:trust}). After the outcome is known, it attributes the change in utility to each experience (\S\ref{sec:attribution}), updates the values (\S\ref{sec:value}) and scores their predictions (\S\ref{sec:trust}).

\subsection{Counterfactual Attribution: What Does an Experience Change?}\label{sec:attribution}

\textbf{Designed drafts.} To know what an experience changes, \methodname compares decisions made on the same date with and without it, as in a controlled experiment: both face the same market path, so the outcome they share cancels in their difference. Comparing every subset of $k$ experiences would need $2^k$ drafts, 16 for $k=4$. A fractional factorial design~\cite{box2008estadistica} measures the effect of each experience with half as many drafts: the half fraction
\begin{equation}\label{eq:design}
  \design = \{\, S\subseteq R_t : |S| \text{ is even} \,\},
\end{equation}
a $2^{4-1}$ design of resolution IV. Its eight subsets include the empty and the full set, every experience appears in half of them and every pair in two, so the effect of one experience is not aliased (mixed) with any interaction between two others (Appendix~\ref{app:design}).

\textbf{Contribution as an estimated Banzhaf value.} Let $U_t(S)=u_t(x_t^S)$ be the utility that the draft made with subset $S$ would have earned at realized prices (the agent is a price taker). The counterfactual contribution of the $j$-th retrieved experience is
\begin{equation}\label{eq:contribution}
  \begin{aligned}
    c_{t,j} = {}&\mean\{U_t(S) : S\in\design,\ j\in S\}\\
    &- \mean\{U_t(S) : S\in\design,\ j\notin S\},
  \end{aligned}
\end{equation}
the main effect of experience $j$ in the $2^{4-1}$ design. Over all 16 subsets, the main effect is the Banzhaf value~\cite{banzhaf1964weighted} of $j$ in the game $U_t$; in the half fraction, it is aliased with the interaction of the other three experiences, which adds 13--17\% to its noise variance on the benchmarks where all 16 subsets were drafted (Appendix~\ref{app:design}). The Banzhaf value weights every context of the other experiences equally, which makes it the most robust semivalue under noisy utilities~\cite{wang2023data}, as for sampled LLM outputs. Any utility shared by all drafts of date $t$ cancels in $c_{t,j}$, and the market move enters it only through the weights that experience $j$ changes. Since $U_t$ is evaluated after the mandate projection, a change that the mandate removes contributes nothing. On the design, the outcome credit of $j$, the mean utility of the drafts that used it, equals $\bar U_t + c_{t,j}/2$, where $\bar U_t$, the mean utility of the eight drafts, follows the market (Table~\ref{tab:confound}). Finally, we centre the contributions within each date,
\begin{equation}\label{eq:centre}
  y_{t,j} = c_{t,j} - \frac{1}{k}\sum_{j'=1}^{k} c_{t,j'},
\end{equation}
which removes what all retrieved experiences do in common on that date and keeps what distinguishes them.

\subsection{Value Estimation: Pooling Noisy Evidence}\label{sec:value}

\textbf{Hierarchical value model.} A contribution measured on a single date is mostly noise, and a new experience has no measurements at all. Small-area estimation faces the same problem when it estimates the income of a small town from a few survey responses, and solves it by borrowing strength from similar towns~\cite{fay1979estimates}. Let $n_h$ be the number of dates with a realized outcome on which experience $h$ was retrieved and $m_h$ the mean of its centred contributions. The value model is
\begin{equation}\label{eq:fh}
  \begin{aligned}
    m_h &= v_h + \varepsilon_h, & \varepsilon_h &\sim \Normal(0,\, \sigma^2/n_h),\\
    v_h &= z_h^{\top}\beta + b_h, & \beta &\sim \Normal(0,\, \sigma_\beta^2 I),\quad b_h \sim \Normal(0,\, \tau^2),
  \end{aligned}
\end{equation}
where $v_h$ is the value of experience $h$, $z_h\in\R^d$ a content feature (the first $d=8$ standardized principal components of its retrieval embedding), $\beta$ maps content to value, and $b_h$ is the experience's own deviation. This is the Fay--Herriot model with content as the covariate, or a Gaussian process with a kernel linear in content plus an identity term~\cite{seeger2004gaussian}. The noise variance $\sigma^2$ is estimated from the spread of each experience's contributions, and $(\sigma_\beta^2,\tau^2)$ maximize the marginal likelihood
\begin{equation}\label{eq:marglik}
  \begin{gathered}
    \log p(m\mid\sigma_\beta^2,\tau^2) = \log\Normal(m;\,0,\,K),\\
    K = \sigma_\beta^2 ZZ^{\top} + \tau^2 I + \sigma^2\diag(1/n_h),
  \end{gathered}
\end{equation}
over a grid. The posterior means are
\begin{equation}\label{eq:posterior}
  \hat v = (\sigma_\beta^2 ZZ^{\top} + \tau^2 I)\,K^{-1}m,\qquad \hat\beta = \sigma_\beta^2 Z^{\top}K^{-1}m,
\end{equation}
so an experience with few measurements is shrunk towards the value predicted from its content, and a new one receives $\hat v_h=z_h^{\top}\hat\beta$ before it is ever used.

\textbf{Learner selection.} Content may or may not predict value, so \methodname keeps two learners, the content-aware model~\eqref{eq:fh} and a per-experience model with $\sigma_\beta^2=0$. It scores both forward (\S\ref{sec:trust}) and uses the one with the higher mean forward score once both have at least 10 scores (before that, the per-experience model).

\subsection{Trust Test: Has the Learned Value Predicted the Future?}\label{sec:trust}

\textbf{Forward scores.} A weather forecaster is judged by how well she predicted tomorrow's weather, not by how well her model fits yesterday's, and prequential analysis judges a statistical model in the same way~\cite{dawid1984present}. Before date $t$, each learner is fitted only on dates whose outcomes are known and predicts the values $\hat v_{t,j}$ of the experiences retrieved at $t$. Once the outcome is known, the prediction is scored against the measured centred contributions,
\begin{equation}\label{eq:forward}
  \rho_t = \corr\big((\hat v_{t,1},\dots,\hat v_{t,k}),\ (y_{t,1},\dots,y_{t,k})\big),
\end{equation}
the Pearson correlation across the retrieved experiences, which is out of sample by construction.

\textbf{Trust test.} Let $\rho_1,\dots,\rho_n$ be the forward scores of the selected learner available at date $t$, with mean $\bar\rho$ and standard deviation $s_\rho$. As in safe policy improvement~\cite{thomas2015high}, \methodname trusts its values only on strong evidence: when $n\ge 10$ and
\begin{equation}\label{eq:ttest}
  p_t = \Pr\!\left(T_{n-1} \ge \frac{\bar\rho}{s_\rho/\sqrt{n}}\right) < \alpha,
\end{equation}
where $T_{n-1}$ follows a Student $t$ distribution with $n-1$ degrees of freedom and $\alpha=0.05$. Trust thus means that the values have ranked the retrieved experiences better than chance on unseen dates; used as a decision rule on every date, the test rarely passes where experiences carry no signal (\S\ref{sec:analysis}). When trusted, \methodname deploys the designed draft with the highest total predicted value,
\begin{equation}\label{eq:trusted}
  S_t^{*} = \argmax_{S\in\design} \sum_{j\in S} \hat v_{t,j},
\end{equation}
with ties broken towards fewer experiences, provided that this total is positive, i.e., that the subset's experiences are predicted to beat the average retrieved one; \S\ref{sec:analysis} measures whether trusted drafts beat the anchored blend. This draft was already generated for Eq.~\eqref{eq:contribution}, so trusting costs no extra LLM call. The test asks whether the values have predicted the future, not whether experiences differed in the past (\S\ref{sec:ablation} compares the two).

\subsection{Anchored Action: What to Do Until Memory Can Be Trusted}\label{sec:anchor}

\textbf{Anchored blend.} The 1/$N$ portfolio is hard to beat out of sample~\cite{demiguel2009optimal}, and combining it with a more ambitious strategy, weighted by how strongly the data support the latter, can improve on both~\cite{tu2011markowitz}. \methodname applies this idea online. When its values are not trusted, it deploys a blend of a conservative reference $\xref$ (on the real benchmarks, the 1/$N$ portfolio projected to the mandate) and the average draft $\bar x_t=\frac{1}{|\design|}\sum_{S\in\design}x_t^S$,
\begin{equation}\label{eq:ftrl}
  \begin{gathered}
    x_t(q) = q\,\xref_t + (1-q)\,\bar x_t,\\
    q_t = \argmax_{q\in[0,1]}\ \frac{1}{n}\sum_{s<t} u_s\big(x_s(q)\big) - \lambda_n\,\KL(q\,\|\,\alpha_0),\qquad \lambda_n = \frac{\lambda_0\,\hat\sigma_n}{\sqrt{n}},
  \end{gathered}
\end{equation}
where $\KL(q\,\|\,\alpha_0)$ is the Kullback--Leibler divergence between Bernoulli distributions, $\alpha_0=0.9$ is the prior weight on the reference, $\lambda_0=4$, $n$ is the number of dates with a realized outcome, and $\hat\sigma_n$, half the mean absolute utility gap between the reference and the average draft, makes the regularizer unit-free. This is follow-the-regularized-leader~\cite{shalev2025online} with a regularizer centred at the prior: with little evidence, $q_t$ stays close to $\alpha_0$; as evidence accumulates, the regularization weakens as $1/\sqrt{n}$ and $q_t$ follows the blend that performed best on these dates.

\subsection{The Closed Loop and Its Cost}\label{sec:loop}

The same eight drafts serve every step: they yield the contributions~\eqref{eq:contribution}, the trusted action~\eqref{eq:trusted} and the average draft in~\eqref{eq:ftrl}. \methodname therefore makes $|\design|=8$ drafts per decision, which run in parallel; its LLM cost is eight times that of a single-draft agent. Fitting the value model costs $O(G\cdot H^3)$ for $G$ grid points and $H$ experiences, a few milliseconds per decision (Appendix~\ref{app:impl}).

\section{Experiments}\label{sec:experiments}

\subsection{Experimental Setup}\label{sec:setup}

\begin{table}[h]
  \caption{\small Three real benchmarks and one semi-synthetic benchmark with planted experiences of known quality.}
  \label{tab:bench}
  \centering
  \scriptsize
  \setlength{\tabcolsep}{4pt}
  \renewcommand{\arraystretch}{1.1}
  \begin{tabular}{lcccc}
    \toprule
    & PortBench & InvestorBench & ClassAlloc & PlantedMem \\
    \midrule
    Assets      & 99 (6 classes) & 4 stocks, cash & 5 classes & 4 classes \\
    Decisions   & monthly        & daily          & monthly   & monthly \\
    Development & 2019--22       & none           & none      & 15 worlds \\
    Test dates  & 20 (2023--24)  & 149 (10/20--05/21) & 71 (05/20--03/26) & 120 / world \\
    Experiences & 338 decisions  & 63 lessons     & 58 lessons & 12/48 planted \\
    Seeds       & 3              & 3              & 3         & 100 worlds \\
    \bottomrule
  \end{tabular}
\end{table}

\textbf{Benchmarks.} Table~\ref{tab:bench} summarizes the four benchmarks, and Appendix~\ref{app:impl} gives their construction. On \emph{PortBench}~\cite{zhao2026portbench}, each decision starts from the equal-weight portfolio, holds for 20 trading days and pays 15~bp on the amount traded. PortBench's native prompt shows prices for only eight of its 99 assets: PortBench-Raw keeps this prompt, and PortBench-Full shows every eligible asset; the prompts of the other benchmarks show every asset. On \emph{InvestorBench}~\cite{li2025investorbench}, the agent also reads daily news and writes its lessons during a warm-up from July to September 2020. \emph{ClassAlloc} tests agents after the LLM's knowledge cutoff: built from PortBench's prices, it ends 21 months after that of gpt-4.1-mini~\cite{openai2025gpt41} (June 2024), carries holdings from month to month, and gives every learner an outcome only after its 20 trading days. Experiences are retrieved by embedding similarity, with Qwen3-Embedding-0.6B~\cite{zhang2025qwen3} on PortBench and text-embedding-3-small~\cite{openai2024embedding} on InvestorBench and ClassAlloc. \emph{PlantedMem} keeps real prices and real LLM drafts but plants twelve experiences of known quality: three useful, six neutral and three misleading. A useful (misleading) experience moves a share $\beta$ of the draft to an asset class, with probability 0.7 the one that does best (worst) next (Appendix~\ref{app:impl}). Retrieval depends on the market regime, which reproduces the market confound. Five core regimes vary how strongly experiences move a draft (no influence, noise only, and $\beta\in\{0.1,0.25,0.5\}$), and two more have 48 experiences whose content is uninformative (\emph{many experiences}) or informative (\emph{informative content}) about their quality.

\textbf{Investors and metric.} We use PortBench's conservative ($M=0.40$, $m=0.40$, $\gamma=10$), balanced ($M=0.65$, $m=0.20$, $\gamma=5$) and aggressive ($M=0.90$, $m=0.05$, $\gamma=2$) investors. We report the utility~\eqref{eq:utility} per decision period, averaged over dates and investors, in \% per month (bp per day on InvestorBench); higher is better.

\textbf{Baselines} (Appendix~\ref{app:baselines}). (i)~Rule-based: 1/$N$ projected to the mandate~\cite{demiguel2009optimal} and Minimum variance~\cite{markowitz1955portfolio}. (ii)~LLM agents without experience learning: Memory-free (zero-shot)~\cite{brown2020language}, Self-consistency~\cite{wang2022self} (the average of three memory-free drafts) and Similarity retrieval~\cite{lewis2020retrieval} of the two or four most similar experiences. (iii)~Experience-learning agents: FinMem~\cite{yu2025finmem}, MemRL~\cite{zhang2026memrl}, Reflexion~\cite{shinn2023reflexion} and ExpeL~\cite{zhao2024expel}. (iv)~Credit-based and ensemble learners: Uplift credit~\cite{gutierrez2017causal}, which regresses outcomes on experience usage with a market control; Counterfactual selection, which selects experiences by our contributions~\eqref{eq:contribution} without value estimation or trust test; Draft averaging~\cite{bates1969combination}, which deploys the average of \methodname's eight drafts; and Hedge~\cite{freund1997decision} over 1/$N$ and the memory configurations. Where a baseline's own module cannot run, we re-implement its memory-scoring rule (marked \dag); on PortBench, ExpeL\dag{} votes over the experiences, and the FinMem and MemRL modules receive no feedback after July 2019 (Appendix~\ref{app:baselines}).

\textbf{Implementation and statistics.} All LLM agents use gpt-4.1-mini (2025-04-14) at temperature 0.7, which gives the distinct drafts that \methodname, Self-consistency and Draft averaging need. \methodname uses $k=4$, the design~\eqref{eq:design}, $d=8$, $\alpha=0.05$ with at least 10 forward scores, $\alpha_0=0.9$ and $\lambda_0=4$. Its reference is 1/$N$ on the real benchmarks and the average of eight memory-free drafts in PlantedMem, where 1/$N$ ignores the risk aversion. It was frozen before any InvestorBench decision, so InvestorBench, ClassAlloc and the runs with other LLMs are held-out tests. On the real benchmarks, each of three seeds is an independent deployment in which every draft is sampled anew; PlantedMem uses 100 test seeds, each a new world. Tables report the mean $\pm$ sd over seeds, with the best result in \textbf{bold} and the second best \underline{underlined}. \emph{Improv.} is the gain of \methodname over the best experience-learning agent of each setting, in percent of that agent's utility, and * marks a significant improvement ($p<0.05$, one-sided paired $t$-test over test dates, or over seeds in PlantedMem). The \emph{trust rate} is the share of decisions on which \methodname acts on its learned values. Appendix~\ref{app:impl} gives further implementation details and Appendix~\ref{app:prompts} the prompts.

\begin{table*}[t]
  \caption{\small Main results: utility per decision period (higher is better). \methodname is the best LLM-based method on PortBench and InvestorBench, where it stays within 2.2\% of 1/$N$, and the best method in PlantedMem.}
  \label{tab:main}
  \centering
  \footnotesize
  \renewcommand{\adapted}{\rlap{\textsuperscript{\dag}}}%
  \renewcommand{\signif}{\rlap{\textsuperscript{*}}}%
  \setlength{\tabcolsep}{6.2pt}
  \renewcommand{\arraystretch}{1.0}
  \begin{tabular}{c l !{\color{black!30}\vrule} c c c c c}
    \toprule
    & \textbf{Method} & \textbf{PortBench-Full} & \textbf{PortBench-Raw} & \textbf{InvestorBench} & \textbf{ClassAlloc} & \textbf{PlantedMem} \\[-0.5ex]
    & & {\scriptsize\color{black!55}\%/month} & {\scriptsize\color{black!55}\%/month} & {\scriptsize\color{black!55}bp/day} & {\scriptsize\color{black!55}\%/month} & {\scriptsize\color{black!55}\%/month} \\
    \midrule
    \multirow{2}{*}{Rule-based} & 1/$N$~\cite{demiguel2009optimal} & \best{\pmsd{2.02}{0.00}} & \best{\pmsd{2.02}{0.00}} & \best{\pmsd{10.45}{0.00}} & \pmsd{0.623}{0.000} & \pmsd{0.57}{0.00} \\
    & Minimum variance~\cite{markowitz1955portfolio} & \pmsd{0.25}{0.00} & \pmsd{0.25}{0.00} & \pmsd{5.53}{0.00} & \pmsd{0.230}{0.000} & \pmsd{$-$0.04}{0.00} \\
    \midrule
    \multirow{4}{*}{\makecell{LLM w/o\\experience\\learning}}
    & Memory-free~\cite{brown2020language} & \pmsd{1.60}{0.05} & \pmsd{1.07}{0.02} & \pmsd{8.43}{0.29} & \second{\pmsd{0.657}{0.001}} & \pmsd{0.76}{0.08} \\
    & Self-consistency~\cite{wang2022self} & \pmsd{1.56}{0.11} & \pmsd{1.17}{0.03} & \pmsd{7.89}{0.21} & \best{\pmsd{0.686}{0.007}} & \pmsd{0.79}{0.03} \\
    & Similarity retrieval (top-2)~\cite{lewis2020retrieval} & \pmsd{1.43}{0.19} & \pmsd{1.24}{0.29} & \pmsd{8.67}{0.19} & \pmsd{0.649}{0.033} & \pmsd{0.72}{0.07} \\
    & Similarity retrieval (top-4)~\cite{lewis2020retrieval} & \pmsd{1.70}{0.06} & \pmsd{1.05}{0.02} & \pmsd{8.12}{0.61} & \pmsd{0.643}{0.039} & \pmsd{0.72}{0.06} \\
    \midrule
    \multirow{4}{*}{\makecell{Experience-\\learning\\agents}}
    & FinMem~\cite{yu2025finmem} & \pmsd{1.63}{0.31} & \pmsd{1.26}{0.14} & \pmsd{8.47}{0.17}\adapted & \pmsd{0.594}{0.017}\adapted & \pmsd{0.75}{0.06}\adapted \\
    & MemRL~\cite{zhang2026memrl} & \pmsd{1.58}{0.29} & \pmsd{1.05}{0.06} & \pmsd{8.90}{0.39}\adapted & \pmsd{0.633}{0.012}\adapted & \pmsd{0.74}{0.08}\adapted \\
    & Reflexion~\cite{shinn2023reflexion} & \pmsd{1.40}{0.18} & \pmsd{1.26}{0.03} & \pmsd{7.29}{0.66} & \pmsd{0.590}{0.061} & \pmsd{0.78}{0.06}\adapted \\
    & ExpeL~\cite{zhao2024expel} & \pmsd{1.53}{0.05}\adapted & \pmsd{1.10}{0.06}\adapted & \pmsd{8.93}{0.32} & \pmsd{0.609}{0.006} & \pmsd{0.77}{0.06}\adapted \\
    \midrule
    \multirow{4}{*}{\makecell{Credit-based\\and ensemble\\learners}}
    & Uplift credit~\cite{gutierrez2017causal} & \pmsd{1.38}{0.09}\adapted & \pmsd{1.20}{0.10}\adapted & \pmsd{8.44}{0.25}\adapted & \pmsd{0.645}{0.022}\adapted & \pmsd{0.78}{0.09}\adapted \\
    & Counterfactual selection~\cite{wang2023data} & \pmsd{1.49}{0.29} & \pmsd{1.30}{0.34} & \pmsd{8.54}{0.26} & \pmsd{0.650}{0.033} & \second{\pmsd{0.85}{0.08}} \\
    & Draft averaging~\cite{bates1969combination} & \pmsd{1.62}{0.05} & \pmsd{1.19}{0.02} & \pmsd{8.69}{0.09} & \pmsd{0.651}{0.008} & \pmsd{0.75}{0.03} \\
    & Hedge~\cite{freund1997decision} & \pmsd{1.73}{0.09} & \pmsd{1.59}{0.04} & \pmsd{8.97}{0.23} & \pmsd{0.634}{0.008} & \pmsd{0.75}{0.03}\adapted \\
    \midrule
    \rowcolor{oursrow}
    & \ours & \second{\pmsd{1.98}{0.00}} & \second{\pmsd{2.01}{0.00}} & \second{\pmsd{10.40}{0.04}} & \pmsd{0.613}{0.017} & \best{\pmsd{0.87}{0.04}}\signif \\
    & \textit{Improv.} (\%) & \improvpos{\pmsd{$+21.3$}{19.3}} & \improvpos{\pmsd{$+59.0$}{2.6}} & \improvpos{\pmsd{$+16.4$}{3.2}} & \improvneg{\pmsd{$-3.2$}{1.4}} & \improvpos{\pmsd{$+12.6$}{10.1}} \\
    \cmidrule(l){2-7}
    & Trust rate of \methodname (\%) & \pmsd{0.0}{0.0} & \pmsd{0.6}{1.0} & \pmsd{0.3}{0.5} & \pmsd{14.6}{21.7} & \pmsd{33.7}{9.2} \\
    \bottomrule
  \end{tabular}
\end{table*}

\begin{figure*}[t]
  \centering
  \includegraphics[width=0.98\textwidth]{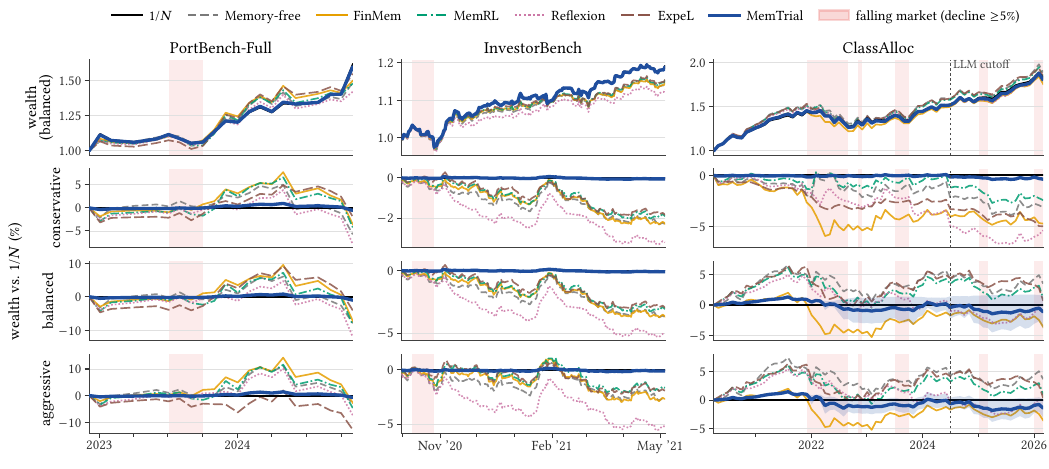}
  \caption{\small Wealth of the balanced investor (top, start $=1$) and relative to 1/$N$ for the conservative, balanced and aggressive investors (rows 2--4). For every investor, \methodname stays close to 1/$N$, while the other LLM-based methods drift away, mostly below.}
  \Description{Twelve line charts in four rows for PortBench-Full, InvestorBench and ClassAlloc. Top row: wealth of the balanced investor over the test period for 1/N, the memory-free LLM, FinMem, MemRL, Reflexion, ExpeL and MemTrial; rows two to four: wealth relative to 1/N in percent for the conservative, balanced and aggressive investors. Falling markets are shaded red, and the ClassAlloc panels mark the LLM's knowledge cutoff with a dotted line. MemTrial stays near 1/N for every investor, while the other agents drift away from it, mostly below.}
  \label{fig:wealth}
\end{figure*}

\begin{table*}[t]
  \caption{\small Return and risk of the balanced investor: cumulative net return (CR), annualized Sharpe ratio (SR) and maximum drawdown (MDD). On PortBench and InvestorBench, \methodname is the only LLM-based method close to 1/$N$ in both return and Sharpe ratio.}
  \label{tab:risk-main}
  \centering
  \footnotesize
  \setlength{\tabcolsep}{4.2pt}
  \renewcommand{\arraystretch}{1.0}
  \begin{tabular}{c l !{\color{black!30}\vrule} ccc !{\color{black!30}\vrule} ccc !{\color{black!30}\vrule} ccc}
    \toprule
    & & \multicolumn{3}{c}{\textbf{PortBench-Full}} & \multicolumn{3}{c}{\textbf{InvestorBench}} & \multicolumn{3}{c}{\textbf{ClassAlloc}} \\
    & \textbf{Method} & CR\,\%\,$\uparrow$ & SR\,$\uparrow$ & MDD\,\%\,$\downarrow$ & CR\,\%\,$\uparrow$ & SR\,$\uparrow$ & MDD\,\%\,$\downarrow$ & CR\,\%\,$\uparrow$ & SR\,$\uparrow$ & MDD\,\%\,$\downarrow$ \\
    \midrule
    \multirow{2}{*}{Rule-based} & 1/$N$~\cite{demiguel2009optimal} & \best{\pmsd{61.2}{0.0}} & \best{\pmsd{1.77}{0.00}} & \pmsd{5.6}{0.0} & \best{\pmsd{19.1}{0.0}} & \best{\pmsd{2.87}{0.00}} & \best{\pmsd{5.7}{0.0}} & \pmsd{82.0}{0.0} & \second{\pmsd{1.22}{0.00}} & \second{\pmsd{12.0}{0.0}} \\
     & Minimum variance~\cite{markowitz1955portfolio} & \pmsd{5.4}{0.0} & \pmsd{1.19}{0.00} & \best{\pmsd{1.5}{0.0}} & \pmsd{9.9}{0.0} & \pmsd{1.47}{0.00} & \pmsd{6.4}{0.0} & \pmsd{19.1}{0.0} & \best{\pmsd{4.19}{0.00}} & \best{\pmsd{0.2}{0.0}} \\
    \midrule
    \makecell{LLM w/o\\exp.\ learning} & Memory-free~\cite{brown2020language} & \pmsd{48.5}{0.8} & \pmsd{1.47}{0.06} & \pmsd{7.3}{1.0} & \pmsd{14.6}{0.5} & \pmsd{2.24}{0.08} & \pmsd{6.3}{0.1} & \second{\pmsd{88.3}{1.5}} & \pmsd{1.11}{0.00} & \pmsd{13.9}{0.3} \\
    \midrule
    \multirow{4}{*}{\makecell{Experience-\\learning\\agents}} & FinMem~\cite{yu2025finmem} & \pmsd{49.6}{9.2} & \pmsd{1.53}{0.21} & \pmsd{6.2}{1.4} & \pmsd{14.7}{0.4} & \pmsd{2.27}{0.03} & \pmsd{6.1}{0.1} & \pmsd{75.4}{2.6} & \pmsd{1.07}{0.03} & \pmsd{15.4}{0.5} \\
     & MemRL~\cite{zhang2026memrl} & \pmsd{47.7}{11.1} & \pmsd{1.45}{0.19} & \pmsd{6.9}{1.1} & \pmsd{15.4}{0.7} & \pmsd{2.33}{0.09} & \pmsd{6.1}{0.3} & \pmsd{84.0}{1.2} & \pmsd{1.06}{0.03} & \pmsd{15.4}{1.1} \\
     & Reflexion~\cite{shinn2023reflexion} & \pmsd{41.8}{6.1} & \pmsd{1.41}{0.14} & \pmsd{7.2}{1.4} & \pmsd{12.9}{1.0} & \pmsd{1.97}{0.14} & \pmsd{6.3}{0.1} & \pmsd{80.6}{8.3} & \pmsd{1.10}{0.08} & \pmsd{15.1}{1.8} \\
     & ExpeL~\cite{zhao2024expel} & \pmsd{54.6}{3.3} & \pmsd{1.61}{0.21} & \pmsd{6.8}{1.1} & \pmsd{15.5}{0.5} & \pmsd{2.43}{0.07} & \second{\pmsd{5.8}{0.1}} & \pmsd{83.2}{0.4} & \pmsd{1.04}{0.01} & \pmsd{16.9}{0.3} \\
    \midrule
    \multirow{2}{*}{\makecell{Credit /\\ensemble}} & Counterfactual selection~\cite{wang2023data} & \pmsd{46.9}{8.9} & \pmsd{1.44}{0.15} & \pmsd{6.1}{1.1} & \pmsd{14.8}{0.5} & \pmsd{2.29}{0.07} & \pmsd{6.0}{0.0} & \best{\pmsd{88.6}{5.1}} & \pmsd{1.10}{0.04} & \pmsd{14.0}{1.2} \\
     & Hedge~\cite{freund1997decision} & \pmsd{51.3}{3.2} & \second{\pmsd{1.64}{0.07}} & \pmsd{5.6}{0.3} & \pmsd{15.8}{0.3} & \pmsd{2.43}{0.05} & \pmsd{6.0}{0.0} & \pmsd{83.9}{1.6} & \pmsd{1.10}{0.01} & \pmsd{14.2}{0.1} \\
    \midrule
    \rowcolor{oursrow}
     & \ours & \second{\pmsd{59.6}{0.1}} & \best{\pmsd{1.77}{0.01}} & \second{\pmsd{5.5}{0.0}} & \second{\pmsd{19.0}{0.1}} & \second{\pmsd{2.86}{0.02}} & \best{\pmsd{5.7}{0.0}} & \pmsd{79.9}{5.1} & \pmsd{1.14}{0.10} & \pmsd{13.1}{1.5} \\
    \bottomrule
  \end{tabular}
\end{table*}

\subsection{Main Results}\label{sec:main}

\textbf{Overall performance.} Table~\ref{tab:main} (PlantedMem: mean of the five core regimes) shows that \methodname is the best of all 15 methods in PlantedMem and the best LLM-based method on PortBench and InvestorBench. Averaged over the five settings, its utility is 21.2\% higher than that of the best experience-learning agent. In PlantedMem, where experiences carry real signal, the gain is 0.10~pp per month. On PortBench and InvestorBench, no LLM-based method beats 1/$N$: the LLM's drafts fail to outperform equal weighting, and their frequent reallocation costs fees and adds risk (Appendix~\ref{app:why1n}). \methodname is the only LLM-based method that stays close to 1/$N$ there, at most 2.2\% below it, whereas the best existing experience-learning agent falls 15--38\% behind. On PortBench-Full, the lead of 1/$N$ and \methodname depends on the last test month (November 2024), in which cryptocurrencies rose by 129\%: without it, Similarity retrieval (top-4) and Hedge come out slightly ahead of both (Appendix~\ref{app:wealth}). On ClassAlloc, the memory-free LLM beats 1/$N$, but no form of memory improves on it (Appendix~\ref{app:classalloc}); \S\ref{sec:analysis} shows that this advantage lies in the months before the LLM's knowledge cutoff (June 2024).

\textbf{Wealth over time.} Figure~\ref{fig:wealth} follows the wealth of the three investors on PortBench-Full, InvestorBench and ClassAlloc (red shading marks falling markets and the dotted line the knowledge cutoff); on PortBench, whose decisions each start from the equal-weight portfolio, wealth compounds their net returns. For every investor, \methodname tracks 1/$N$ through rising and falling markets, while the other LLM-based methods drift away from it, mostly below. In Table~\ref{tab:risk-main}, \methodname has the best Sharpe ratio and drawdown of the LLM-based methods on every benchmark, and the highest return on PortBench and InvestorBench, where no LLM-based method has a higher return than 1/$N$. Across all 162 runs (both PortBench variants and eight LLMs on InvestorBench and ClassAlloc, each with three investors and three seeds), the wealth of \methodname never fell more than 5\% behind that of 1/$N$, whereas the memory-free LLM and the experience-learning agents did so in 22--44\% of their runs (Appendix~\ref{app:wealth}).

\textbf{Why does \methodname help?} Three components account for this. First, counterfactual attribution removes the market from the contribution. In PlantedMem, counterfactual selection beats selection by outcome credit (MemRL\dag), whereas adjusting outcome credit for the market by regression (Uplift credit) closes only a third of this gap. Second, the trust test keeps \methodname from acting on values that do not carry over to unseen dates. On PortBench and InvestorBench, the values learned from past dates did not predict the following dates, so the test almost never passed (on at most 0.6\% of decisions, Table~\ref{tab:main}). Counterfactual selection, which acts on past contributions without such a test, trails 1/$N$ as much as the other agents. Third, the anchor protects \methodname on these benchmarks. The drafts alone do not: Draft averaging deploys the same eight drafts and falls well behind 1/$N$. With the same anchor, however, every LLM-based baseline also comes within 1\% of \methodname. With signal in PlantedMem, \methodname earns 11\% more than the best anchored baseline, as it acts fully on values that have predicted unseen dates (Appendix~\ref{app:ablation}).

\begin{figure}[t]
  \centering
  \includegraphics[width=0.93\columnwidth]{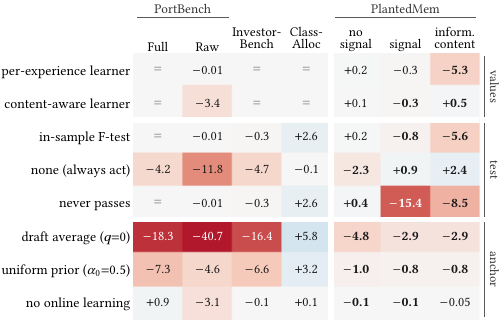}
  \caption{\small Ablation: change in utility when one step of \methodname is changed (\%, mean over seeds; $=$: same actions; bold: $p<0.05$). On real markets, the anchor and the trust test matter most; with signal, acting on the learned values does.}
  \Description{Heat map with eight ablation variants in rows, grouped into value estimation, trust test and anchored action, and seven settings in columns: PortBench-Full, PortBench-Raw, InvestorBench and ClassAlloc, and PlantedMem without signal, with signal and with informative content. Each cell gives the change in utility relative to MemTrial in percent; red cells are losses and blue cells gains. Falling back to the draft average loses 16 to 41 percent on PortBench and InvestorBench, and never acting on the learned values loses 15 percent in PlantedMem with signal.}
  \label{fig:ablation}
\end{figure}

\subsection{Ablation Study}\label{sec:ablation}

Figure~\ref{fig:ablation} changes one step of \methodname at a time; PlantedMem is split into the core regimes without signal (no influence, noise only), those with signal ($\beta>0$), and the regime with 48 experiences and informative content. On real markets, the anchor and the trust test matter: replacing the anchored blend by the draft average costs 16--41\% on PortBench and InvestorBench, and acting on the values without the trust test 4--12\%, as these values do not hold on unseen dates. The prior weight on the reference sets a trade-off: a uniform prior loses 5--7\% on PortBench and InvestorBench but gains 3.2\% on ClassAlloc, where the drafts beat 1/$N$, and the opposite extreme, always keeping the reference, gains where 1/$N$ cannot be beaten but loses where the drafts are better (Appendix~\ref{app:anchor-sens}). Learning the blend online matters less: fixing it at the prior changes the utility by $-3$\% to $+1$\%. On ClassAlloc, where the test passed most often, never acting on values is 2.6\% better (Appendices~\ref{app:ablation} and~\ref{app:case}).

In PlantedMem, the learned values matter where experiences carry signal: never acting on them costs 15\% (8.5\% with informative content), and almost nothing without signal. The trust test costs little there: removing it gains 0.9--2.4\% with signal but loses 2.3\% without. Content-aware value estimation and the trust test matter when each experience is retrieved too rarely for per-experience estimates or for an in-sample F-test (5--6\% with informative content).

\subsection{Robustness to the LLM}\label{sec:robust}

We repeat InvestorBench and ClassAlloc with seven other LLMs: gpt-4.1-nano~\cite{openai2025gpt41}, gpt-5-mini~\cite{openai2025gpt5}, Gemini 2.5 Flash~\cite{comanici2025gemini}, Claude Haiku 4.5~\cite{anthropic2025haiku}, and the open-weight Llama-3.3-70B~\cite{grattafiori2024llama}, Qwen3-235B~\cite{yang2025qwen3} and DeepSeek-V3.1~\cite{deepseek2024v3}. We also run InvestorBench at temperatures 0, 0.3 and 1.0. Appendix~\ref{app:robust-all} gives every LLM, temperature and method. On InvestorBench, no LLM-based baseline beats 1/$N$ in any of the eleven configurations, and \methodname beats all of them in every configuration, by 4--33\% over the best experience-learning agent. On ClassAlloc, \methodname stays within 0.021~pp per month of 1/$N$ with every LLM; it beats the best experience-learning agent with four of the eight LLMs and trails it with the other four. Without its trust test, \methodname does worse in most configurations.

\subsection{Analysis}\label{sec:analysis}

\begin{figure}[t]
  \centering
  \includegraphics[width=0.94\columnwidth]{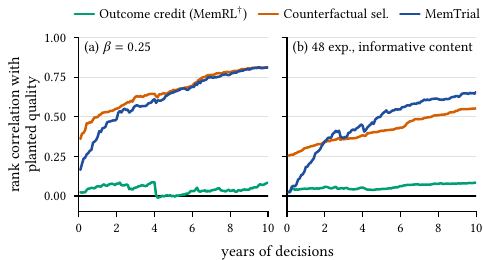}
  \caption{\small Outcome credit does not learn which experiences are useful; counterfactual contributions do, and pooling them by content helps when content is informative.}
  \Description{Two line charts over ten years of PlantedMem decisions (balanced investor, mean over 100 worlds); the y-axis is the rank correlation between the learned values and the planted quality of the experiences. Panel (a), beta 0.25: outcome credit stays below 0.1; counterfactual selection rises from about 0.36 to 0.81; MemTrial rises from about 0.17 to 0.81. Panel (b), 48 experiences with informative content: outcome credit stays below 0.1; counterfactual selection rises from 0.26 to 0.55; MemTrial rises from 0.02 to 0.66 and overtakes counterfactual selection after about two years.}
  \label{fig:recovery}
\end{figure}

\textbf{Does \methodname learn what each experience is worth?} In PlantedMem, the quality of every experience is known, so we can check how well the learned values rank them (Figure~\ref{fig:recovery}; balanced investor). Outcome credit hardly ranks them: after ten years of monthly decisions, the correlation is 0.08 at $\beta=0.25$ and only 0.29 even at $\beta=0.5$, because it follows the markets in which an experience was retrieved. Counterfactual contributions rank them well, and the pooled values of \methodname reach the same level (0.81 at $\beta=0.25$); with 48 experiences whose content is informative, pooling across similar experiences ranks them better than the raw contributions (0.66 against 0.55). Early on, the shrunk values are less accurate and \methodname rarely acts on them; at $\beta=0.25$, its trust rate rises from 12\% in the first year to 91\% in the tenth as evidence accumulates.

\begin{figure}[t]
  \centering
  \includegraphics[width=0.94\columnwidth]{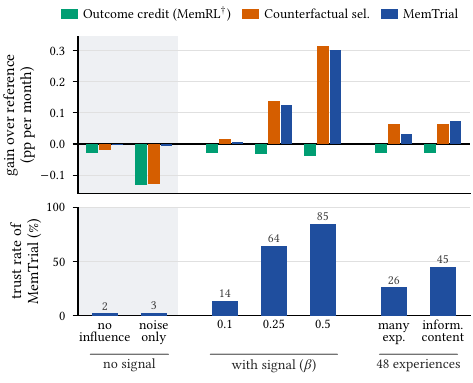}
  \caption{\small PlantedMem by regime. \methodname abstains when experiences carry no signal and acts on its values when they matter.}
  \Description{Two bar charts over seven PlantedMem regimes. Top: gain over the reference in pp per month for outcome credit (MemRL), counterfactual selection and MemTrial. Outcome credit loses in every regime; counterfactual selection loses up to 0.13 when experiences carry no signal and gains up to 0.31 when they matter; MemTrial stays near zero without signal and gains up to 0.30. Bottom: MemTrial's trust rate, 2 to 3 percent without signal and 64 to 85 percent at beta 0.25 and 0.5.}
  \label{fig:regime}
\end{figure}

\textbf{When does \methodname trust its values?} Figure~\ref{fig:regime} shows PlantedMem by regime. When experiences carry no signal, the trust test passes on only 2--3\% of decisions and \methodname stays at the reference, whereas Counterfactual selection and outcome credit fall below it. As experiences matter more, the test passes more often, on up to 85\% of decisions, and \methodname gains up to 0.30~pp per month over the reference. On the real benchmarks, the test responds to whether learned values carry over to unseen dates, not to whether experiences matter: on InvestorBench, experiences change the decision on 85\% of days, yet the test passed on only 0.3\% of decisions. Where it passed most often (15\% of ClassAlloc decisions), the trusted drafts earned $0.05$~pp less (s.e.\ $0.07$) than the anchored blend they replaced (Appendices~\ref{app:classalloc} and~\ref{app:case}).

\textbf{What happens after the LLM's knowledge cutoff?} On ClassAlloc, the memory-free LLM beats 1/$N$ by 0.08~pp per month before the knowledge cutoff of gpt-4.1-mini and trails it by as much afterwards ($p=0.36$ for the difference), and five of the six LLMs whose cutoff is published trail 1/$N$ after theirs. Given only a date, each LLM recalls the asset-class returns of the following month before its cutoff (a look-ahead bias) but hardly at all after it; yet hiding every date from the agent does not remove the advantage before the cutoff (Appendices~\ref{app:cb-llms}--\ref{app:lookahead}, Figure~\ref{fig:cutoff}). Anchored at 1/$N$, \methodname gives up most of the drafts' advantage before the cutoff but loses less than any other LLM-based method after it (Appendix~\ref{app:classalloc}). On PortBench extended to 2026, it is again the best LLM-based method over the eleven months after the cutoff of gpt-4.1-mini (Appendix~\ref{app:pb-forward}).

\textbf{Hyperparameters and cost.} Appendix~\ref{app:hyper} varies the level $\alpha$ of the trust test (Figure~\ref{fig:sens}). Up to the default $\alpha=0.05$, the results on the real benchmarks barely change. Looser levels let chance forward scores pass, on up to 17\% of PortBench-Full decisions, and cost up to 4\% of the utility, whereas in the informative regime of PlantedMem they let \methodname act earlier on values that do hold. The anchor needs a strong prior on the reference (\S\ref{sec:ablation}) but not a precise strength, and more retrieved experiences help only when experiences matter (Appendices~\ref{app:hyper}--\ref{app:k-regime}). A decision costs about USD 0.05 on PortBench; at the same LLM spend, Draft averaging does clearly worse.

\section{Conclusion}\label{sec:conclusion}

When an LLM portfolio agent credits an experience with the outcome of the decisions that used it, the credit tracks the market rather than the experience. \methodname instead measures what each experience changes with a small designed experiment on every date, pools these measurements into values across dates and similar experiences, trusts the values only after they have predicted unseen dates, and until then keeps an anchored blend near a conservative reference. Where the quality of experiences is known, it achieves the highest utility among 15 methods. On real markets, where the learned values did not carry over to unseen dates, it stayed close to 1/$N$, unlike existing experience-learning agents.

\textbf{Limitations.} On the real benchmarks, \methodname's values rarely passed the trust test, and where they did (ClassAlloc), acting on them did not beat the anchor. Its advantage there comes from the anchor, which protects the baselines just as well, so gains from memory on real markets remain to be shown. LLMs recall market outcomes from before their knowledge cutoffs, a look-ahead bias that can flatter any LLM-based method on earlier periods; PlantedMem and the months after the cutoff are free of it. When the LLM's drafts beat the reference, as on ClassAlloc before the cutoff, the anchored action gives up most of their advantage (Appendix~\ref{app:limitations}).

\appendix
\counterwithin{table}{section}
\counterwithin{figure}{section}
\counterwithin{equation}{section}
\counterwithin{algorithm}{section}
\renewcommand{\topfraction}{0.95}
\renewcommand{\bottomfraction}{0.7}
\renewcommand{\textfraction}{0.05}
\renewcommand{\floatpagefraction}{0.8}
\renewcommand{\dbltopfraction}{0.95}
\renewcommand{\dblfloatpagefraction}{0.8}
\setcounter{topnumber}{4}
\setcounter{bottomnumber}{2}
\setcounter{totalnumber}{6}
\setcounter{dbltopnumber}{4}
\setlength{\floatsep}{8pt plus 2pt minus 2pt}
\setlength{\textfloatsep}{10pt plus 2pt minus 3pt}
\setlength{\dblfloatsep}{8pt plus 2pt minus 2pt}
\setlength{\dbltextfloatsep}{10pt plus 2pt minus 3pt}
\setlength{\intextsep}{8pt plus 2pt minus 2pt}

\section*{Appendices}
\noindent\textbf{Contents.} \ref{app:related}~More Related Work (\ref{app:rw-memory}~memory and experience in LLM agents; \ref{app:rw-valuation}~valuation and causal effects; \ref{app:rw-portfolio}~portfolio choice under estimation error; \ref{app:rw-forecast}~forecast evaluation and safe deployment) $\cdot$ \ref{app:notation}~Notation $\cdot$ \ref{app:derivations}~Derivations (\ref{app:design}~the half-fraction design; \ref{app:value}~the value model; \ref{app:trusttest}~the trust test; \ref{app:anchor}~the anchored action) $\cdot$ \ref{app:experiments}~More Experiments (\ref{app:stats}~dataset statistics; \ref{app:baselines}~baselines; \ref{app:impl}~implementation details; \ref{app:why1n}~why LLM agents trail 1/$N$; \ref{app:ablation}~ablation details; \ref{app:wealth}~wealth, market phases and survival; \ref{app:hyper}~hyperparameter analysis; \ref{app:anchor-sens}~strength of the anchor; \ref{app:k-regime}~number of retrieved experiences by regime; \ref{app:regime}~PlantedMem by regime; \ref{app:classalloc}~ClassAlloc before and after the knowledge cutoff; \ref{app:robust-all}~other LLMs and temperatures; \ref{app:cb-llms}~ClassAlloc with other LLMs; \ref{app:lookahead}~look-ahead checks; \ref{app:pb-forward}~PortBench after the knowledge cutoff) $\cdot$ \ref{app:case}~Case Study $\cdot$ \ref{app:algorithms}~Algorithms $\cdot$ \ref{app:prompts}~LLM Prompts (\ref{app:prompts-ib}~InvestorBench; \ref{app:prompts-cb}~ClassAlloc; \ref{app:prompts-pb}~PortBench) $\cdot$ \ref{app:limitations}~Limitations.

\section{More Related Work}\label{app:related}

\subsection{Memory and experience in LLM agents}\label{app:rw-memory}
In more detail than \S\ref{sec:related}: MemGPT~\cite{packer2023memgpt} manages a hierarchy of memory tiers for long interactions, MemoryBank~\cite{zhong2024memorybank} stores memories and lets them fade with a rule inspired by the forgetting curve, and Voyager~\cite{wang2023voyager} builds a library of executable skills that it reuses when they succeed; in finance, FinCon~\cite{yu2024fincon} updates the investment beliefs of a team of LLM agents by verbal reinforcement from realized profit and loss. These methods decide what to store, keep or reuse from the success of the episodes in which a memory took part, i.e., by outcome credit, and none of them tests whether its rule for using memory predicts future gains before acting on it.

\subsection{Valuation and causal effects}\label{app:rw-valuation}
The Shapley value~\cite{shapley1953value} and the Banzhaf value~\cite{banzhaf1964weighted} are the two classic ways to divide a joint payoff among players. Data valuation adapts them to training data~\cite{ghorbani2019data,jia2019towards,kwon2021beta,wang2023data}: Beta Shapley reweights subsets by their size to reduce noise, and Data Banzhaf weights all subsets equally, which makes it the most robust semivalue when the utility is noisy. Our counterfactual contribution~\eqref{eq:contribution} is the main effect of a factor in a $2^{4-1}$ fractional factorial experiment~\cite{box2008estadistica}, the classic way to estimate several effects at once with few runs; Appendix~\ref{app:design} shows that it equals the Banzhaf value when the three-way interaction aliased with it is negligible. Estimating what an experience does to a decision is also a problem of heterogeneous treatment effects~\cite{athey2016recursive,kunzel2019metalearners}. Uplift modelling~\cite{gutierrez2017causal} and counterfactual learning from logged decisions~\cite{bottou2013counterfactual,swaminathan2015counterfactual} estimate such effects from observational data, and online controlled experiments~\cite{kohavi2009controlled} randomize users into groups. \methodname instead runs the whole designed experiment on every date, so the market is held fixed within each comparison.

\subsection{Portfolio choice under estimation error}\label{app:rw-portfolio}
Mean--variance optimization~\cite{markowitz1955portfolio} is sensitive to errors in the estimated means and covariances, which has motivated shrinkage estimators~\cite{jorion1986bayes,ledoit2003honey}, portfolios that account for parameter and model uncertainty~\cite{kan2007optimal,garlappi2007portfolio}, and combinations of an optimized portfolio with 1/$N$~\cite{tu2011markowitz}. \citet{demiguel2009optimal} find that none of the optimized portfolios they evaluate consistently beats 1/$N$ out of sample. Online portfolio selection~\cite{cover1991universal,helmbold1998line,li2014online} adapts the weights over time with regret guarantees. The anchored action of \methodname combines two of these ideas: shrinkage towards a conservative reference, and an online blend learned with a regularizer.

\subsection{Forecast evaluation and safe deployment}\label{app:rw-forecast}
Prequential analysis~\cite{dawid1984present} judges a model by its sequential out-of-sample predictions; tests of predictive accuracy~\cite{diebold2002comparing} and proper scoring rules~\cite{gneiting2007strictly} formalize such judgments. The trust test of \methodname applies this idea to the values of experiences, as a one-sided test of whether their out-of-sample correlations with the measured contributions are positive on average. Safe policy improvement~\cite{thomas2015high,laroche2019safe} deploys a new policy only when it improves on a baseline with high confidence; in \methodname, the trust test plays the role of this confidence check, and the anchored action that of the baseline policy. Finally, shrinkage across units~\cite{efron1975data,fay1979estimates} underlies our value model.

\section{Notation}\label{app:notation}

\begin{table}[h]
  \caption{Notation.}
  \label{tab:notation}
  \small
  \begin{tabular}{@{}L{0.30\columnwidth}L{0.66\columnwidth}@{}}
    \toprule
    Symbol & Meaning \\
    \midrule
    $t$, $T$ & decision date, number of dates \\
    $s_t$ & market state at date $t$ \\
    $N$ & number of assets (plus cash) \\
    $\theta=(M,m,\gamma)$ & investor profile: maximum share of risky assets, minimum share of defensive assets, risk aversion \\
    $P_\theta$ & projection of a portfolio onto the investor's mandate \\
    $u_t(x)$; $r_t$, $V_t$ & investor utility (certainty-equivalent return) of portfolio $x$, Eq.~\eqref{eq:utility}; net return and realized variance over the holding period \\
    $\pi$; $x\sim\pi(\cdot\mid s_t,S)$ & LLM policy; a draft made with the experiences in $S$ \\
    $k$; $R_t=(h_{t,1},\dots,h_{t,k})$ & number of retrieved experiences (4); experiences retrieved at date $t$ \\
    $\design$ & designed subsets, Eq.~\eqref{eq:design} \\
    $U_t(S)$ & utility of the draft made with subset $S$, at realized prices \\
    $c_{t,j}$; $y_{t,j}$ & counterfactual contribution of the $j$-th retrieved experience, Eq.~\eqref{eq:contribution}; centred contribution, Eq.~\eqref{eq:centre} \\
    $v_h$; $\hat v_h$ & value of experience $h$; its posterior mean, Eq.~\eqref{eq:posterior} \\
    $z_h\in\R^d$ & content feature of experience $h$ ($d=8$ principal components of its embedding) \\
    $\beta$; $b_h$ & map from content to value; experience-specific deviation \\
    $\sigma^2$; $\sigma_\beta^2$, $\tau^2$ & noise variance; prior variances, Eq.~\eqref{eq:fh} \\
    $n_h$; $m_h$ & number of dates with a realized outcome on which $h$ was retrieved; mean of its centred contributions \\
    $\rho_t$; $n$; $\alpha$ & forward score at date $t$, Eq.~\eqref{eq:forward}; number of forward scores; level of the trust test, Eq.~\eqref{eq:ttest} \\
    $S^*_t$ & trusted subset, Eq.~\eqref{eq:trusted} \\
    $\xref$; $\bar x_t$ & conservative reference; average draft \\
    $q_t$; $\alpha_0$; $\lambda_0$ & weight on the reference; its prior; regularization strength, Eq.~\eqref{eq:ftrl} \\
    \bottomrule
  \end{tabular}
\end{table}

\section{Derivations}\label{app:derivations}

\subsection{The half-fraction design}\label{app:design}
Code the presence of the $j$-th retrieved experience in a subset $S$ as $x_j=+1$ if $j\in S$ and $x_j=-1$ otherwise, and write the utility of a draft as a full factorial model,
\begin{multline}\label{eq:factorial}
  U_t(S) = \mu_t + \sum_{j} \tfrac{a_j}{2}\, x_j + \sum_{j<l} \tfrac{a_{jl}}{2}\, x_j x_l \\
  + \sum_{j<l<m} \tfrac{a_{jlm}}{2}\, x_j x_l x_m + \tfrac{a_{1234}}{2}\, x_1 x_2 x_3 x_4 + \varepsilon_{t,S},
\end{multline}
where $\mu_t$ collects everything the drafts of date $t$ share, including the market move, $a_j$ is the main effect of experience $j$ (over the 16 subsets, the mean utility of those with it minus that of those without it), the other $a$'s are interactions, and $\varepsilon_{t,S}$ is the sampling noise of the draft. Over all 16 subsets, the main effect equals the Banzhaf value of $j$, because both are the mean over subsets with $j$ minus the mean over subsets without it. Since $k=4$ is even, a subset has an even number of experiences exactly when $x_1x_2x_3x_4=+1$, so the design~\eqref{eq:design} is the half fraction with defining relation $I=x_1x_2x_3x_4$. Multiplying by $x_j$ shows that on the design $x_j$ equals the product of the other three columns: each main effect is aliased with the three-way interaction of the other three experiences, and each two-way interaction with the complementary one ($x_1x_2=x_3x_4$), which is resolution IV. The contribution~\eqref{eq:contribution} is the contrast of column $x_j$ over the design, so under~\eqref{eq:factorial}
\begin{equation}\label{eq:alias}
  c_{t,j} = a_j + a_{(-j)} + \bar\varepsilon^{+}_{t,j} - \bar\varepsilon^{-}_{t,j},
\end{equation}
where $a_{(-j)}$ is the three-way interaction of the other three experiences and $\bar\varepsilon^{+}$, $\bar\varepsilon^{-}$ are the mean noise of the four drafts with and the four without experience $j$. Three consequences follow. First, $\mu_t$, everything the drafts of date $t$ share, cancels; the market move enters $c_{t,j}$ only through the weights that experience $j$ changes. Second, $c_{t,j}$ estimates the main effect without bias when three-way interactions are negligible, and the four contrasts are orthogonal to each other and to every two-way interaction, so the contribution of one experience is not affected by two-way interactions. Third, the noise term has variance $\sigma_\varepsilon^2/2$ for draft-level noise variance $\sigma_\varepsilon^2$, against $\sigma_\varepsilon^2/4$ for the full factorial with twice as many drafts. On ClassAlloc and InvestorBench, where all 16 subsets were drafted on every date and seed, the difference between $c_{t,j}$ and the full-factorial value, half the contrast between the two half fractions, has 1.26--1.34 times the mean square that sampling noise alone gives ($\sigma_\varepsilon^2/4$, with $\sigma_\varepsilon^2$ estimated from the spread of a draft's utility across seeds): the three-way interaction aliased with the main effect adds 13--17\% to the noise variance of a contribution, and $c_{t,j}$ correlates with the full-factorial value at 0.76--0.86.

\subsection{The value model}\label{app:value}
Stack the $H$ experiences that have at least one measured contribution. Under Eq.~\eqref{eq:fh}, the values $v=Z\beta+b$ and the observed means $m$ are jointly Gaussian with
\begin{equation}\label{eq:joint}
  \begin{aligned}
    v &\sim \Normal(0, K_v), & K_v &= \sigma_\beta^2 ZZ^{\top} + \tau^2 I,\\
    m &\sim \Normal(0, K), & K &= K_v + \sigma^2 \diag(1/n_h),
  \end{aligned}
\end{equation}
which gives the marginal likelihood~\eqref{eq:marglik}. Gaussian conditioning gives $\mathbb{E}[v\mid m]=K_vK^{-1}m$ and $\mathbb{E}[\beta\mid m]=\sigma_\beta^2Z^{\top}K^{-1}m$, i.e., Eq.~\eqref{eq:posterior}. For a new experience $h'$ without measurements, $\operatorname{cov}(v_{h'},m)=\sigma_\beta^2 z_{h'}^{\top}Z^{\top}$, so $\mathbb{E}[v_{h'}\mid m]=z_{h'}^{\top}\hat\beta$: its value is predicted from its content alone. Without content ($\sigma_\beta^2=0$), $K$ is diagonal and
\begin{equation}\label{eq:shrink}
  \hat v_h = \frac{\tau^2}{\tau^2 + \sigma^2/n_h}\, m_h,
\end{equation}
the James--Stein-type shrinkage of the Fay--Herriot model~\cite{fay1979estimates}: an experience measured on few dates is shrunk towards zero, which after the centring~\eqref{eq:centre} is the value of an average experience. The noise variance $\sigma^2$ is the pooled within-experience variance of the centred contributions, and $(\sigma_\beta^2,\tau^2)$ maximize~\eqref{eq:marglik} over a grid.

\subsection{The trust test}\label{app:trusttest}
Suppose that the learned values carry no information about future contributions, so that every forward score has mean zero. If the scores were independent and approximately normal, the statistic in Eq.~\eqref{eq:ttest} would follow a Student $t$ distribution with $n-1$ degrees of freedom, and a single test at a fixed $n$ would pass with probability at most $\alpha$. \methodname uses the test as a decision rule: it applies the test on every date with $n\ge10$, to the scores of the learner whose mean score is higher. Scores of nearby dates are also dependent when the same experiences recur. The level $\alpha$ is therefore the level of one test, not the rate at which the rule passes over a run. That rate is measured where every pass is a false one: in the two PlantedMem regimes in which experiences carry no signal, the rule passed on 2.3\% and 3.0\% of decisions (Table~\ref{tab:trust-rate}). The minimum of 10 scores and the level $\alpha=0.05$ were fixed on the development seeds, and Table~\ref{tab:trust-sens} varies both.

\subsection{The anchored action}\label{app:anchor}
Both the reference and the average draft satisfy the mandate, which is a convex set, so every blend $x_t(q)$ does too. Because $q$ is a single number, Eq.~\eqref{eq:ftrl} is solved by evaluating the objective on a grid of 201 values, which finds its maximizer up to the grid spacing without any assumption on its shape; the utility of every blend is computed exactly on the realized price path, including trading costs. Before any outcome is realized, $q_t=\alpha_0$. As $n$ grows, $\lambda_n=\lambda_0\hat\sigma_n/\sqrt{n}$ tends to zero, so the weight of the prior fades and $q_t$ approaches the blend with the highest mean utility over the past dates, as in follow-the-regularized-leader~\cite{shalev2025online}. Scaling by $\hat\sigma_n$ makes $\lambda_0$ independent of the units of utility, so the same $\lambda_0=4$ is used on all benchmarks.

\section{More Experiments}\label{app:experiments}

\emph{Improv.} is the gain of \methodname over the best experience-learning agent of each setting, in percent of that agent's utility.

\subsection{Dataset statistics}\label{app:stats}
Table~\ref{tab:stats} gives the size of each benchmark and the cost of our runs. The smallest sample is the PortBench test period (20 months): with significance tested across test dates, none of the PortBench differences in Table~\ref{tab:main} is significant. The ten further InvestorBench runs of \S\ref{sec:robust} make 8,940 executions each (the 16 subsets, ExpeL and the three sequential agents, over 149 days and three seeds) and cost USD 1.21--16.90 each; the seven further ClassAlloc runs make 4,318 executions each and cost USD 0.38--6.01 each, and the two extra memory-free draws per date and seed of Self-consistency cost USD 6.76 in all. Replaying the sequential agents of all ten ClassAlloc runs (eight LLMs and two with dates hidden, Appendix~\ref{app:lookahead}) with the feedback timing of Appendix~\ref{app:impl} cost USD 1.26.

\begin{table*}[t]
  \caption{Benchmark statistics and cost.}
  \label{tab:stats}
  \footnotesize
  \begin{tabular}{@{}L{2.1cm}L{4.1cm}L{3.2cm}L{3.4cm}L{3.2cm}@{}}
    \toprule
     & PortBench (Full / Raw) & InvestorBench & ClassAlloc & PlantedMem \\
    \midrule
    Decision dates & 59 monthly, Aug 2019 -- Dec 2024 (a month whose 20-day window would overlap the previous one is skipped); 56 usable per configuration & 63 warm-up and 149 test trading days, Jul 2020 -- May 2021 & 129 monthly, Jul 2015 -- Mar 2026: 58 warm-up and 71 test & 132 monthly, Apr 2015 -- Mar 2026, per episode; the first 12 are a warm-up \\
    Development / test & 36 / 20 months & none (fully held out) / 149 days & none (fully held out) / 71 months & seeds 0--14 / 15--114 \\
    Assets & 99 in 6 classes, including cash & HON, JNJ, MSFT, UVV and cash & equities, bonds, commodities, real estate and cash: equal-weighted indices of 89 assets & equities, bonds, cash, commodities \\
    Experiences & 338 records of earlier decisions (Apr 2015 -- Jul 2019); 29 retrieved & 63 lessons written in the warm-up & 58 lessons written in the warm-up & 12 or 48 planted per episode \\
    Drafts per date and seed & 19: the 16 subsets of the 4 retrieved experiences, FinMem, MemRL and the benchmark's native memory; Reflexion runs separately & 20: the 16 subsets, ExpeL, FinMem, MemRL and Reflexion & 20: the 16 subsets, ExpeL, FinMem, MemRL and Reflexion & resampled from 2,828 logged LLM allocations \\
    LLM executions & 6,612 (6,518 valid) and 354 Reflexion decisions & 9,003 & 4,318 (58 in the warm-up), and 210 redrawn (Appendix~\ref{app:impl}) & none \\
    API cost & about USD 43.6 & USD 5.09 & USD 1.61, and USD 0.12 to redraw & none \\
    Retrieval ($k=4$) & Qwen3 embeddings & text-embedding-3-small & text-embedding-3-small & regime match with noise \\
    Trading costs (15~bp) & on the distance from equal weights & on the amount traded & on the amount traded & on the distance from equal weights \\
    Seeds & 3 & 3 & 3 & 100 (and 15 for development) \\
    Evaluated test decisions per method & 360 (20 months $\times$ 3 investors $\times$ 3 seeds $\times$ 2 configurations) & 1,341 (149 days $\times$ 3 investors $\times$ 3 seeds) & 639 (71 months $\times$ 3 investors $\times$ 3 seeds) & 252,000 (100 seeds $\times$ 120 months $\times$ 7 regimes $\times$ 3 investors) \\
    \bottomrule
  \end{tabular}
\end{table*}

\subsection{Baselines}\label{app:baselines}
Several methods are evaluated on the same logged drafts, so that they differ only in how they use memory. \emph{1/$N$}~\cite{demiguel2009optimal} holds equal weights in all assets and cash, projected to the mandate. \emph{Minimum variance}~\cite{markowitz1955portfolio} is the long-only minimum-variance portfolio estimated from the past 60 daily returns, projected to the mandate. \emph{Memory-free} (zero-shot)~\cite{brown2020language} deploys the draft made without experiences. \emph{Self-consistency}~\cite{wang2022self} averages memory-free drafts of the same decision: on the real benchmarks, three per seed (the seed's own draft and two further draws, made after the main runs with the same prompt, LLM and temperature); in PlantedMem, eight. \emph{Similarity retrieval}~\cite{lewis2020retrieval} deploys the draft made with the two most similar or with all four retrieved experiences. \emph{FinMem}~\cite{yu2025finmem} ranks memories by similarity, recency and an importance score that is reinforced when the decision that used them earns a positive return. \emph{MemRL}~\cite{zhang2026memrl} learns a value for each memory from the rewards of the decisions that used it and selects among the most similar candidates, with $\varepsilon$-greedy exploration. \emph{Reflexion}~\cite{shinn2023reflexion} writes a verbal reflection after each outcome and conditions later decisions on the most recent reflections. \emph{ExpeL}~\cite{zhao2024expel} distils insights from successful and failed decisions; on InvestorBench and ClassAlloc the LLM writes them, and on PortBench and in PlantedMem we use its insight votes (\dag): an experience gains a vote when the decision it took part in beats 1/$N$ and loses one otherwise, and the two experiences with the most votes are used. \emph{Uplift credit}~\cite{gutierrez2017causal} regresses the utility of the deployed decision on indicators of which experiences were used, with the utility of 1/$N$ as a market control and a small ridge penalty, and uses the two experiences with the highest coefficients, exploring a random pair on 10\% of dates. \emph{Counterfactual selection}~\cite{wang2023data} measures the Banzhaf contribution of each retrieved experience on every date with a realized outcome and deploys the draft with the two experiences of highest mean past contribution, without value estimation, trust test or anchor. \emph{Draft averaging}~\cite{bates1969combination} deploys the average of \methodname's eight designed drafts. \emph{Hedge}~\cite{freund1997decision} mixes 1/$N$ and the portfolios of the memory configurations available on each benchmark (memory-free, top-4 retrieval, FinMem and MemRL, plus the benchmark's native memory on PortBench) with exponential weights, whose learning rate is scaled as in Eq.~\eqref{eq:ftrl}. On PortBench, FinMem and MemRL select experiences from the same pool with their own memory modules; where a baseline's own module cannot run (\dag{} in Table~\ref{tab:main}), we re-implement its memory-scoring rule over the same experience pool. The baselines learn as follows. On PortBench, the FinMem and MemRL modules receive no feedback after July 2019, before the first decision date, and score the pool by the outcomes recorded up to then; ExpeL\dag{} is the vote rule above (ExpeL-style voting), and Reflexion writes a reflection once the outcome of each decision is known. On InvestorBench and ClassAlloc, FinMem\dag{} and MemRL\dag{} update their importance and values, and Reflexion its reflections, once the outcome of each decision is known; FinMem\dag{} and MemRL\dag{} learn from the return of their own draft over the holding period, before the mandate and fees. ExpeL's insights are written once from the warm-up lessons. In PlantedMem, every adapted agent updates after every period. Uplift credit, Counterfactual selection, Hedge and \methodname learn from every realized outcome on all benchmarks.

\subsection{Implementation details}\label{app:impl}
\begin{itemize}
  \item \emph{Agents.} All LLM agents use gpt-4.1-mini (2025-04-14)~\cite{openai2025gpt41} at temperature 0.7. On PortBench, a draft is one run of the benchmark's staged agent (market interpretation, signal generation and weight optimization), which makes three LLM calls with JSON output; the retrieved experiences enter the prompt as a memory block of at most 4,096 tokens. PortBench-Raw uses the benchmark's native prompt, which shows a price table of eight assets, and PortBench-Full extends the table to all eligible assets. On InvestorBench, a draft is one chat completion that returns a JSON object with the weights of the four stocks and cash; the prompt gives, for each stock, the close, the 1-, 5- and 20-day returns, the 20-day volatility and up to three news headlines of the day, and an invalid reply is retried once. The runs with other LLMs (\S\ref{sec:robust}) also use gpt-4.1-nano (2025-04-14)~\cite{openai2025gpt41}, gpt-5-mini (2025-08-07; minimal reasoning effort, temperature fixed by the API)~\cite{openai2025gpt5}, and, through OpenRouter, Llama-3.3-70B-Instruct~\cite{grattafiori2024llama}, Qwen3-235B-A22B-Instruct-2507~\cite{yang2025qwen3}, DeepSeek-V3.1 (non-thinking mode)~\cite{deepseek2024v3}, Gemini 2.5 Flash (thinking budget 0)~\cite{comanici2025gemini} and Claude Haiku 4.5~\cite{anthropic2025haiku}, all with unchanged prompts; on InvestorBench the lessons and their embeddings are those of the main run, and on ClassAlloc each LLM writes its own lessons; the ExpeL insights and Reflexion notes are always written by the LLM under test.
  \item \emph{ClassAlloc.} Each class index averages, every day, the daily returns of PortBench's assets in that class that have prices on both days (44 equities, 15 bond, 16 commodity, 10 real-estate and 4 cash assets); an asset listed after January 2015 enters its index only after 20 sessions of prices, because the first quotes of new listings can be wrong (one coin gains 10,200\% on its third day). Decisions are made on the first session of each month from July 2015 (on 6 July, the first session with six months of history) to March 2026, the last month with 20 sessions of outcome in the data. Each decision is scored over the 20 sessions after it, and the holdings are carried to the next decision date. Every learner uses an outcome only after its 20 sessions have passed: in 9 of the 70 test months the next decision comes after 19 sessions, and that outcome is first used at the decision after next. The sequential agents of our runs (FinMem, MemRL and Reflexion) were replayed under this timing from the logged runs: a logged decision whose prompt is unchanged is kept, and any other decision, with the reflection written after it, is drawn anew (210 of 639 in the main run). The mandate treats equities and real estate as risky and bonds and cash as defensive, as PortBench does. Each lesson is written after the agent sees the returns of the five classes over the 20 holding sessions, its own portfolio return and that of the equal-weight portfolio, and is retrieved by text-embedding-3-small~\cite{openai2024embedding} as on InvestorBench. FinMem's recency decays over 4.5 months (about 90 trading days); all other settings are as on InvestorBench.
  \item \emph{PlantedMem.} An episode runs over 132 monthly dates (Table~\ref{tab:stats}) with four asset classes, the equal-weighted indices of PortBench's equities, bonds, cash and commodities, and 20-session outcomes; every decision starts from the equal-weight portfolio, and its outcome is known before the next one. Of the 12 (or 48) planted experiences, a quarter are useful, half neutral and a quarter misleading, in random order. Each experience has a home regime, drawn independently of its type; on date $t$ it receives the retrieval score $\mathbb{1}[\text{home}=g_t]+\epsilon$ with $\epsilon\sim\Normal(0,0.5^2)$, where $g_t=1$ if the equities rose on average over the previous 60 sessions, and the four highest scores are retrieved. On each date, a retrieved experience points to an asset class $c_{t,h}$: a useful (misleading) experience points with probability 0.7 to the class with the highest (lowest) return over the coming 20 sessions, and otherwise, like a neutral one, to a class drawn uniformly. A draft made with the experiences $S$ is $x^S=(1-\beta)\,b+\beta\,\frac{1}{|S|}\sum_{h\in S}e_{c_{t,h}}$, where $b$ is an allocation drawn for each draft from 2,828 logged gpt-4.1-mini allocations of PortBench (aggregated to the four classes) and $e_c$ is the indicator of class $c$; the memory-free draft is $b$. The regimes set $\beta\in\{0,0.1,0.25,0.5\}$, and \emph{noise only} uses $\beta=0.5$ with every experience neutral. The content feature is $z_h=\kappa\,q_h e_1+\xi_h$, with type $q_h\in\{+1,0,-1\}$, $\xi_h\sim\Normal(0,I_8)$, $\kappa=1.5$ in the regime with informative content and 0 otherwise. Usefulness is thus defined by the direction in which an experience moves the draft, and its gain in utility depends on the date and the investor. The adapted baselines act on the same retrieved experiences: FinMem\dag{} adds an importance that rises when the decision beats 1/$N$, MemRL\dag{} learns values from the utilities of the decisions that used the experiences, ExpeL\dag{} votes, and Reflexion\dag{} tilts the draft towards the classes that did best in the last three periods.
  \item \emph{Retrieval.} On PortBench, the query is a short JSON text with the task description, the date, the public market features, the number of assets and the risk profile of the benchmark task. Each experience is the JSON record of an earlier decision: its allocation by asset class, the public market features at the time and the realized 20-day outcome. Both are embedded with Qwen3-Embedding-0.6B~\cite{zhang2025qwen3} (1,024 dimensions, run locally), and the four records with the highest cosine similarity are retrieved. All 338 records became available between April 2015 and July 2019, before the first decision date. On InvestorBench, the query is the day's market context described above, each experience is a lesson of at most 700 characters that the agent wrote after seeing the outcome of a warm-up decision, and both are embedded with text-embedding-3-small. Ties are broken by experience identifier.
  \item \emph{Mandate.} On PortBench, $P_\theta$ is the Euclidean projection onto the long-only portfolios whose risky share (equities, real estate and cryptocurrency) is at most $M$ and whose defensive share (bonds and cash) is at least $m$, computed by alternating projections. On InvestorBench, negative weights are set to zero and, if the stock share exceeds $\min(M,1-m)$, the stock weights are scaled down to this cap and the rest goes to cash.
  \item \emph{Trading costs and utility.} Costs are 15~bp of the L1 distance moved. Each PortBench decision is an episode that starts from the equal-weight portfolio, as in the benchmark, so the distance is measured from it; PlantedMem does the same. On InvestorBench, holdings drift with prices overnight and the distance is measured from the drifted holdings, so every method pays for what it actually trades. The variance term of Eq.~\eqref{eq:utility} is the variance of daily portfolio returns times the number of days held (20); on InvestorBench, with one day held, it is the squared daily return.
  \item \emph{\methodname.} Content features $z_h$ are the first $d=8$ principal components of the retrieval embeddings of all experiences, standardized; in PlantedMem, the first coordinate carries $\kappa$ times the experience's quality plus standard normal noise and the others are noise. $(\sigma_\beta^2,\tau^2)$ maximize the marginal likelihood~\eqref{eq:marglik} over a small grid; Eq.~\eqref{eq:ftrl} is solved over 201 values of $q$. All learning uses only outcomes realized before the decision date.
  \item \emph{Seeds.} A seed means different things on the two kinds of benchmark. On the real benchmarks, a seed is an independent deployment in which every LLM draft is sampled anew. There is only one market history, so additional seeds measure only the randomness of the LLM, and each one costs a full set of paid calls (about USD 14.5 per seed on PortBench, USD 43.6 for all methods and seeds); we use three seeds and test significance across test dates, where the randomness of the market lies. In PlantedMem, a seed is a newly generated world (which experiences are useful, the retrieval noise and the resampled drafts), which costs no LLM call, so we use 100 test seeds to make the estimates precise. A standard deviation shown as 0.00 is below 0.005; this happens when \methodname stays anchored at the reference, which does not depend on the seed.
  \item \emph{LLM cost and learning overhead.} \methodname makes eight drafts per decision. The drafts are independent and run in parallel, so the latency of a decision is that of a single draft. With gpt-4.1-mini, one draft costs about USD 0.006 on PortBench and USD 0.0006 on InvestorBench, whose prompts are shorter, so a \methodname decision costs about USD 0.05 and USD 0.005, respectively; the complete InvestorBench run, covering all 15 methods with three seeds, cost USD 5.09, and the ClassAlloc run USD 1.61 (Appendix~\ref{app:stats}). At equal LLM spend, the relevant comparison is Draft averaging, which uses the same eight drafts and is 0.36 and 0.81~pp per month worse than \methodname on PortBench, 1.70~bp per day worse on InvestorBench and 0.12~pp worse in PlantedMem (Table~\ref{tab:main}). Fitting both value models, scoring the trust test and solving Eq.~\eqref{eq:ftrl} takes 3.6~ms per decision on average with 12 experiences and 6.9~ms with 48 experiences (95th percentiles 6.2 and 10.4~ms, one CPU core), and recording a realized outcome takes 0.1~ms, which is negligible next to a single LLM call.
  \item \emph{Statistics.} Mean $\pm$ sd over seeds; one-sided paired $t$-tests over test dates on the real benchmarks (after averaging over seeds and investors) and over seeds in PlantedMem. Every number in the paper is computed from the logged drafts; after \methodname was frozen, only the names in its source file were changed (the sha256 of the file before and after is recorded with the code).
\end{itemize}

\subsection{Why do LLM agents trail 1/$N$?}\label{app:why1n}
Table~\ref{tab:decomp} splits the test-period utility into gross return, trading cost and risk penalty (utility = gross return $-$ cost $-$ risk penalty, up to rounding; mean $\pm$ sd over seeds, each averaged over the investors, in \% per month on PortBench and bp per day on InvestorBench). L1 is the mean distance from the equal-weight portfolio, on which PortBench charges costs. On PortBench-Full, the memory-free draft, the top-4 retrieval draft and the average of the eight designed drafts trail 1/$N$ by 0.33--0.42~pp per month, and the gap has three parts of similar size: 0.06--0.16~pp less gross return, 0.13--0.15~pp more trading cost and a 0.11--0.14~pp larger risk penalty. Each month, the LLM moves about half of the capital away from the equal-weight portfolio (an L1 distance of 1.0--1.2, against 0.17 for 1/$N$ projected to the mandate), which costs money and adds risk without earning more. On PortBench-Raw, whose prompt shows only eight assets, the drafts are also more defensive than 1/$N$: they earn 0.72--0.85~pp less gross return at a slightly smaller risk penalty. On InvestorBench, the other LLM-based methods earn about the same gross return as 1/$N$ (10.7--11.7 against 11.4~bp per day), but they change their portfolios substantially every day and pay 1.0--2.5~bp per day in trading costs, against 0.1~bp for 1/$N$ and 0.2~bp for \methodname. With trading costs set to zero for all methods, \methodname included, every LLM draft still trails 1/$N$ on PortBench, by 0.17--0.27~pp per month on Full and 0.68--0.81~pp on Raw, while \methodname stays within 0.04~pp of it. In short, the drafts do not pick better assets than equal weighting, and their frequent reallocation costs money. Experience learning would have to close this gap to beat 1/$N$. Outcome credit does not close it, and \methodname's trust test finds that its learned values do not hold on unseen dates, so it stays near the reference.

\begin{table*}[t]
  \caption{Decomposition of the test-period utility. The LLM drafts trail 1/$N$ in all three parts on PortBench-Full, mostly in gross return on PortBench-Raw and mostly in trading costs on InvestorBench.}
  \label{tab:decomp}
  \footnotesize
  \setlength{\tabcolsep}{5pt}
  \begin{tabular}{@{}llccccc@{}}
    \toprule
    Benchmark & Method & Utility & \makecell{Gross\\return} & \makecell{Trading\\cost} & \makecell{Risk\\penalty} & L1 \\
    \midrule
    \multirow{4}{*}{PortBench-Full} & 1/$N$ & \pmsd{2.02}{0.00} & \pmsd{2.25}{0.00} & \pmsd{0.03}{0.00} & \pmsd{0.21}{0.00} & \pmsd{0.17}{0.00} \\
      & Memory-free & \pmsd{1.60}{0.05} & \pmsd{2.12}{0.05} & \pmsd{0.17}{0.02} & \pmsd{0.34}{0.03} & \pmsd{1.13}{0.11} \\
      & Sim.\ retrieval (top-4) & \pmsd{1.70}{0.06} & \pmsd{2.19}{0.04} & \pmsd{0.18}{0.02} & \pmsd{0.31}{0.02} & \pmsd{1.17}{0.14} \\
      & Draft averaging & \pmsd{1.62}{0.05} & \pmsd{2.09}{0.06} & \pmsd{0.16}{0.00} & \pmsd{0.32}{0.02} & \pmsd{1.02}{0.01} \\
    \midrule
    \multirow{4}{*}{PortBench-Raw} & 1/$N$ & \pmsd{2.02}{0.00} & \pmsd{2.25}{0.00} & \pmsd{0.03}{0.00} & \pmsd{0.21}{0.00} & \pmsd{0.17}{0.00} \\
      & Memory-free & \pmsd{1.07}{0.02} & \pmsd{1.43}{0.03} & \pmsd{0.18}{0.01} & \pmsd{0.17}{0.00} & \pmsd{1.19}{0.08} \\
      & Sim.\ retrieval (top-4) & \pmsd{1.05}{0.02} & \pmsd{1.40}{0.02} & \pmsd{0.19}{0.01} & \pmsd{0.17}{0.00} & \pmsd{1.22}{0.05} \\
      & Draft averaging & \pmsd{1.19}{0.02} & \pmsd{1.53}{0.02} & \pmsd{0.17}{0.00} & \pmsd{0.17}{0.00} & \pmsd{1.13}{0.03} \\
    \midrule
    \multirow{6}{*}{InvestorBench} & 1/$N$ & \pmsd{10.45}{0.00} & \pmsd{11.44}{0.00} & \pmsd{0.11}{0.00} & \pmsd{0.88}{0.00} & -- \\
      & Memory-free & \pmsd{8.43}{0.29} & \pmsd{11.53}{0.31} & \pmsd{2.16}{0.07} & \pmsd{0.94}{0.01} & -- \\
      & Sim.\ retrieval (top-4) & \pmsd{8.12}{0.61} & \pmsd{10.74}{0.58} & \pmsd{1.72}{0.05} & \pmsd{0.90}{0.01} & -- \\
      & Draft averaging & \pmsd{8.69}{0.09} & \pmsd{11.23}{0.11} & \pmsd{1.62}{0.03} & \pmsd{0.91}{0.00} & -- \\
      & ExpeL & \pmsd{8.93}{0.32} & \pmsd{10.78}{0.34} & \pmsd{0.96}{0.02} & \pmsd{0.89}{0.00} & -- \\
      & \ours & \pmsd{10.40}{0.04} & \pmsd{11.44}{0.06} & \pmsd{0.16}{0.01} & \pmsd{0.88}{0.00} & -- \\
    \bottomrule
  \end{tabular}
\end{table*}

\subsection{Ablation details}\label{app:ablation}
This appendix completes \S\ref{sec:ablation} (Figure~\ref{fig:ablation}). Table~\ref{tab:trust-all} gives the trust rate of every variant, the share of test decisions on which it acted on its learned values (mean $\pm$ sd over seeds: three seeds on the real benchmarks, each pooling the configurations and investors; 100 test seeds in PlantedMem and 50 for the $k$ designs). PlantedMem (core) averages the five core regimes, and PlantedMem (informative) is the regime with 48 experiences and informative content; -- marks settings that were not evaluated. A large sd on the real benchmarks means that the test passed mostly in one seed. Table~\ref{tab:ablation-abs} gives the utilities behind Figure~\ref{fig:ablation}. The variant whose test never passes always takes the anchored action: it is identical to \methodname on PortBench-Full, where the test of \methodname never passed either, and in PlantedMem it stays at the reference in every regime (Table~\ref{tab:regime}). On ClassAlloc it earns 0.629\% per month against 0.613\% for \methodname: acting on the trusted values cost 0.016~pp per month there ($p=0.28$; Appendix~E). Fixing the blend of the anchored action at its prior ($q=\alpha_0=0.9$, no online learning) and always keeping the reference ($q=1$) are two fixed alternatives to the learned blend: the first differs from \methodname by $-3.1$\% to $+0.9$\%, and the second is the extreme $\alpha_0=1$ of the prior (Appendix~\ref{app:anchor-sens}).

\begin{table*}[t]
  \caption{Trust rate (\%) of every variant. With the trust test, \methodname acts on its values on at most 0.3\% of PortBench and InvestorBench decisions (0.6\% on PortBench-Raw alone); without it, on 25--35\%.}
  \label{tab:trust-all}
  \footnotesize
  \setlength{\tabcolsep}{3pt}
  \begin{tabular}{@{}lccccc@{}}
    \toprule
    Variant & PortBench & InvestorBench & ClassAlloc & \makecell{PlantedMem\\(core)} & \makecell{PlantedMem\\(informative)} \\
    \midrule
    \multicolumn{6}{l}{\textit{Ablation (Figure~\ref{fig:ablation})}} \\
    \ours & \pmsd{0.3}{0.5} & \pmsd{0.3}{0.5} & \pmsd{14.6}{21.7} & \pmsd{33.7}{9.2} & \pmsd{44.9}{22.5} \\
    per-experience only (no content) & \pmsd{0.0}{0.0} & \pmsd{0.3}{0.5} & \pmsd{14.6}{21.7} & \pmsd{31.8}{8.9} & \pmsd{22.0}{20.7} \\
    content-aware only (no learner selection) & \pmsd{1.1}{1.3} & \pmsd{0.3}{0.5} & \pmsd{14.6}{21.7} & \pmsd{32.6}{9.0} & \pmsd{44.5}{23.1} \\
    in-sample F-test & \pmsd{0.0}{0.0} & \pmsd{0.0}{0.0} & \pmsd{2.3}{4.1} & \pmsd{29.3}{8.4} & \pmsd{15.1}{17.8} \\
    no trust test & \pmsd{24.7}{21.4} & \pmsd{34.5}{18.8} & \pmsd{53.8}{24.7} & \pmsd{54.3}{12.6} & \pmsd{64.9}{18.6} \\
    test never passes (always anchored action) & \pmsd{0.0}{0.0} & \pmsd{0.0}{0.0} & \pmsd{0.0}{0.0} & \pmsd{0.0}{0.0} & \pmsd{0.0}{0.0} \\
    draft average ($q=0$) & \pmsd{0.3}{0.5} & \pmsd{0.0}{0.0} & \pmsd{14.9}{20.5} & \pmsd{33.7}{9.2} & \pmsd{44.9}{22.5} \\
    always the reference ($q=1$) & \pmsd{0.3}{0.5} & \pmsd{0.3}{0.5} & \pmsd{14.4}{21.8} & \pmsd{33.7}{9.2} & \pmsd{44.9}{22.5} \\
    uniform prior ($\alpha_0=0.5$) & \pmsd{0.3}{0.5} & \pmsd{0.0}{0.0} & \pmsd{14.7}{20.6} & \pmsd{33.7}{9.2} & \pmsd{44.9}{22.5} \\
    no online learning ($q$ fixed at $\alpha_0$) & \pmsd{0.3}{0.5} & \pmsd{0.3}{0.5} & \pmsd{14.4}{21.8} & \pmsd{33.7}{9.2} & \pmsd{44.9}{22.5} \\
    \midrule
    \multicolumn{6}{l}{\textit{Trust test (Table~\ref{tab:trust-sens}); default $\alpha=0.05$, at least 10 forward scores}} \\
    $\alpha=0.01$ & \pmsd{0.0}{0.0} & \pmsd{0.0}{0.0} & -- & \pmsd{28.2}{8.1} & \pmsd{32.9}{22.3} \\
    $\alpha=0.02$ & \pmsd{0.0}{0.0} & \pmsd{0.0}{0.0} & -- & \pmsd{30.3}{8.5} & \pmsd{37.8}{22.7} \\
    $\alpha=0.1$ & \pmsd{5.3}{5.0} & \pmsd{0.5}{0.9} & -- & \pmsd{36.9}{10.0} & \pmsd{50.6}{21.4} \\
    $\alpha=0.2$ & \pmsd{12.2}{13.0} & \pmsd{1.6}{2.3} & -- & \pmsd{40.7}{11.0} & \pmsd{56.3}{20.4} \\
    at least 5 forward scores & \pmsd{0.3}{0.5} & \pmsd{0.3}{0.5} & -- & \pmsd{33.9}{9.2} & \pmsd{44.6}{22.2} \\
    at least 20 forward scores & \pmsd{0.0}{0.0} & \pmsd{0.0}{0.0} & -- & \pmsd{32.0}{8.5} & \pmsd{43.4}{21.6} \\
    \midrule
    \multicolumn{6}{l}{\textit{Number of retrieved experiences and design (Table~\ref{tab:k}); PlantedMem on seeds 15--64}} \\
    $k=2$, full factorial (4 drafts) & \pmsd{0.0}{0.0} & \pmsd{2.0}{2.0} & -- & \pmsd{8.5}{7.7} & \pmsd{7.2}{12.3} \\
    $k=3$, half fraction (4 drafts) & \pmsd{0.8}{1.4} & \pmsd{4.5}{5.2} & -- & \pmsd{18.5}{9.4} & \pmsd{9.5}{13.7} \\
    $k=3$, full factorial (8 drafts) & \pmsd{0.0}{0.0} & \pmsd{10.8}{10.7} & -- & \pmsd{29.4}{11.0} & \pmsd{43.4}{25.6} \\
    $k=4$, half fraction (8 drafts; seeds 15--64) & \pmsd{0.3}{0.5} & \pmsd{0.3}{0.5} & -- & \pmsd{35.4}{11.0} & \pmsd{49.6}{24.3} \\
    $k=4$, full factorial (16 drafts) & \pmsd{1.1}{1.9} & \pmsd{8.5}{7.4} & -- & \pmsd{42.0}{8.1} & \pmsd{69.9}{20.1} \\
    $k=5$, half fraction (16 drafts) & -- & -- & -- & \pmsd{43.4}{10.3} & \pmsd{74.7}{17.0} \\
    $k=6$, half fraction (32 drafts) & -- & -- & -- & \pmsd{49.1}{9.8} & \pmsd{86.4}{10.6} \\
    \bottomrule
  \end{tabular}
\end{table*}

\begin{table*}[t]
  \caption{Utilities behind Figure~\ref{fig:ablation}, with the variant that always keeps the reference ($q=1$; Appendix~\ref{app:anchor-sens}) (\% per month, bp per day on InvestorBench; mean $\pm$ sd over seeds). Without signal, the variants differ little in PlantedMem; with signal, never acting on the learned values stays at the reference.}
  \label{tab:ablation-abs}
  \footnotesize
  \setlength{\tabcolsep}{2.6pt}
  \begin{tabular}{@{}L{1.5cm}L{3.5cm}cccccccc@{}}
    \toprule
     & & & & & & \multicolumn{4}{c}{PlantedMem} \\
    \cmidrule(l){7-10}
    Step changed & Variant & PortBench-Full & PortBench-Raw & InvestorBench & ClassAlloc & no signal & signal & \makecell{informative\\content} & core \\
    \midrule
    -- & \ours & \pmsd{1.98}{0.00} & \second{\pmsd{2.01}{0.00}} & \second{\pmsd{10.40}{0.04}} & \pmsd{0.613}{0.017} & \pmsd{0.784}{0.031} & \pmsd{0.932}{0.070} & \pmsd{0.862}{0.068} & \second{\pmsd{0.873}{0.045}} \\
    \midrule
    \multirow{2}{=}{Value estimation} & per-experience learner only & \pmsd{1.98}{0.00} & \second{\pmsd{2.01}{0.00}} & \second{\pmsd{10.40}{0.04}} & \pmsd{0.613}{0.017} & \second{\pmsd{0.786}{0.030}} & \pmsd{0.929}{0.069} & \pmsd{0.817}{0.048} & \pmsd{0.872}{0.045} \\
     & content-aware learner only & \pmsd{1.98}{0.00} & \pmsd{1.94}{0.12} & \second{\pmsd{10.40}{0.04}} & \pmsd{0.613}{0.017} & \pmsd{0.785}{0.031} & \pmsd{0.929}{0.072} & \second{\pmsd{0.866}{0.072}} & \pmsd{0.871}{0.046} \\
    \midrule
    \multirow{3}{=}{Trust test} & in-sample F-test & \pmsd{1.98}{0.00} & \second{\pmsd{2.01}{0.00}} & \pmsd{10.37}{0.00} & \pmsd{0.628}{0.001} & \second{\pmsd{0.786}{0.029}} & \pmsd{0.924}{0.064} & \pmsd{0.814}{0.045} & \pmsd{0.869}{0.043} \\
     & none (always act on values) & \pmsd{1.89}{0.11} & \pmsd{1.77}{0.41} & \pmsd{9.91}{0.45} & \pmsd{0.612}{0.005} & \pmsd{0.767}{0.038} & \best{\pmsd{0.940}{0.080}} & \best{\pmsd{0.882}{0.068}} & \pmsd{0.871}{0.052} \\
     & never passes (always anchored action) & \pmsd{1.98}{0.00} & \second{\pmsd{2.01}{0.00}} & \pmsd{10.37}{0.00} & \pmsd{0.629}{0.001} & \best{\pmsd{0.787}{0.029}} & \pmsd{0.788}{0.029} & \pmsd{0.789}{0.029} & \pmsd{0.788}{0.028} \\
    \midrule
    \multirow{4}{=}{Anchored action} & draft average ($q=0$) & \pmsd{1.62}{0.05} & \pmsd{1.19}{0.02} & \pmsd{8.69}{0.09} & \best{\pmsd{0.648}{0.007}} & \pmsd{0.747}{0.032} & \pmsd{0.905}{0.080} & \pmsd{0.837}{0.073} & \pmsd{0.842}{0.049} \\
     & uniform prior ($\alpha_0=0.5$) & \pmsd{1.83}{0.02} & \pmsd{1.92}{0.01} & \pmsd{9.71}{0.02} & \second{\pmsd{0.632}{0.008}} & \pmsd{0.777}{0.024} & \pmsd{0.924}{0.072} & \pmsd{0.855}{0.066} & \pmsd{0.865}{0.044} \\
     & no online learning ($q$ fixed at $\alpha_0$) & \second{\pmsd{1.99}{0.01}} & \pmsd{1.94}{0.00} & \pmsd{10.39}{0.04} & \pmsd{0.613}{0.018} & \pmsd{0.784}{0.031} & \pmsd{0.931}{0.070} & \pmsd{0.862}{0.068} & \pmsd{0.872}{0.045} \\
     & always the reference ($q=1$) & \best{\pmsd{2.02}{0.00}} & \best{\pmsd{2.02}{0.00}} & \best{\pmsd{10.48}{0.05}} & \pmsd{0.605}{0.018} & \pmsd{0.785}{0.034} & \second{\pmsd{0.933}{0.070}} & \pmsd{0.864}{0.068} & \best{\pmsd{0.874}{0.045}} \\
    \bottomrule
  \end{tabular}
\end{table*}

\textbf{Why do some values coincide?} On PortBench and InvestorBench, the three variants that change how values are learned or tested give the same numbers as \methodname in all but two cells. Such a change can affect a decision only when the values are acted upon, and on these two benchmarks the test (or the F-test) passes on at most 1.1\% of decisions (Table~\ref{tab:trust-all}), so on all other decisions the four methods take the same anchored action. The two exceptions come from the few decisions on which the tests disagree: the content-aware learner alone passes its test by chance on a few PortBench-Raw decisions (1.94~$\pm$~0.12; observation 3), and the F-test never passes on InvestorBench, whereas \methodname's test passes on 0.3\% of decisions (10.37 against 10.40). On ClassAlloc, the value model estimates the content term at zero on every date, so the content-aware learner coincides with the per-experience learner, and the two value-estimation variants give the same numbers as \methodname although the test passes on 15\% of decisions. Where no test passes, the anchored action stays close to the reference, which does not depend on the seed, so the sd is below 0.005; a test that passes on a few decisions of one seed raises it (0.04 for \methodname on InvestorBench). The same steps make a difference in PlantedMem, where these variants trust their values on 15--45\% of decisions and the columns differ. The anchored-action variants change what the agent does when its values are not trusted, so they matter most on PortBench and InvestorBench, where the values are rarely trusted.

We make the following observations. (1)~\emph{The trust test matters most on real markets.} Without it, \methodname acts on its learned values whenever they favour a designed subset, on 25\% of PortBench and 34.5\% of InvestorBench test decisions, and loses 0.08~pp per month on PortBench-Full, 0.24~pp on PortBench-Raw and 0.49~bp per day on InvestorBench, with a much larger spread across seeds. In PlantedMem, removing the test leaves the core average unchanged and gains 0.02~pp in the informative regime, where acting early pays off; \S\ref{sec:analysis} shows that the test trades this small delay for protection when experiences carry no signal. On ClassAlloc, where the test passes on 15\% of decisions, acting on the values does not pay either way: without the test, \methodname loses 0.001~pp, and the F-test, which passes on 2\% of decisions, gains 0.016~pp. (2)~\emph{The test must look at future dates, not past ones.} An in-sample F-test of whether experiences differed in the past leaves PortBench and the core regimes nearly unchanged and never passes on InvestorBench ($-0.03$~bp per day), but it loses 0.05~pp in the informative regime: with 48 experiences, each retrieved only occasionally, the F-test, which needs repeated measurements of the same experience, passes on only 15\% of decisions there, against 45\% for the trust test. (3)~\emph{Content-aware value estimation helps when content is informative and costs nothing otherwise.} Without it, \methodname trusts its values on only 22\% of decisions in the informative regime and loses 0.05~pp. Using the content-aware learner alone, without learner selection, is as good there but costs 0.07~pp on PortBench-Raw, with a larger spread across seeds, because its chance forward scores pass the test more often (1.1\% against 0.3\% of PortBench decisions); selecting the learner by its forward score keeps the benefit and avoids the cost. (4)~\emph{When the values are not trusted, the agent must stay anchored to the reference.} Replacing the anchored blend by the draft average costs 0.36~pp (Full), 0.82~pp (Raw), 1.70~bp per day (InvestorBench) and 0.03~pp (PlantedMem, core), and learning the blend from a uniform prior ($\alpha_0=0.5$) costs 0.14~pp (Full), 0.09~pp (Raw) and 0.68~bp per day (InvestorBench). Always keeping the reference ($q=1$) is as good as \methodname in PlantedMem and slightly better on PortBench and InvestorBench (by up to 0.04~pp per month and 0.08~bp per day), because there the average draft never beats the reference; online learning matters when the reference is weak (Appendix~\ref{app:anchor-sens}). ClassAlloc is such a case: there the drafts beat 1/$N$, so the draft average would gain 0.035~pp per month and the uniform prior 0.019~pp, whereas always keeping the reference loses 0.008~pp.

\textbf{Baselines with the same anchor.} Table~\ref{tab:anchored} gives every LLM-based baseline the anchored action of \methodname: on every date, it deploys $q_t\,\xref_t+(1-q_t)\,x_t$, where $x_t$ is the portfolio that the baseline would deploy and $q_t$ follows Eq.~\eqref{eq:ftrl} on the baseline's own realized outcomes ($\alpha_0=0.9$, $\lambda_0=4$); holdings, fees and the timing of outcomes are those of each benchmark. For Draft averaging, this is exactly the variant whose test never passes. On PortBench and InvestorBench, the anchor helps every baseline, and all of them come within 1\% of \methodname; on ClassAlloc, where the trusted decisions of \methodname lost to the anchored blend, all of them are ahead of it (0.621--0.635\% against 0.613\% per month). No difference between \methodname and the best anchored baseline on a real benchmark is significant ($p\ge0.13$, two-sided paired $t$-test over test dates), and Hedge with a prior weight of 0.9 on the reference does as well. In PlantedMem, the anchor keeps the baselines close to the reference: with signal, they earn 0.787--0.838\% per month against 0.788\% for the reference, whereas \methodname, which acts fully on its values once they have predicted unseen dates, earns 0.932\% per month, 11\% more than the best of them (Counterfactual selection; $p<0.001$ over seeds), and 7.5\% more with informative content. Without signal, the anchored baselines stay at the reference and \methodname is 0.6\% below them, the cost of its false passes.

\begin{table*}[t]
  \caption{Every LLM-based baseline with the anchored action of \methodname (Eq.~\eqref{eq:ftrl}; \% per month, bp per day on InvestorBench; mean $\pm$ sd over seeds). With the same anchor, the baselines do as well as \methodname on the real benchmarks; with signal in PlantedMem, \methodname earns 11\% more than the best of them.}
  \label{tab:anchored}
  \footnotesize
  \setlength{\tabcolsep}{2.8pt}
  \begin{tabular}{@{}lccccccc@{}}
    \toprule
     & & & & & \multicolumn{3}{c}{PlantedMem} \\
    \cmidrule(l){6-8}
    Method & PortBench-Full & PortBench-Raw & InvestorBench & ClassAlloc & no signal & signal & \makecell{informative\\content} \\
    \midrule
    1/$N$ & \best{\pmsd{2.021}{0.000}} & \best{\pmsd{2.021}{0.000}} & \best{\pmsd{10.45}{0.00}} & \pmsd{0.623}{0.000} & \pmsd{0.575}{0.000} & \pmsd{0.575}{0.000} & \pmsd{0.575}{0.000} \\
    \ours & \pmsd{1.976}{0.005} & \pmsd{2.007}{0.002} & \pmsd{10.40}{0.04} & \pmsd{0.613}{0.017} & \pmsd{0.784}{0.031} & \best{\pmsd{0.932}{0.070}} & \best{\pmsd{0.862}{0.068}} \\
    Hedge, prior 0.9 on the reference & \second{\pmsd{1.992}{0.009}} & \pmsd{2.003}{0.003} & \second{\pmsd{10.41}{0.01}} & \pmsd{0.627}{0.001} & \pmsd{0.786}{0.029} & \pmsd{0.787}{0.029} & \pmsd{0.787}{0.030} \\
    \midrule
    \multicolumn{8}{@{}l}{\textit{With the anchored action of \methodname}} \\
    Memory-free & \pmsd{1.985}{0.008} & \pmsd{1.998}{0.004} & \pmsd{10.34}{0.02} & \second{\pmsd{0.630}{0.001}} & \best{\pmsd{0.789}{0.030}} & \pmsd{0.789}{0.032} & \pmsd{0.788}{0.030} \\
    Self-consistency & \pmsd{1.980}{0.022} & \pmsd{2.000}{0.005} & \pmsd{10.32}{0.01} & \best{\pmsd{0.635}{0.002}} & \second{\pmsd{0.788}{0.032}} & \pmsd{0.788}{0.032} & \pmsd{0.788}{0.032} \\
    Similarity retrieval (top-2) & \pmsd{1.961}{0.021} & \pmsd{2.003}{0.010} & \pmsd{10.36}{0.01} & \pmsd{0.629}{0.006} & \pmsd{0.787}{0.029} & \pmsd{0.788}{0.030} & \pmsd{0.790}{0.031} \\
    Similarity retrieval (top-4) & \pmsd{1.984}{0.011} & \pmsd{2.006}{0.003} & \pmsd{10.34}{0.03} & \pmsd{0.627}{0.006} & \pmsd{0.787}{0.029} & \pmsd{0.787}{0.028} & \pmsd{0.789}{0.031} \\
    FinMem & \pmsd{1.981}{0.041} & \pmsd{2.005}{0.001} & \pmsd{10.35}{0.00} & \pmsd{0.621}{0.002} & \pmsd{0.787}{0.030} & \pmsd{0.796}{0.030} & \pmsd{0.789}{0.031} \\
    MemRL & \pmsd{1.963}{0.044} & \pmsd{1.998}{0.007} & \pmsd{10.36}{0.03} & \pmsd{0.625}{0.002} & \pmsd{0.787}{0.030} & \pmsd{0.792}{0.031} & \pmsd{0.790}{0.030} \\
    Reflexion & \pmsd{1.964}{0.042} & \pmsd{1.999}{0.001} & \pmsd{10.30}{0.01} & \pmsd{0.621}{0.006} & \best{\pmsd{0.789}{0.029}} & \pmsd{0.790}{0.029} & \pmsd{0.790}{0.029} \\
    ExpeL & \pmsd{1.971}{0.015} & \pmsd{2.007}{0.003} & \pmsd{10.34}{0.01} & \pmsd{0.622}{0.001} & \pmsd{0.787}{0.029} & \pmsd{0.803}{0.032} & \pmsd{0.791}{0.029} \\
    Uplift credit & \pmsd{1.969}{0.012} & \pmsd{2.004}{0.003} & \pmsd{10.35}{0.01} & \pmsd{0.628}{0.003} & \pmsd{0.787}{0.030} & \pmsd{0.804}{0.034} & \pmsd{0.792}{0.031} \\
    Counterfactual selection & \pmsd{1.970}{0.026} & \pmsd{2.006}{0.009} & \pmsd{10.34}{0.01} & \pmsd{0.629}{0.005} & \second{\pmsd{0.788}{0.031}} & \second{\pmsd{0.838}{0.040}} & \second{\pmsd{0.802}{0.032}} \\
    Draft averaging (= test never passes) & \pmsd{1.976}{0.005} & \pmsd{2.007}{0.001} & \pmsd{10.37}{0.00} & \pmsd{0.629}{0.001} & \pmsd{0.787}{0.029} & \pmsd{0.788}{0.029} & \pmsd{0.789}{0.029} \\
    Hedge & \pmsd{1.984}{0.006} & \second{\pmsd{2.013}{0.000}} & \pmsd{10.39}{0.01} & \pmsd{0.626}{0.001} & \pmsd{0.787}{0.030} & \pmsd{0.788}{0.030} & \pmsd{0.788}{0.030} \\
    \bottomrule
  \end{tabular}
\end{table*}

\subsection{Wealth over time, market phases and survival}\label{app:wealth}

\textbf{Wealth.} For every method, seed and setting, the wealth path starts at 1 and follows the net returns after the 15~bp trading fee, before the risk penalty of the utility; we divide it by the wealth path of 1/$N$ in the same setting. On InvestorBench and ClassAlloc, which carry the holdings from one decision to the next, it is the value of the account from each decision date to the next (on ClassAlloc, a decision is scored over the 20 sessions after it and held until the next decision date). On PortBench, whose decisions each start from the equal-weight portfolio, it compounds the net returns of the 20-day decision periods. The net returns are those of the portfolios deployed in the reported runs: evaluating these portfolios reproduces every reported utility exactly, so the wealth paths need no new LLM calls. \textbf{Market phases.} For each benchmark we form the equal-weighted index of its equities (PortBench: the equities in the date's universe; InvestorBench: the four stocks; ClassAlloc: the equities class) and call \emph{falling} every phase from a peak of this index to the following trough in which the index lost at least 5\%; all other periods are \emph{rising}. This gives one falling phase on PortBench (August to October 2023, $-8.7\%$), one on InvestorBench (October 2020, $-8.7\%$) and four on ClassAlloc, the largest of which is the 2022 bear market ($-20.6\%$). \textbf{Survival.} A run survives at a given shortfall if its wealth never fell further behind the wealth of 1/$N$ than this shortfall at any time in the test period.

\begin{figure*}[t]
  \centering
  \includegraphics[width=\textwidth]{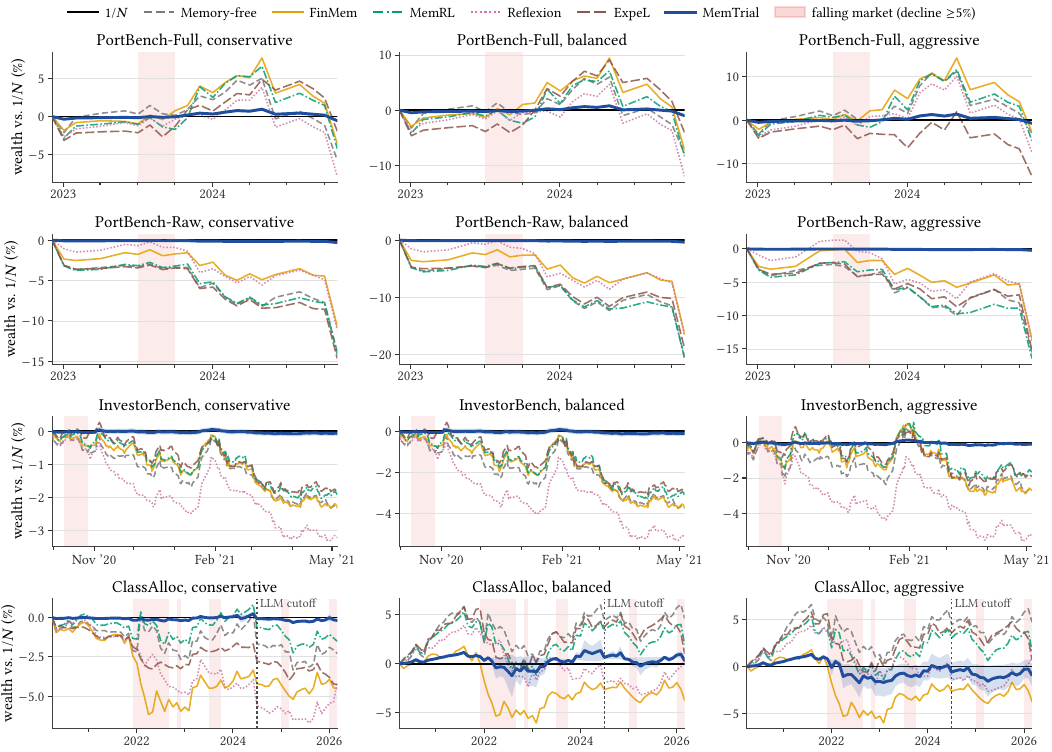}
  \caption{Wealth relative to 1/$N$ in every setting. \methodname stays within 2\% of 1/$N$ throughout, whereas the other agents fall up to 21\% behind.}
  \Description{Twelve line charts in four rows (PortBench-Full, PortBench-Raw, InvestorBench, ClassAlloc) and three columns (conservative, balanced and aggressive investor) showing wealth relative to 1/N in percent for 1/N, the memory-free LLM, FinMem, MemRL, Reflexion, ExpeL and MemTrial. MemTrial stays near zero; the other agents fall up to 21 percent behind.}
  \label{fig:wealth-rel-all}
\end{figure*}

\begin{figure*}[t]
  \centering
  \includegraphics[width=\textwidth]{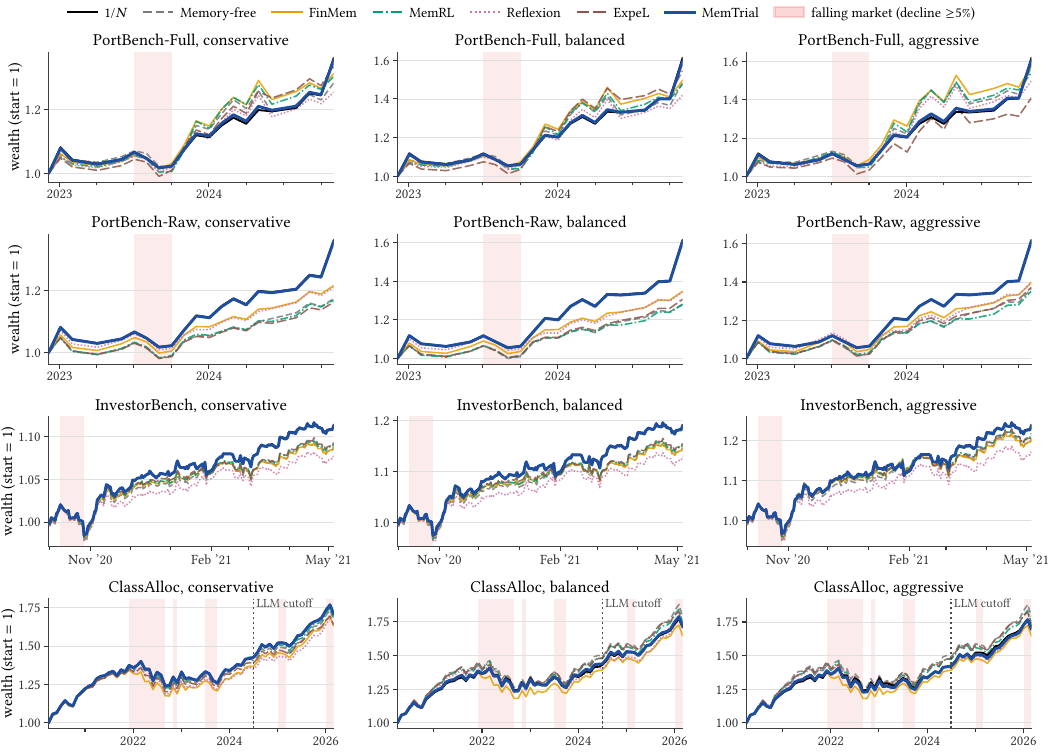}
  \caption{Wealth in every setting (start $=1$). \methodname tracks 1/$N$; on PortBench and InvestorBench, the others end below it.}
  \Description{Twelve line charts in the same grid showing wealth, starting at 1, for the same methods. MemTrial tracks 1/N; the other agents end below it on PortBench and InvestorBench.}
  \label{fig:wealth-raw}
\end{figure*}

\begin{figure*}[t]
  \centering
  \includegraphics[width=\textwidth]{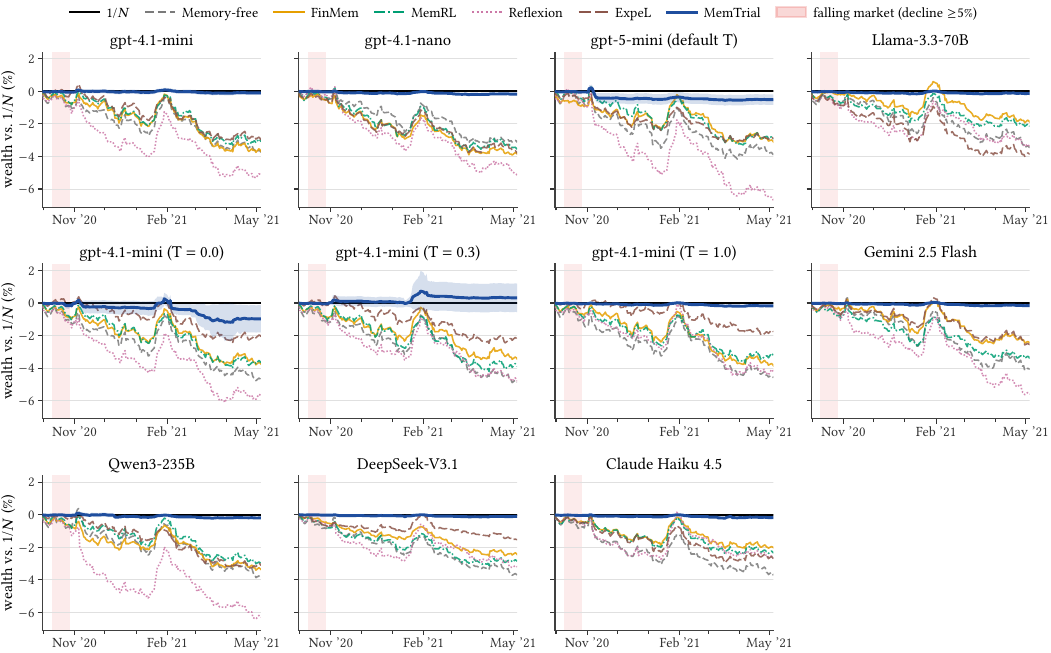}
  \caption{InvestorBench with other LLMs and temperatures: wealth relative to 1/$N$. \methodname stays within 1.2\% of 1/$N$ in every configuration; the other agents end 1.5--6.7\% behind.}
  \Description{Eleven line charts, one per InvestorBench configuration (eight LLMs and three further temperatures), showing wealth relative to 1/N for the balanced investor. MemTrial stays near zero; the other agents end 1.5 to 6.7 percent behind.}
  \label{fig:wealth-llms}
\end{figure*}

Figure~\ref{fig:wealth-rel-all} shows the wealth relative to 1/$N$ in every setting, Figure~\ref{fig:wealth-raw} the wealth itself, Figure~\ref{fig:wealth-llms} InvestorBench with the other LLMs and temperatures, and Figure~\ref{fig:wealth-cb-llms} ClassAlloc with the other LLMs. All use gpt-4.1-mini unless stated and show the mean over three seeds, with a band of $\pm$~sd for \methodname; shading marks the falling phases and the dotted line the LLM's knowledge cutoff, as in Figure~\ref{fig:wealth}. Figures~\ref{fig:wealth-llms} and~\ref{fig:wealth-cb-llms} show the balanced investor; on InvestorBench, the experience pool, retrieval and test days are those of the main run. Table~\ref{tab:wealth-phase} gives the relative wealth at the end of the test period and accumulated over the falling and the rising phases, Table~\ref{tab:survival} the share of runs whose wealth never fell further behind that of 1/$N$ than a given shortfall, with the median over runs of the largest shortfall, and Table~\ref{tab:risk} the cumulative net return, annualized Sharpe ratio and maximum drawdown over the test period, computed from the wealth paths of Figure~\ref{fig:wealth-raw}. In Tables~\ref{tab:wealth-phase} and~\ref{tab:risk}, each seed averages the settings of a benchmark (PortBench: two configurations and three investors; InvestorBench and ClassAlloc: eight LLMs and three investors); Table~\ref{tab:survival} pools the 54 runs of each seed over the three benchmarks. All three report the mean $\pm$ sd over the three seeds. On PortBench, the experience-learning agents were ahead of 1/$N$ in mid-2024 in the Full configuration; the last test month (November 2024) reverses this: the equal-weighted cryptocurrency class of the universe gained 129\%, and 1/$N$, which holds 10\% in it for the balanced investor, gained 15\%, whereas the agents' drafts held 0--7\%. Without this month, Similarity retrieval (top-4) (1.50\% per month) and Hedge (1.49\%) come out ahead of 1/$N$ and \methodname (1.46\% each) on PortBench-Full; dropping any other single test month leaves \methodname the best LLM-based method on both PortBench variants. On ClassAlloc, the memory-free LLM and most memory-based methods end above 1/$N$ for the balanced and aggressive investors, with their gains made before the knowledge cutoff of gpt-4.1-mini; after it, every LLM-based method trails 1/$N$, \methodname the least for every investor.

\textbf{Summary over all settings.} Every method gains over the test periods (for the balanced investor, 1/$N$ ends at 1.61, 1.19 and 1.82 times its initial wealth on PortBench-Full, InvestorBench and ClassAlloc), so we compare the methods by their wealth relative to 1/$N$. Averaged over all settings, including every LLM, the experience-learning agents fall behind 1/$N$ on PortBench and InvestorBench mostly while markets rise (by 9.8--11.5\% and 2.0--4.1\% over the rising phases), and on ClassAlloc they lose 1.5--3.5\% against 1/$N$ over the falling phases, which include the 2022 bear market (Table~\ref{tab:wealth-phase}). Over the falling phases of InvestorBench and ClassAlloc, \methodname loses less than the memory-free LLM and every experience-learning agent (0.06\% and 0.9\%, against 0.38--0.65\% and 1.5--3.5\%). Its terminal wealth is the highest of all LLM-based methods in 38 of the 39 PortBench and InvestorBench settings and within 1.4\% of 1/$N$, and its Sharpe ratio and maximum drawdown are the closest to those of 1/$N$ (Table~\ref{tab:risk}). On ClassAlloc, the memory-free LLM ends 0.3\% above 1/$N$.

\begin{table*}[t]
  \caption{Wealth relative to 1/$N$ (\%) at the end and by market phase. \methodname stays within 1\% of 1/$N$ in every phase; the other agents lose most in the rising phases of PortBench and InvestorBench.}
  \label{tab:wealth-phase}
  \footnotesize
  \setlength{\tabcolsep}{2.5pt}
  \begin{tabular}{@{}lccccccccc@{}}
    \toprule
    Method & \multicolumn{3}{c}{PortBench} & \multicolumn{3}{c}{InvestorBench} & \multicolumn{3}{c}{ClassAlloc} \\
    \cmidrule(lr){2-4}\cmidrule(lr){5-7}\cmidrule(lr){8-10}
     & End & Falling & Rising & End & Falling & Rising & End & Falling & Rising \\
    \midrule
    Memory-free & \pmsd{$-11.10$}{0.46} & \pmsd{$-0.65$}{0.58} & \pmsd{$-10.50$}{0.61} & \pmsd{$-3.15$}{0.36} & \pmsd{$-0.65$}{0.06} & \pmsd{$-2.52$}{0.39} & \best{\pmsd{$+0.31$}{0.97}} & \pmsd{$-1.59$}{0.25} & \best{\pmsd{$+2.08$}{0.89}} \\
    FinMem & \second{\pmsd{$-9.05$}{1.93}} & \best{\pmsd{$+0.75$}{1.48}} & \second{\pmsd{$-9.80$}{0.96}} & \pmsd{$-2.54$}{0.20} & \pmsd{$-0.41$}{0.03} & \pmsd{$-2.14$}{0.22} & \pmsd{$-4.62$}{0.25} & \pmsd{$-3.17$}{0.10} & \pmsd{$-1.33$}{0.35} \\
    MemRL & \pmsd{$-11.44$}{2.82} & \pmsd{$-0.93$}{0.38} & \pmsd{$-10.59$}{2.64} & \second{\pmsd{$-2.46$}{0.10}} & \pmsd{$-0.50$}{0.11} & \second{\pmsd{$-1.97$}{0.11}} & \pmsd{$-2.50$}{0.22} & \pmsd{$-3.53$}{0.20} & \second{\pmsd{$+1.35$}{0.07}} \\
    Reflexion & \pmsd{$-11.42$}{1.16} & \second{\pmsd{$+0.11$}{1.31}} & \pmsd{$-11.50$}{1.78} & \pmsd{$-4.60$}{0.10} & \pmsd{$-0.52$}{0.06} & \pmsd{$-4.09$}{0.11} & \pmsd{$-2.53$}{0.87} & \second{\pmsd{$-1.45$}{0.67}} & \pmsd{$-0.98$}{0.71} \\
    ExpeL & \pmsd{$-11.29$}{0.88} & \pmsd{$-0.05$}{1.03} & \pmsd{$-11.27$}{0.83} & \pmsd{$-2.75$}{0.16} & \second{\pmsd{$-0.38$}{0.03}} & \pmsd{$-2.37$}{0.18} & \pmsd{$-3.88$}{0.39} & \pmsd{$-3.44$}{0.26} & \pmsd{$+0.17$}{0.33} \\
    \ours & \best{\pmsd{$-0.53$}{0.04}} & \pmsd{$+0.03$}{0.03} & \best{\pmsd{$-0.55$}{0.03}} & \best{\pmsd{$-0.21$}{0.06}} & \best{\pmsd{$-0.06$}{0.02}} & \best{\pmsd{$-0.15$}{0.04}} & \second{\pmsd{$-0.33$}{0.30}} & \best{\pmsd{$-0.94$}{0.15}} & \pmsd{$+0.64$}{0.18} \\
    \midrule
    Falling / rising periods & \multicolumn{3}{c}{3 / 17 months} & \multicolumn{3}{c}{12 / 137 days} & \multicolumn{3}{c}{17 / 54 months} \\
    \bottomrule
  \end{tabular}
\end{table*}

\begin{table}[t]
  \caption{Survival against 1/$N$ (\% of runs). \methodname never falls 5\% behind 1/$N$; the others do in 22--44\% of runs.}
  \label{tab:survival}
  \footnotesize
  \setlength{\tabcolsep}{2pt}
  \begin{tabular}{@{}lcccc@{}}
    \toprule
    Method & \makecell{Shortfall\\$<2\%$} & \makecell{Shortfall\\$<5\%$} & \makecell{Shortfall\\$<10\%$} & \makecell{Median largest\\shortfall (\%)} \\
    \midrule
    Memory-free & \pmsd{27.8}{3.7} & \second{\pmsd{77.8}{3.7}} & \second{\pmsd{94.4}{0.0}} & \pmsd{3.15}{0.19} \\
    FinMem & \pmsd{19.1}{3.9} & \pmsd{67.9}{3.9} & \pmsd{90.7}{3.2} & \pmsd{3.59}{0.12} \\
    MemRL & \second{\pmsd{31.5}{4.9}} & \pmsd{74.7}{1.1} & \pmsd{93.2}{1.1} & \pmsd{3.04}{0.26} \\
    Reflexion & \pmsd{14.2}{4.3} & \pmsd{55.6}{1.9} & \pmsd{88.9}{3.7} & \pmsd{4.60}{0.24} \\
    ExpeL & \pmsd{26.5}{2.1} & \pmsd{69.8}{3.9} & \pmsd{83.3}{3.2} & \second{\pmsd{3.01}{0.36}} \\
    \ours & \best{\pmsd{93.2}{1.1}} & \best{\pmsd{100.0}{0.0}} & \best{\pmsd{100.0}{0.0}} & \best{\pmsd{0.20}{0.01}} \\
    \bottomrule
  \end{tabular}
\end{table}

\begin{table*}[t]
  \caption{Cumulative net return (CR, \%), annualized Sharpe ratio (SR) and maximum drawdown (MDD, \%). On every benchmark, \methodname has the highest Sharpe ratio and the smallest drawdown of the LLM-based agents.}
  \label{tab:risk}
  \footnotesize
  \setlength{\tabcolsep}{2.5pt}
  \begin{tabular}{@{}lccccccccc@{}}
    \toprule
    Method & \multicolumn{3}{c}{PortBench} & \multicolumn{3}{c}{InvestorBench} & \multicolumn{3}{c}{ClassAlloc} \\
    \cmidrule(lr){2-4}\cmidrule(lr){5-7}\cmidrule(lr){8-10}
     & CR & SR & MDD & CR & SR & MDD & CR & SR & MDD \\
    \midrule
    1/$N$ & \best{\pmsd{53.1}{0.0}} & \best{\pmsd{1.72}{0.00}} & \best{\pmsd{5.7}{0.0}} & \best{\pmsd{18.1}{0.0}} & \best{\pmsd{2.87}{0.00}} & \best{\pmsd{5.4}{0.0}} & \second{\pmsd{82.0}{0.0}} & \best{\pmsd{1.22}{0.00}} & \best{\pmsd{12.0}{0.0}} \\
    Memory-free & \pmsd{36.0}{0.8} & \pmsd{1.43}{0.04} & \pmsd{6.8}{0.8} & \pmsd{14.4}{0.4} & \pmsd{2.26}{0.06} & \pmsd{6.0}{0.1} & \best{\pmsd{82.5}{1.8}} & \pmsd{1.14}{0.02} & \pmsd{12.7}{0.3} \\
    FinMem & \pmsd{39.1}{3.0} & \second{\pmsd{1.53}{0.08}} & \second{\pmsd{6.2}{0.6}} & \pmsd{15.1}{0.2} & \pmsd{2.42}{0.03} & \pmsd{5.8}{0.0} & \pmsd{73.5}{0.5} & \pmsd{1.12}{0.00} & \pmsd{13.4}{0.0} \\
    MemRL & \pmsd{35.4}{4.4} & \pmsd{1.39}{0.05} & \pmsd{6.5}{0.4} & \pmsd{15.2}{0.1} & \pmsd{2.39}{0.03} & \pmsd{5.9}{0.1} & \pmsd{77.4}{0.4} & \pmsd{1.11}{0.00} & \pmsd{13.5}{0.3} \\
    Reflexion & \pmsd{35.4}{1.8} & \pmsd{1.43}{0.09} & \pmsd{6.8}{0.8} & \pmsd{12.6}{0.1} & \pmsd{2.02}{0.01} & \pmsd{5.9}{0.0} & \pmsd{77.4}{1.6} & \pmsd{1.14}{0.01} & \pmsd{13.0}{0.4} \\
    ExpeL & \pmsd{35.5}{1.2} & \pmsd{1.47}{0.11} & \pmsd{6.7}{0.7} & \pmsd{14.8}{0.2} & \pmsd{2.41}{0.03} & \pmsd{5.8}{0.0} & \pmsd{74.9}{0.7} & \pmsd{1.11}{0.01} & \pmsd{13.4}{0.3} \\
    \ours & \second{\pmsd{52.3}{0.1}} & \best{\pmsd{1.72}{0.01}} & \best{\pmsd{5.7}{0.0}} & \second{\pmsd{17.9}{0.1}} & \second{\pmsd{2.84}{0.01}} & \second{\pmsd{5.5}{0.0}} & \pmsd{81.4}{0.5} & \second{\pmsd{1.19}{0.01}} & \second{\pmsd{12.3}{0.2}} \\
    \bottomrule
  \end{tabular}
\end{table*}

\subsection{Hyperparameter analysis}\label{app:hyper}
\begin{figure}[t]
  \centering
  \includegraphics[width=\columnwidth]{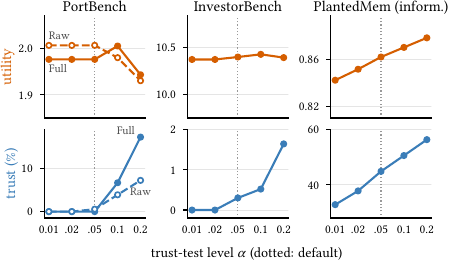}
  \caption{Effect of the trust-test level $\alpha$ (top: utility; bottom: trust rate). On real markets, a looser test acts on chance and loses; in PlantedMem, it acts earlier on values that hold.}
  \Description{Six small line charts in two rows for PortBench (Full and Raw), InvestorBench and PlantedMem (informative regime), with the trust-test level alpha at 0.01, 0.02, 0.05 (default), 0.1 and 0.2. Top row, utility: PortBench is flat up to 0.05 and falls by about 2 to 4 percent at 0.2; InvestorBench is flat; PlantedMem rises from 0.84 to 0.88 percent per month. Bottom row, trust rate: PortBench rises from 0 to 17 percent (Full) and 7 percent (Raw) at 0.2; InvestorBench stays below 2 percent; PlantedMem rises from 33 to 56 percent.}
  \label{fig:sens}
\end{figure}

We vary one hyperparameter at a time around the frozen setting: the trust test (Figure~\ref{fig:sens} and Table~\ref{tab:trust-sens}), the number of retrieved experiences $k$ and the design (Table~\ref{tab:k}), and the strength of the anchor (Appendix~\ref{app:anchor-sens}, Table~\ref{tab:anchor}). Tables report the mean $\pm$ sd over seeds, in \% per month (bp per day on InvestorBench). In Table~\ref{tab:k}, \emph{Drafts} is the number of drafts per decision; a draft is one LLM call on InvestorBench and one run of the three-call agent on PortBench, so the LLM cost grows in proportion. On the real benchmarks, drafts were generated for all 16 subsets of the four retrieved experiences, so every design with $k\le4$ is evaluated offline from these drafts, without new LLM calls. In PlantedMem, drafts are resampled, which also allows $k=5$ and 6; these results come from a separate run on 50 test seeds (15--64) with common random numbers across designs, in which the default scores 0.868\% per month (0.87 on 100 seeds in Table~\ref{tab:main}); they are shown with three decimals. Trust rates, with sd, are in Table~\ref{tab:trust-all}.

We make the following observations. (1)~\emph{The test level trades protection against speed.} Levels up to 0.05 give identical results on PortBench, because at these levels the test passes on at most 0.3\% of PortBench decisions. At $\alpha=0.1$ and 0.2, chance forward scores pass the test on 5\% and 12\% of PortBench decisions; $\alpha=0.1$ gains 0.03~pp on PortBench-Full but costs 0.03~pp on PortBench-Raw, and $\alpha=0.2$ costs 0.03 and 0.08~pp, with a larger spread across seeds. In PlantedMem, a looser test trusts the values earlier and more often (on 28\% of core decisions at $\alpha=0.01$ against 41\% at $\alpha=0.2$), which helps in the informative regime (0.84\% against 0.88\%). On InvestorBench, all levels stay within 0.05~bp per day of each other. The frozen level $\alpha=0.05$ sits between these regimes. (2)~\emph{The minimum number of forward scores} (5, 10 or 20) changes the InvestorBench result by at most 0.03~bp per day and no other result at two decimals. (3)~\emph{More retrieved experiences help when experiences matter, with diminishing returns per draft.} In PlantedMem, the core utility rises from 0.812\% per month with $k=2$ (4 drafts) to 0.868 with $k=4$ (8 drafts), 0.896 with $k=5$ (16 drafts) and 0.909 with $k=6$ (32 drafts): going from 8 to 16 drafts adds 0.03~pp, and doubling again to 32 drafts adds 0.01~pp. (4)~\emph{At equal cost, the half fraction is the better design}, because it measures one more experience with the same number of drafts. With 4 drafts, $k=3$ in a half fraction beats $k=2$ in a full factorial (0.829\% against 0.812\%); with 8 drafts, $k=4$ in a half fraction matches $k=3$ in a full factorial (0.868\% against 0.865\%) and trusts its values more often (35\% against 29\%); with 16 drafts, $k=5$ in a half fraction beats $k=4$ in a full factorial (0.896\% against 0.889\%). (5)~\emph{On the real benchmarks}, where the values are rarely trusted, $k$ and the design barely matter on PortBench (1.95--2.01\% per month). On InvestorBench, every other design passes the test more often (on 2--11\% of decisions), on values that do not carry over to the next day, and loses 0.13--0.44~bp per day against the default; the full factorial at $k=4$, for example, doubles the drafts and loses 0.41~bp per day. We therefore keep $k=4$ with eight drafts: in PlantedMem its gain over the reference per draft (0.011~pp) is close to the largest (0.012~pp, for $k=3$ in a half fraction), and on InvestorBench it is the best design. A larger $k$ is worth its cost only when experiences are known to matter. (6)~\emph{The anchor matters more than its exact strength} (Table~\ref{tab:anchor}). A uniform prior ($\alpha_0=0.5$) costs up to 0.18~pp on PortBench-Full, up to 0.29~pp on PortBench-Raw and 0.40--0.74~bp per day on InvestorBench, whereas with $\alpha_0\in\{0.9,0.95\}$ every $\lambda_0$ stays within 0.06~pp per month and 0.06~bp per day of the default; in PlantedMem, all twelve settings give 0.86--0.87\% per month on the core regimes.

\begin{table*}[t]
  \caption{Sensitivity to the trust test. Test levels up to 0.05 give identical PortBench results; looser levels help consistently only in the informative PlantedMem regime.}
  \label{tab:trust-sens}
  \small
  \begin{tabular}{llccccc}
    \toprule
    Hyperparameter & Value & PortBench-Full & PortBench-Raw & InvestorBench & \makecell{PlantedMem\\(core)} & \makecell{PlantedMem\\(informative)} \\
    \midrule
    \multirow{5}{*}{Test level $\alpha$} & 0.01 & \second{\pmsd{1.98}{0.00}} & \best{\pmsd{2.01}{0.00}} & \pmsd{10.37}{0.00} & \best{\pmsd{0.87}{0.04}} & \pmsd{0.84}{0.07} \\
      & 0.02 & \second{\pmsd{1.98}{0.00}} & \best{\pmsd{2.01}{0.00}} & \pmsd{10.37}{0.00} & \best{\pmsd{0.87}{0.04}} & \pmsd{0.85}{0.07} \\
      & 0.05 (default) & \second{\pmsd{1.98}{0.00}} & \best{\pmsd{2.01}{0.00}} & \second{\pmsd{10.40}{0.04}} & \best{\pmsd{0.87}{0.04}} & \pmsd{0.86}{0.07} \\
      & 0.1 & \best{\pmsd{2.01}{0.05}} & \second{\pmsd{1.98}{0.04}} & \best{\pmsd{10.42}{0.09}} & \best{\pmsd{0.87}{0.05}} & \second{\pmsd{0.87}{0.07}} \\
      & 0.2 & \pmsd{1.94}{0.05} & \pmsd{1.93}{0.13} & \pmsd{10.39}{0.07} & \best{\pmsd{0.87}{0.05}} & \best{\pmsd{0.88}{0.07}} \\
    \midrule
    \multirow{3}{*}{Minimum forward scores} & 5 & \best{\pmsd{1.98}{0.00}} & \best{\pmsd{2.01}{0.00}} & \best{\pmsd{10.40}{0.04}} & \best{\pmsd{0.87}{0.05}} & \best{\pmsd{0.86}{0.07}} \\
      & 10 (default) & \best{\pmsd{1.98}{0.00}} & \best{\pmsd{2.01}{0.00}} & \best{\pmsd{10.40}{0.04}} & \best{\pmsd{0.87}{0.04}} & \best{\pmsd{0.86}{0.07}} \\
      & 20 & \best{\pmsd{1.98}{0.00}} & \best{\pmsd{2.01}{0.00}} & \pmsd{10.37}{0.00} & \best{\pmsd{0.87}{0.04}} & \best{\pmsd{0.86}{0.06}} \\
    \bottomrule
  \end{tabular}
\end{table*}

\begin{table*}[t]
  \caption{Number of retrieved experiences $k$ and the design. More experiences help when experiences matter, with diminishing returns per draft; at equal cost, the half fraction is the better design.}
  \label{tab:k}
  \small
  \setlength{\tabcolsep}{4.5pt}
  \begin{tabular}{clcccccc}
    \toprule
    $k$ & Design & Drafts & PortBench-Full & PortBench-Raw & InvestorBench & \makecell{PlantedMem\\(core)} & \makecell{PlantedMem\\(informative)} \\
    \midrule
    2 & full factorial & 4 & \best{\pmsd{1.98}{0.01}} & \best{\pmsd{2.01}{0.00}} & \second{\pmsd{10.27}{0.11}} & \pmsd{0.812}{0.043} & \pmsd{0.798}{0.042} \\
    3 & half fraction & 4 & \pmsd{1.97}{0.02} & \best{\pmsd{2.01}{0.00}} & \pmsd{10.14}{0.21} & \pmsd{0.829}{0.041} & \pmsd{0.799}{0.047} \\
    3 & full factorial & 8 & \best{\pmsd{1.98}{0.01}} & \best{\pmsd{2.01}{0.00}} & \pmsd{9.96}{0.36} & \pmsd{0.865}{0.051} & \pmsd{0.850}{0.070} \\
    4 & half fraction (default) & 8 & \best{\pmsd{1.98}{0.00}} & \best{\pmsd{2.01}{0.00}} & \best{\pmsd{10.40}{0.04}} & \pmsd{0.868}{0.045} & \pmsd{0.862}{0.072} \\
    4 & full factorial & 16 & \pmsd{1.97}{0.00} & \pmsd{1.95}{0.11} & \pmsd{9.99}{0.34} & \pmsd{0.889}{0.047} & \pmsd{0.903}{0.085} \\
    5 & half fraction & 16 & -- & -- & -- & \second{\pmsd{0.896}{0.044}} & \second{\pmsd{0.914}{0.078}} \\
    6 & half fraction & 32 & -- & -- & -- & \best{\pmsd{0.909}{0.048}} & \best{\pmsd{0.932}{0.070}} \\
    \bottomrule
  \end{tabular}
\end{table*}

\subsection{Strength of the anchor}\label{app:anchor-sens}
Table~\ref{tab:anchor} varies the prior weight $\alpha_0$ on the reference and the regularization strength $\lambda_0$ of Eq.~\eqref{eq:ftrl} (mean $\pm$ sd over seeds; \% per month, bp per day on InvestorBench; default $\alpha_0=0.9$ and $\lambda_0=4$). Its last row is the extreme $\alpha_0=1$, which always keeps the reference when the values are not trusted. It is the best choice where 1/$N$ cannot be beaten, 2.2\%, 0.7\% and 0.8\% better than \methodname on PortBench-Full, PortBench-Raw and InvestorBench, but it cannot move towards better drafts: it is 1.2\% worse on ClassAlloc and 4.4\% worse in PlantedMem when 1/$N$ is the reference (Table~\ref{tab:weakref}). The default $\alpha_0=0.9$ gives up at most 2.2\% where the reference is strong and keeps the ability to follow better drafts.

\begin{table}[t]
  \caption{Strength of the anchor ($\alpha_0=1$: always the reference). A uniform prior ($\alpha_0=0.5$) costs the most; with $\alpha_0\ge0.9$, every $\lambda_0$ stays close to the default.}
  \label{tab:anchor}
  \footnotesize
  \setlength{\tabcolsep}{2pt}
  \begin{tabular}{@{}cccccc@{}}
    \toprule
    $\alpha_0$ & $\lambda_0$ & \makecell{PortBench-\\Full} & \makecell{PortBench-\\Raw} & InvestorBench & \makecell{PlantedMem\\(core)} \\
    \midrule
    0.5 & 1 & \pmsd{1.80}{0.02} & \best{\pmsd{2.02}{0.00}} & \pmsd{9.99}{0.03} & \best{\pmsd{0.87}{0.04}} \\
    0.5 & 4 & \pmsd{1.83}{0.02} & \pmsd{1.92}{0.01} & \pmsd{9.71}{0.02} & \best{\pmsd{0.87}{0.04}} \\
    0.5 & 16 & \pmsd{1.85}{0.02} & \pmsd{1.72}{0.01} & \pmsd{9.65}{0.04} & \second{\pmsd{0.86}{0.04}} \\
    0.75 & 1 & \pmsd{1.86}{0.03} & \best{\pmsd{2.02}{0.00}} & \pmsd{10.29}{0.05} & \best{\pmsd{0.87}{0.04}} \\
    0.75 & 4 & \pmsd{1.92}{0.01} & \pmsd{1.98}{0.00} & \pmsd{10.16}{0.02} & \best{\pmsd{0.87}{0.04}} \\
    0.75 & 16 & \pmsd{1.94}{0.01} & \pmsd{1.89}{0.00} & \pmsd{10.12}{0.01} & \best{\pmsd{0.87}{0.04}} \\
    0.9 & 1 & \pmsd{1.92}{0.03} & \best{\pmsd{2.02}{0.00}} & \pmsd{10.42}{0.05} & \best{\pmsd{0.87}{0.04}} \\
    0.9 & 4 (default) & \pmsd{1.98}{0.00} & \second{\pmsd{2.01}{0.00}} & \pmsd{10.40}{0.04} & \best{\pmsd{0.87}{0.04}} \\
    0.9 & 16 & \pmsd{1.99}{0.01} & \pmsd{1.97}{0.00} & \pmsd{10.39}{0.04} & \best{\pmsd{0.87}{0.04}} \\
    0.95 & 1 & \pmsd{1.95}{0.02} & \best{\pmsd{2.02}{0.00}} & \pmsd{10.45}{0.05} & \best{\pmsd{0.87}{0.04}} \\
    0.95 & 4 & \pmsd{2.00}{0.00} & \second{\pmsd{2.01}{0.00}} & \pmsd{10.45}{0.04} & \best{\pmsd{0.87}{0.05}} \\
    0.95 & 16 & \second{\pmsd{2.01}{0.00}} & \pmsd{2.00}{0.00} & \second{\pmsd{10.46}{0.04}} & \best{\pmsd{0.87}{0.05}} \\
    1 & -- & \best{\pmsd{2.02}{0.00}} & \best{\pmsd{2.02}{0.00}} & \best{\pmsd{10.48}{0.05}} & \best{\pmsd{0.87}{0.05}} \\
    \bottomrule
  \end{tabular}
\end{table}

The best $\lambda_0$ differs between the two PortBench configurations (16 on Full, 1 on Raw), and the default is within 0.03~pp of the best setting on both. A higher prior ($\alpha_0=0.95$) would have done slightly better on the real benchmarks, where the reference is strong, but by the trade-off shown in \S\ref{sec:ablation} it would adapt more slowly when the reference is weak.

\subsection{Number of retrieved experiences by regime}\label{app:k-regime}
Table~\ref{tab:k-regime} breaks the PlantedMem part of Table~\ref{tab:k} down by regime (\% per month; mean $\pm$ sd over test seeds 15--64, each averaged over the three investors; trust rates in Table~\ref{tab:trust-all}). When experiences carry no signal (first two columns), every design stays within 0.01~pp of the reference, so a larger $k$ adds no risk there; its gains come entirely from the regimes in which experiences matter.

\begin{table*}[t]
  \caption{PlantedMem utility by regime for each $k$ and design (\% per month). A larger $k$ gains only where experiences matter.}
  \label{tab:k-regime}
  \footnotesize
  \setlength{\tabcolsep}{2.5pt}
  \begin{tabular}{lccccccccc}
    \toprule
    Design & Drafts & No influence & Noise only & $\beta=0.1$ & $\beta=0.25$ & $\beta=0.5$ & Many exp. & Informative & Core avg. \\
    \midrule
    Reference & 8 & \pmsd{0.782}{0.031} & \second{\pmsd{0.782}{0.031}} & \pmsd{0.782}{0.031} & \pmsd{0.782}{0.031} & \pmsd{0.782}{0.031} & \pmsd{0.782}{0.031} & \pmsd{0.782}{0.031} & \pmsd{0.782}{0.031} \\
    $k=2$, full factorial & 4 & \best{\pmsd{0.784}{0.029}} & \second{\pmsd{0.782}{0.038}} & \pmsd{0.785}{0.029} & \pmsd{0.804}{0.048} & \pmsd{0.904}{0.130} & \pmsd{0.782}{0.029} & \pmsd{0.798}{0.042} & \pmsd{0.812}{0.043} \\
    $k=3$, half fraction & 4 & \best{\pmsd{0.784}{0.032}} & \best{\pmsd{0.784}{0.042}} & \pmsd{0.787}{0.038} & \pmsd{0.821}{0.061} & \pmsd{0.970}{0.123} & \pmsd{0.789}{0.036} & \pmsd{0.799}{0.047} & \pmsd{0.829}{0.041} \\
    $k=3$, full factorial & 8 & \pmsd{0.782}{0.028} & \pmsd{0.778}{0.032} & \pmsd{0.784}{0.031} & \pmsd{0.904}{0.089} & \pmsd{1.078}{0.154} & \pmsd{0.812}{0.048} & \pmsd{0.850}{0.070} & \pmsd{0.865}{0.051} \\
    $k=4$, half fraction (default) & 8 & \second{\pmsd{0.783}{0.028}} & \pmsd{0.777}{0.037} & \pmsd{0.794}{0.045} & \pmsd{0.901}{0.084} & \pmsd{1.085}{0.114} & \pmsd{0.836}{0.079} & \pmsd{0.862}{0.072} & \pmsd{0.868}{0.045} \\
    $k=4$, full factorial & 16 & \pmsd{0.780}{0.030} & \pmsd{0.778}{0.029} & \pmsd{0.792}{0.049} & \pmsd{0.950}{0.080} & \pmsd{1.143}{0.134} & \pmsd{0.869}{0.094} & \pmsd{0.903}{0.085} & \pmsd{0.889}{0.047} \\
    $k=5$, half fraction & 16 & \pmsd{0.780}{0.032} & \pmsd{0.775}{0.034} & \second{\pmsd{0.800}{0.054}} & \second{\pmsd{0.956}{0.081}} & \second{\pmsd{1.170}{0.110}} & \second{\pmsd{0.873}{0.075}} & \second{\pmsd{0.914}{0.078}} & \second{\pmsd{0.896}{0.044}} \\
    $k=6$, half fraction & 32 & \second{\pmsd{0.783}{0.029}} & \pmsd{0.776}{0.034} & \best{\pmsd{0.822}{0.073}} & \best{\pmsd{0.981}{0.082}} & \best{\pmsd{1.184}{0.126}} & \best{\pmsd{0.921}{0.066}} & \best{\pmsd{0.932}{0.070}} & \best{\pmsd{0.909}{0.048}} \\
    \bottomrule
  \end{tabular}
\end{table*}

\subsection{PlantedMem by regime}\label{app:regime}
Tables~\ref{tab:trust-rate}--\ref{tab:weakref} break the PlantedMem results down by regime (mean $\pm$ sd over the 100 test seeds, each averaged over the three investors). The trust rate is the share of decisions on which a method acts on its learned values, and the reference is the average of eight memory-free drafts, except in Table~\ref{tab:weakref}, where it is 1/$N$.

\textbf{When does \methodname trust its values?} Table~\ref{tab:trust-rate} shows how often the trust test passes in each regime of PlantedMem. When experiences carry no signal, \methodname acts on its values on only 2--3\% of decisions, below the nominal 5\% level of a single test. The trust rate rises with the influence of experiences, from 14\% at $\beta=0.1$ to 64.5\% at $\beta=0.25$ and 85\% at $\beta=0.5$, and with 48 experiences it rises from 26\% to 45\% when content becomes informative, because the content-aware learner then predicts the contributions of rarely seen experiences. On PortBench, the test passed on 1 of 360 test decisions and on 14 of 1,008 decisions over all months, consistent with Table~\ref{tab:confound}, where the effect of experiences on PortBench decisions is rarely detectable. On InvestorBench, it passed on only 0.3\% of decisions, even though experiences change most daily decisions detectably: the values learned from past days do not carry over to the following days, and the mean forward score of the learned values, averaged over investors and seeds, is $-0.07$. The test thus responds to whether learned values transfer to unseen dates, not to whether experiences matter on a given date.

\textbf{Where does \methodname gain and where does it lose?} Table~\ref{tab:regime} breaks the PlantedMem results down by regime. When experiences carry no signal, acting on learned values is costly: without its trust test, \methodname falls 0.01--0.03~pp below the reference, and Counterfactual selection falls up to 0.13~pp below it, whereas the full \methodname stays at the reference. When experiences matter, \methodname gains up to 0.30~pp over the reference, but it gives up 0.01--0.03~pp to Counterfactual selection, because it waits for forward evidence before acting; the exception is the informative regime, where content-aware value estimation puts it 0.01~pp ahead. Averaged over the core regimes, protection outweighs delay ($+0.02$~pp, $p=0.001$; Table~\ref{tab:main}). Outcome credit (MemRL\dag) never gains from experiences, even at $\beta=0.5$, because its credit follows the market.

\textbf{What if the reference is weak?} On PortBench, InvestorBench and PlantedMem, the reference is never beaten by the average draft, so always keeping it ($q=1$) is at least as good as \methodname; on ClassAlloc, where the average draft beats 1/$N$ by 0.03~pp per month, it is 0.008~pp worse (Table~\ref{tab:ablation-abs}). To test the online learning of the anchored action, we replace the reference in PlantedMem with the 1/$N$ portfolio, which ignores the investor's risk aversion and is about 0.18~pp per month worse than the average draft there. Table~\ref{tab:weakref} shows that, if \methodname always kept 1/$N$ when its values are not trusted, it would earn 0.71\% per month on the core regimes. The anchored action instead learns to move towards the drafts (0.75\%), and a uniform prior moves faster (0.81\%). In every regime, the learned blend is at least as good as always keeping 1/$N$. Never acting on its values, \methodname would earn only 0.62\%, so with a weak reference the learned values matter even more. The prior weight $\alpha_0$ therefore sets a trade-off: a high prior protects when the reference is strong (Table~\ref{tab:anchor}), and a low prior adapts faster when it is weak.

\begin{table*}[t]
  \caption{Trust rate (\%) by PlantedMem regime. \methodname rarely trusts its values when experiences carry no signal and trusts them more often the more experiences matter.}
  \label{tab:trust-rate}
  \footnotesize
  \setlength{\tabcolsep}{3pt}
  \begin{tabular}{lccccccc}
    \toprule
    Trust rule & No influence & Noise only & $\beta=0.1$ & $\beta=0.25$ & $\beta=0.5$ & Many exp. & Informative \\
    \midrule
    \methodname (trust test) & \pmsd{2.3}{6.6} & \pmsd{3.0}{10.1} & \pmsd{13.7}{20.7} & \pmsd{64.5}{23.3} & \pmsd{84.8}{8.9} & \pmsd{26.1}{22.6} & \pmsd{44.9}{22.5} \\
    Per-experience learner only & \pmsd{1.5}{5.1} & \pmsd{2.0}{8.3} & \pmsd{10.5}{18.0} & \pmsd{61.4}{24.0} & \pmsd{83.5}{10.1} & \pmsd{22.0}{20.7} & \pmsd{22.0}{20.7} \\
    In-sample F-test & \pmsd{2.4}{6.7} & \pmsd{2.1}{7.1} & \pmsd{8.1}{13.7} & \pmsd{52.4}{22.9} & \pmsd{81.7}{10.5} & \pmsd{14.8}{17.7} & \pmsd{15.1}{17.8} \\
    \bottomrule
  \end{tabular}
\end{table*}

\begin{table*}[t]
  \caption{Utility by PlantedMem regime (\% per month). \methodname stays at the reference when experiences carry no signal and gains when they matter.}
  \label{tab:regime}
  \footnotesize
  \setlength{\tabcolsep}{3pt}
  \begin{tabular}{lccccccc}
    \toprule
    Method & No influence & Noise only & $\beta=0.1$ & $\beta=0.25$ & $\beta=0.5$ & Many exp. & Informative \\
    \midrule
    Reference (memory-free average) & \best{\pmsd{0.788}{0.032}} & \best{\pmsd{0.788}{0.032}} & \pmsd{0.788}{0.032} & \pmsd{0.788}{0.032} & \pmsd{0.788}{0.032} & \pmsd{0.788}{0.032} & \pmsd{0.788}{0.032} \\
    Outcome credit (MemRL\dag) & \pmsd{0.759}{0.097} & \pmsd{0.658}{0.089} & \pmsd{0.758}{0.090} & \pmsd{0.755}{0.098} & \pmsd{0.749}{0.144} & \pmsd{0.760}{0.085} & \pmsd{0.760}{0.085} \\
    Counterfactual selection & \pmsd{0.768}{0.097} & \pmsd{0.659}{0.087} & \best{\pmsd{0.805}{0.093}} & \best{\pmsd{0.926}{0.100}} & \best{\pmsd{1.101}{0.138}} & \best{\pmsd{0.852}{0.078}} & \pmsd{0.852}{0.078} \\
    \methodname w/o trust test & \pmsd{0.776}{0.047} & \pmsd{0.758}{0.056} & \second{\pmsd{0.803}{0.059}} & \second{\pmsd{0.924}{0.092}} & \second{\pmsd{1.094}{0.131}} & \second{\pmsd{0.843}{0.061}} & \best{\pmsd{0.882}{0.068}} \\
    \methodname, never acting on its values & \best{\pmsd{0.788}{0.028}} & \second{\pmsd{0.787}{0.029}} & \pmsd{0.788}{0.028} & \pmsd{0.788}{0.029} & \pmsd{0.788}{0.029} & \pmsd{0.789}{0.029} & \pmsd{0.789}{0.029} \\
    \ours & \second{\pmsd{0.787}{0.030}} & \pmsd{0.782}{0.035} & \pmsd{0.793}{0.043} & \pmsd{0.912}{0.088} & \pmsd{1.090}{0.124} & \pmsd{0.821}{0.050} & \second{\pmsd{0.862}{0.068}} \\
    \bottomrule
  \end{tabular}
\end{table*}

\begin{table*}[t]
  \caption{PlantedMem with 1/$N$ as the reference (\% per month). The anchored blend learns to leave a weak reference.}
  \label{tab:weakref}
  \footnotesize
  \setlength{\tabcolsep}{3pt}
  \begin{tabular}{lcccccccc}
    \toprule
    Method & No influence & Noise only & $\beta=0.1$ & $\beta=0.25$ & $\beta=0.5$ & Many exp. & Informative & Core avg. \\
    \midrule
    1/$N$ & \pmsd{0.575}{0.000} & \pmsd{0.575}{0.000} & \pmsd{0.575}{0.000} & \pmsd{0.575}{0.000} & \pmsd{0.575}{0.000} & \pmsd{0.575}{0.000} & \pmsd{0.575}{0.000} & \pmsd{0.57}{0.00} \\
    Draft averaging & \best{\pmsd{0.782}{0.032}} & \best{\pmsd{0.715}{0.052}} & \best{\pmsd{0.777}{0.031}} & \pmsd{0.764}{0.044} & \pmsd{0.719}{0.080} & \best{\pmsd{0.775}{0.041}} & \second{\pmsd{0.775}{0.041}} & \second{\pmsd{0.75}{0.03}} \\
    \methodname, always 1/$N$ when not trusted ($q=1$) & \pmsd{0.578}{0.016} & \pmsd{0.580}{0.021} & \pmsd{0.596}{0.055} & \pmsd{0.783}{0.124} & \pmsd{1.030}{0.136} & \pmsd{0.626}{0.073} & \pmsd{0.700}{0.096} & \pmsd{0.71}{0.05} \\
    \methodname, never acting on its values & \pmsd{0.633}{0.009} & \pmsd{0.608}{0.017} & \pmsd{0.630}{0.009} & \pmsd{0.624}{0.014} & \pmsd{0.611}{0.026} & \pmsd{0.626}{0.013} & \pmsd{0.626}{0.013} & \pmsd{0.62}{0.01} \\
    \methodname, anchored ($\alpha_0=0.9$) & \pmsd{0.635}{0.015} & \pmsd{0.612}{0.024} & \pmsd{0.646}{0.047} & \second{\pmsd{0.805}{0.114}} & \second{\pmsd{1.033}{0.133}} & \pmsd{0.669}{0.063} & \pmsd{0.733}{0.085} & \second{\pmsd{0.75}{0.05}} \\
    \methodname, uniform prior ($\alpha_0=0.5$) & \second{\pmsd{0.735}{0.022}} & \second{\pmsd{0.675}{0.041}} & \second{\pmsd{0.736}{0.039}} & \best{\pmsd{0.852}{0.100}} & \best{\pmsd{1.043}{0.130}} & \second{\pmsd{0.751}{0.055}} & \best{\pmsd{0.797}{0.073}} & \best{\pmsd{0.81}{0.05}} \\
    \bottomrule
  \end{tabular}
\end{table*}

\subsection{ClassAlloc before and after the knowledge cutoff}\label{app:classalloc}
Table~\ref{tab:classalloc} splits the ClassAlloc results at the knowledge cutoff of gpt-4.1-mini (June 2024; \% per month, mean $\pm$ sd over seeds, averaged over the three investors). Before the cutoff, every LLM-based method except FinMem and Reflexion beats 1/$N$, by up to 0.12~pp per month (Self-consistency); after it, every LLM-based method trails 1/$N$, \methodname the least (0.746\% against 0.787\%) and the memory-free draft by 0.08~pp. As on the other benchmarks, outcome credit tracks the market (Spearman correlation 0.97, Table~\ref{tab:confound}), and the retrieved experiences change the decision detectably in 53--57 of the 71 months. Over all 71 months, the memory-free draft beats 1/$N$ by 0.03~pp per month, but no form of memory improves on it: retrieval, the experience-learning agents and the credit-based learners earn 0.59--0.65\%, against 0.66\% for the memory-free draft and 0.69\% for Self-consistency (Table~\ref{tab:main}). For the memory-free draft, the difference between the periods before and after the cutoff is not significant ($p=0.36$, Welch $t$-test). The trust test of \methodname passed on none of the decisions of 2020, on 3\% of those of 2021 and on 13--33\% of those of each later year, and on 0\%, 39\% and 4\% of the decisions of the three seeds. On the 93 decisions on which it passed, the trusted draft earned $0.05$~pp less (s.e.\ $0.07$) than the anchored blend it replaced, and more on 48\% of them.

\begin{table}[t]
  \caption{ClassAlloc before and after the knowledge cutoff of gpt-4.1-mini (\% per month). Most LLM-based methods beat 1/$N$ before the cutoff; after it, all trail it, \methodname the least.}
  \label{tab:classalloc}
  \footnotesize
  \setlength{\tabcolsep}{3pt}
  \begin{tabular}{@{}lccc@{}}
    \toprule
    Method & \makecell{All 71\\months} & \makecell{50 months\\before cutoff} & \makecell{21 months\\after cutoff} \\
    \midrule
    1/$N$ & \pmsd{0.623}{0.000} & \pmsd{0.554}{0.000} & \best{\pmsd{0.787}{0.000}} \\
    Minimum variance & \pmsd{0.230}{0.000} & \pmsd{0.178}{0.000} & \pmsd{0.353}{0.000} \\
    Memory-free & \second{\pmsd{0.657}{0.001}} & \second{\pmsd{0.636}{0.008}} & \pmsd{0.705}{0.017} \\
    Self-consistency & \best{\pmsd{0.686}{0.007}} & \best{\pmsd{0.672}{0.010}} & \pmsd{0.721}{0.013} \\
    Similarity retr.\ (top-2) & \pmsd{0.649}{0.033} & \pmsd{0.616}{0.043} & \pmsd{0.728}{0.008} \\
    Similarity retr.\ (top-4) & \pmsd{0.643}{0.039} & \pmsd{0.621}{0.047} & \pmsd{0.696}{0.027} \\
    FinMem\adapted & \pmsd{0.594}{0.017} & \pmsd{0.538}{0.021} & \pmsd{0.725}{0.026} \\
    MemRL\adapted & \pmsd{0.633}{0.012} & \pmsd{0.612}{0.030} & \pmsd{0.683}{0.033} \\
    Reflexion & \pmsd{0.590}{0.061} & \pmsd{0.545}{0.072} & \pmsd{0.697}{0.064} \\
    ExpeL & \pmsd{0.609}{0.006} & \pmsd{0.593}{0.005} & \pmsd{0.648}{0.012} \\
    Uplift credit\adapted & \pmsd{0.645}{0.022} & \pmsd{0.617}{0.031} & \pmsd{0.711}{0.001} \\
    Counterfactual sel. & \pmsd{0.650}{0.033} & \pmsd{0.632}{0.039} & \pmsd{0.691}{0.047} \\
    Draft averaging & \pmsd{0.651}{0.008} & \pmsd{0.632}{0.007} & \pmsd{0.696}{0.012} \\
    Hedge & \pmsd{0.634}{0.008} & \pmsd{0.601}{0.009} & \pmsd{0.714}{0.012} \\
    \ours & \pmsd{0.613}{0.017} & \pmsd{0.557}{0.009} & \second{\pmsd{0.746}{0.050}} \\
    \methodname w/o trust test & \pmsd{0.612}{0.005} & \pmsd{0.572}{0.030} & \pmsd{0.707}{0.061} \\
    \bottomrule
  \end{tabular}
\end{table}

\subsection{Other LLMs and temperatures}\label{app:robust-all}
Table~\ref{tab:robust-full} summarizes every configuration of \S\ref{sec:robust}: on InvestorBench, in bp per day (1/$N$ earns 10.45 in every configuration), with the experience pool, retrieval and test days of the main run; on ClassAlloc, in \% per month (1/$N$ earns 0.623), where each LLM writes its own warm-up lessons. The temperature is 0.7 unless stated, and gpt-5-mini runs at the temperature fixed by its API. \emph{Best other LLM-based method} is the best of the 12 LLM-based baselines; Improv.\ and the trust rate are defined as in Table~\ref{tab:main}, and $p$ is the $p$-value of the one-sided paired $t$-test over test dates. Tables~\ref{tab:robust-all} and~\ref{tab:robust-all-llm} give every method on InvestorBench (the column for temperature 0.7 is the main run), and Table~\ref{tab:cb-models} every method on ClassAlloc; all report the mean $\pm$ sd over seeds, averaged over the three investors.

\begin{table*}[t]
  \caption{Robustness to the LLM and its temperature, all configurations. \methodname beats every LLM-based baseline in all eleven InvestorBench configurations and stays within 0.021~pp of 1/$N$ with all eight LLMs on ClassAlloc.}
  \label{tab:robust-full}
  \footnotesize
  \setlength{\tabcolsep}{3.2pt}
  \begin{tabular}{@{}lccccccc@{}}
    \toprule
    LLM, temperature & Memory-free & \makecell{Best experience-\\learning agent} & \makecell{Best other\\LLM-based method} & \ours & \makecell{\methodname\\w/o trust test} & \makecell{Improv.\\(\%, $p$)} & \makecell{Trust\\rate (\%)} \\
    \midrule
    \multicolumn{8}{@{}l}{\textit{(a) InvestorBench (bp per day)}} \\
    gpt-4.1-mini, 0 & \pmsd{7.74}{0.35} & ExpeL \pmsd{9.46}{0.09} & ExpeL \pmsd{9.46}{0.09} & \best{\pmsd{9.84}{0.51}} & \second{\pmsd{9.69}{0.22}} & \improvpos{\pmsd{$+4.0$}{6.0}} (0.32) & \pmsd{23.2}{7.9} \\
    gpt-4.1-mini, 0.3 & \pmsd{7.63}{0.25} & ExpeL \second{\pmsd{9.40}{0.15}} & ExpeL \second{\pmsd{9.40}{0.15}} & \best{\pmsd{10.54}{0.32}} & \pmsd{9.33}{0.28} & \improvpos{\pmsd{$+12.1$}{4.0}} (0.09) & \pmsd{7.0}{12.1} \\
    gpt-4.1-mini, 0.7 (main) & \pmsd{8.43}{0.29} & ExpeL \pmsd{8.93}{0.32} & Hedge \pmsd{8.97}{0.23} & \best{\pmsd{10.40}{0.04}} & \second{\pmsd{9.91}{0.45}} & \improvpos{\pmsd{$+16.4$}{3.2}} (0.12) & \pmsd{0.3}{0.5} \\
    gpt-4.1-mini, 1.0 & \pmsd{7.83}{0.47} & ExpeL \pmsd{9.51}{0.45} & ExpeL \pmsd{9.51}{0.45} & \best{\pmsd{10.36}{0.01}} & \second{\pmsd{10.10}{0.54}} & \improvpos{\pmsd{$+9.0$}{4.8}} (0.15) & \pmsd{0.0}{0.0} \\
    gpt-4.1-nano, 0.7 & \pmsd{8.45}{0.74} & MemRL \pmsd{7.75}{0.55} & Self-consistency \pmsd{9.07}{0.39} & \best{\pmsd{10.30}{0.06}}\signif & \second{\pmsd{9.22}{1.25}} & \improvpos{\pmsd{$+33.0$}{7.4}} (0.02) & \pmsd{2.1}{3.6} \\
    gpt-5-mini, default & \pmsd{8.36}{0.54} & MemRL \pmsd{8.98}{0.07} & Hedge \pmsd{9.19}{0.11} & \best{\pmsd{10.29}{0.12}} & \second{\pmsd{9.48}{0.79}} & \improvpos{\pmsd{$+14.7$}{2.1}} (0.14) & \pmsd{2.3}{0.9} \\
    Llama-3.3-70B, 0.7 & \pmsd{8.34}{0.39} & FinMem \pmsd{9.18}{0.37} & Counterfactual sel. \pmsd{9.39}{0.44} & \second{\pmsd{10.38}{0.04}} & \best{\pmsd{10.41}{0.16}} & \improvpos{\pmsd{$+13.1$}{4.0}} (0.10) & \pmsd{0.8}{1.1} \\
    Qwen3-235B, 0.7 & \pmsd{8.31}{0.21} & MemRL \pmsd{8.86}{0.64} & Hedge \second{\pmsd{8.97}{0.21}} & \best{\pmsd{10.07}{0.25}} & \pmsd{8.94}{0.20} & \improvpos{\pmsd{$+13.6$}{8.6}} (0.12) & \pmsd{9.2}{10.3} \\
    DeepSeek-V3.1, 0.7 & \pmsd{8.37}{0.55} & ExpeL \second{\pmsd{9.71}{0.14}} & ExpeL \second{\pmsd{9.71}{0.14}} & \best{\pmsd{10.33}{0.12}} & \pmsd{9.57}{0.27} & \improvpos{\pmsd{$+6.4$}{2.3}} (0.13) & \pmsd{1.1}{1.0} \\
    Gemini 2.5 Flash, 0.7 & \pmsd{7.66}{0.23} & FinMem \pmsd{8.97}{0.27} & FinMem \pmsd{8.97}{0.27} & \best{\pmsd{10.36}{0.00}} & \second{\pmsd{9.53}{0.47}} & \improvpos{\pmsd{$+15.5$}{3.0}} (0.09) & \pmsd{0.0}{0.0} \\
    Claude Haiku 4.5, 0.7 & \pmsd{8.42}{0.10} & FinMem \pmsd{9.40}{0.11} & FinMem \pmsd{9.40}{0.11} & \best{\pmsd{10.35}{0.03}} & \second{\pmsd{9.48}{0.69}} & \improvpos{\pmsd{$+10.1$}{1.6}} (0.14) & \pmsd{1.3}{1.8} \\
    \midrule
    \multicolumn{8}{@{}l}{\textit{(b) ClassAlloc (\% per month)}} \\
    gpt-4.1-mini (main) & \second{\pmsd{0.657}{0.001}} & MemRL \pmsd{0.633}{0.012} & Self-consistency \best{\pmsd{0.686}{0.007}} & \pmsd{0.613}{0.017} & \pmsd{0.612}{0.005} & \improvneg{\pmsd{$-3.2$}{1.4}} (0.65) & \pmsd{14.6}{21.7} \\
    gpt-4.1-nano & \pmsd{0.587}{0.007} & Reflexion \pmsd{0.575}{0.040} & Counterfactual sel. \best{\pmsd{0.617}{0.016}} & \pmsd{0.609}{0.005} & \second{\pmsd{0.616}{0.007}} & \improvpos{\pmsd{$+5.9$}{7.8}} (0.26) & \pmsd{8.5}{6.0} \\
    gpt-5-mini & \second{\pmsd{0.713}{0.013}} & ExpeL \pmsd{0.690}{0.013} & Self-consistency \best{\pmsd{0.719}{0.012}} & \pmsd{0.643}{0.017} & \pmsd{0.642}{0.015} & \improvneg{\pmsd{$-6.7$}{0.7}} (0.76) & \pmsd{16.3}{27.8} \\
    Llama-3.3-70B & \second{\pmsd{0.619}{0.059}} & FinMem \pmsd{0.589}{0.015} & Memory-free \second{\pmsd{0.619}{0.059}} & \best{\pmsd{0.622}{0.000}} & \best{\pmsd{0.622}{0.013}} & \improvpos{\pmsd{$+5.6$}{2.5}} (0.26) & \pmsd{0.9}{1.6} \\
    Qwen3-235B & \second{\pmsd{0.688}{0.022}} & MemRL \pmsd{0.609}{0.016} & Self-consistency \best{\pmsd{0.698}{0.021}} & \pmsd{0.626}{0.001} & \pmsd{0.640}{0.015} & \improvpos{\pmsd{$+2.8$}{2.7}} (0.38) & \pmsd{3.4}{2.4} \\
    DeepSeek-V3.1 & \pmsd{0.597}{0.018} & Reflexion \second{\pmsd{0.624}{0.049}} & Sim.\ retrieval (top-4) \best{\pmsd{0.625}{0.016}} & \pmsd{0.607}{0.013} & \pmsd{0.604}{0.014} & \improvneg{\pmsd{$-2.7$}{10.0}} (0.65) & \pmsd{12.8}{11.7} \\
    Gemini 2.5 Flash & \pmsd{0.616}{0.012} & ExpeL \pmsd{0.613}{0.010} & Sim.\ retrieval (top-2) \best{\pmsd{0.656}{0.026}} & \pmsd{0.621}{0.010} & \second{\pmsd{0.625}{0.039}} & \improvpos{\pmsd{$+1.3$}{0.7}} (0.44) & \pmsd{21.9}{26.4} \\
    Claude Haiku 4.5 & \best{\pmsd{0.668}{0.020}} & MemRL \second{\pmsd{0.637}{0.006}} & Memory-free \best{\pmsd{0.668}{0.020}} & \pmsd{0.629}{0.000} & \pmsd{0.618}{0.004} & \improvneg{\pmsd{$-1.2$}{0.9}} (0.56) & \pmsd{0.0}{0.0} \\
    \bottomrule
  \end{tabular}
\end{table*}

In every InvestorBench configuration, the best LLM-based baseline earns 8.97--9.71~bp per day, 1/$N$ 10.45 and \methodname 9.84--10.54.

\begin{table*}[t]
  \caption{InvestorBench with gpt-4.1-mini at four temperatures, all methods (bp per day). At every temperature, no LLM-based baseline beats 1/$N$ and \methodname is the best LLM-based method.}
  \label{tab:robust-all}
  \footnotesize
  \setlength{\tabcolsep}{2.4pt}
  \begin{tabular}{@{}lcccc@{}}
    \toprule
    & \multicolumn{4}{c}{gpt-4.1-mini, temperature} \\
    \cmidrule(l){2-5}
    Method & 0 & 0.3 & 0.7 (main) & 1.0 \\
    \midrule
    1/$N$ & \best{\pmsd{10.45}{0.00}} & \second{\pmsd{10.45}{0.00}} & \best{\pmsd{10.45}{0.00}} & \best{\pmsd{10.45}{0.00}} \\
    Minimum variance & \pmsd{5.53}{0.00} & \pmsd{5.53}{0.00} & \pmsd{5.53}{0.00} & \pmsd{5.53}{0.00} \\
    Memory-free & \pmsd{7.74}{0.35} & \pmsd{7.63}{0.25} & \pmsd{8.43}{0.29} & \pmsd{7.83}{0.47} \\
    Self-consistency & \pmsd{7.86}{0.17} & \pmsd{7.81}{0.08} & \pmsd{7.89}{0.21} & \pmsd{7.83}{0.24} \\
    Similarity retrieval (top-2) & \pmsd{8.36}{0.14} & \pmsd{8.73}{0.28} & \pmsd{8.67}{0.19} & \pmsd{8.60}{0.45} \\
    Similarity retrieval (top-4) & \pmsd{8.06}{0.10} & \pmsd{7.91}{0.27} & \pmsd{8.12}{0.61} & \pmsd{8.65}{0.40} \\
    FinMem\adapted & \pmsd{8.45}{0.17} & \pmsd{8.66}{0.37} & \pmsd{8.47}{0.17} & \pmsd{8.36}{0.14} \\
    MemRL\adapted & \pmsd{8.46}{0.25} & \pmsd{8.32}{0.39} & \pmsd{8.90}{0.39} & \pmsd{8.77}{0.25} \\
    Reflexion & \pmsd{6.87}{0.15} & \pmsd{7.59}{0.54} & \pmsd{7.29}{0.66} & \pmsd{7.93}{0.36} \\
    ExpeL & \pmsd{9.46}{0.09} & \pmsd{9.40}{0.15} & \pmsd{8.93}{0.32} & \pmsd{9.51}{0.45} \\
    Uplift credit\adapted & \pmsd{8.52}{0.28} & \pmsd{8.35}{0.24} & \pmsd{8.44}{0.25} & \pmsd{8.44}{0.20} \\
    Counterfactual selection & \pmsd{8.50}{0.07} & \pmsd{8.39}{0.12} & \pmsd{8.54}{0.26} & \pmsd{8.60}{0.65} \\
    Draft averaging & \pmsd{8.55}{0.06} & \pmsd{8.51}{0.06} & \pmsd{8.69}{0.09} & \pmsd{8.69}{0.16} \\
    Hedge & \pmsd{8.70}{0.07} & \pmsd{8.69}{0.12} & \pmsd{8.97}{0.23} & \pmsd{8.96}{0.10} \\
    \ours & \second{\pmsd{9.84}{0.51}} & \best{\pmsd{10.54}{0.32}} & \second{\pmsd{10.40}{0.04}} & \second{\pmsd{10.36}{0.01}} \\
    \methodname w/o trust test & \pmsd{9.69}{0.22} & \pmsd{9.33}{0.28} & \pmsd{9.91}{0.45} & \pmsd{10.10}{0.54} \\
    \bottomrule
  \end{tabular}
\end{table*}

\begin{table*}[t]
  \caption{InvestorBench with eight LLMs, all methods (bp per day). With every LLM, no LLM-based baseline beats 1/$N$ and \methodname beats every LLM-based baseline.}
  \label{tab:robust-all-llm}
  \footnotesize
  \setlength{\tabcolsep}{2.4pt}
  \begin{tabular}{@{}lcccccccc@{}}
    \toprule
    Method & \makecell{gpt-4.1-mini} & \makecell{gpt-4.1-nano} & \makecell{gpt-5-mini} & \makecell{Llama-3.3-70B} & \makecell{Qwen3-235B} & \makecell{DeepSeek-V3.1} & \makecell{Gemini\\2.5 Flash} & \makecell{Claude\\Haiku 4.5} \\
    \midrule
    1/$N$ & \best{\pmsd{10.45}{0.00}} & \best{\pmsd{10.45}{0.00}} & \best{\pmsd{10.45}{0.00}} & \best{\pmsd{10.45}{0.00}} & \best{\pmsd{10.45}{0.00}} & \best{\pmsd{10.45}{0.00}} & \best{\pmsd{10.45}{0.00}} & \best{\pmsd{10.45}{0.00}} \\
    Minimum variance & \pmsd{5.53}{0.00} & \pmsd{5.53}{0.00} & \pmsd{5.53}{0.00} & \pmsd{5.53}{0.00} & \pmsd{5.53}{0.00} & \pmsd{5.53}{0.00} & \pmsd{5.53}{0.00} & \pmsd{5.53}{0.00} \\
    Memory-free & \pmsd{8.43}{0.29} & \pmsd{8.45}{0.74} & \pmsd{8.36}{0.54} & \pmsd{8.34}{0.39} & \pmsd{8.31}{0.21} & \pmsd{8.37}{0.55} & \pmsd{7.66}{0.23} & \pmsd{8.42}{0.10} \\
    Self-consistency & \pmsd{7.89}{0.21} & \pmsd{9.07}{0.39} & \pmsd{8.31}{0.31} & \pmsd{8.26}{0.22} & \pmsd{8.37}{0.17} & \pmsd{8.63}{0.51} & \pmsd{7.92}{0.23} & \pmsd{8.48}{0.07} \\
    Similarity retrieval (top-2) & \pmsd{8.67}{0.19} & \pmsd{6.89}{0.92} & \pmsd{8.41}{0.17} & \pmsd{8.90}{0.28} & \pmsd{8.15}{0.44} & \pmsd{8.66}{0.47} & \pmsd{8.50}{0.41} & \pmsd{8.98}{0.12} \\
    Similarity retrieval (top-4) & \pmsd{8.12}{0.61} & \pmsd{7.71}{0.66} & \pmsd{8.46}{0.24} & \pmsd{8.51}{0.14} & \pmsd{8.00}{0.32} & \pmsd{8.55}{0.45} & \pmsd{8.22}{0.08} & \pmsd{8.83}{0.15} \\
    FinMem\adapted & \pmsd{8.47}{0.17} & \pmsd{7.47}{0.72} & \pmsd{8.87}{0.22} & \pmsd{9.18}{0.37} & \pmsd{8.59}{0.48} & \pmsd{8.94}{0.35} & \pmsd{8.97}{0.27} & \pmsd{9.40}{0.11} \\
    MemRL\adapted & \pmsd{8.90}{0.39} & \pmsd{7.75}{0.55} & \pmsd{8.98}{0.07} & \pmsd{9.10}{0.34} & \pmsd{8.86}{0.64} & \pmsd{8.84}{0.55} & \pmsd{8.35}{0.22} & \pmsd{9.38}{0.31} \\
    Reflexion & \pmsd{7.29}{0.66} & \pmsd{6.58}{0.94} & \pmsd{5.91}{1.45} & \pmsd{8.33}{0.61} & \pmsd{6.10}{1.13} & \pmsd{8.24}{0.22} & \pmsd{6.57}{0.82} & \pmsd{9.05}{0.68} \\
    ExpeL & \pmsd{8.93}{0.32} & \pmsd{7.38}{0.41} & \pmsd{8.82}{0.51} & \pmsd{7.79}{0.19} & \pmsd{8.29}{0.27} & \pmsd{9.71}{0.14} & \pmsd{8.72}{0.21} & \pmsd{9.11}{0.14} \\
    Uplift credit\adapted & \pmsd{8.44}{0.25} & \pmsd{6.62}{0.55} & \pmsd{8.37}{0.33} & \pmsd{9.04}{0.29} & \pmsd{8.38}{0.32} & \pmsd{8.82}{0.30} & \pmsd{8.44}{0.18} & \pmsd{8.71}{0.20} \\
    Counterfactual selection & \pmsd{8.54}{0.26} & \pmsd{6.79}{0.44} & \pmsd{9.01}{0.09} & \pmsd{9.39}{0.44} & \pmsd{8.68}{0.53} & \pmsd{9.25}{0.31} & \pmsd{8.61}{0.37} & \pmsd{8.80}{0.26} \\
    Draft averaging & \pmsd{8.69}{0.09} & \pmsd{8.42}{0.10} & \pmsd{8.80}{0.11} & \pmsd{9.25}{0.04} & \pmsd{8.56}{0.09} & \pmsd{9.14}{0.13} & \pmsd{8.57}{0.07} & \pmsd{8.93}{0.04} \\
    Hedge & \pmsd{8.97}{0.23} & \pmsd{8.98}{0.21} & \pmsd{9.19}{0.11} & \pmsd{9.29}{0.16} & \pmsd{8.97}{0.21} & \pmsd{9.33}{0.23} & \pmsd{8.96}{0.07} & \pmsd{9.39}{0.05} \\
    \ours & \second{\pmsd{10.40}{0.04}} & \second{\pmsd{10.30}{0.06}} & \second{\pmsd{10.29}{0.12}} & \pmsd{10.38}{0.04} & \second{\pmsd{10.07}{0.25}} & \second{\pmsd{10.33}{0.12}} & \second{\pmsd{10.36}{0.00}} & \second{\pmsd{10.35}{0.03}} \\
    \methodname w/o trust test & \pmsd{9.91}{0.45} & \pmsd{9.22}{1.25} & \pmsd{9.48}{0.79} & \second{\pmsd{10.41}{0.16}} & \pmsd{8.94}{0.20} & \pmsd{9.57}{0.27} & \pmsd{9.53}{0.47} & \pmsd{9.48}{0.69} \\
    \bottomrule
  \end{tabular}
\end{table*}

\subsection{ClassAlloc with other LLMs}\label{app:cb-llms}
\begin{figure*}[t]
  \centering
  \includegraphics[width=\textwidth]{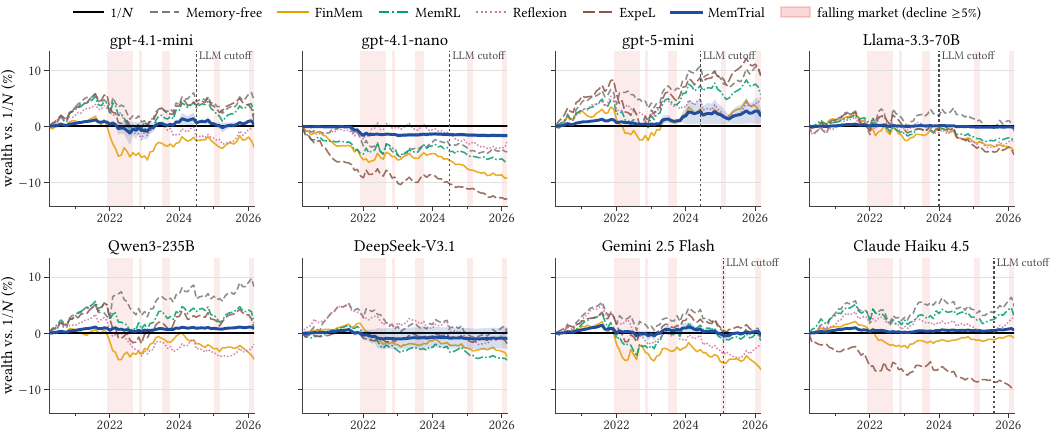}
  \caption{ClassAlloc with other LLMs: wealth relative to 1/$N$. \methodname stays within 3\% of 1/$N$ with every LLM, whereas the memory-free LLM and the other agents move up to 14\% away.}
  \Description{Eight line charts, one per LLM on ClassAlloc, showing wealth relative to 1/N for the balanced investor, with a dotted line at the LLM's knowledge cutoff. MemTrial stays within 3 percent of 1/N; the other agents move up to 14 percent away.}
  \label{fig:wealth-cb-llms}
\end{figure*}

Table~\ref{tab:cb-models} gives every method of Table~\ref{tab:robust-full}b on ClassAlloc, Figure~\ref{fig:wealth-cb-llms} the wealth relative to 1/$N$, and Table~\ref{tab:cb-cutoff-llm} the results before and after each LLM's own knowledge cutoff (utility minus that of 1/$N$, mean over seeds; \emph{Months} counts the test months before and after the cutoff). For Claude Haiku 4.5 we use the end of its training data, July 2025 (its reliable-knowledge cutoff is February 2025); Qwen3-235B and DeepSeek-V3.1 have no published knowledge cutoff. Each LLM writes its own warm-up lessons, which are embedded and retrieved as in the main run. The memory-free draft beats 1/$N$ over the whole test period with four of the eight LLMs (gpt-4.1-mini, gpt-5-mini, Qwen3-235B and Claude Haiku 4.5). Of the six LLMs whose knowledge cutoff is published, it trails 1/$N$ after the cutoff with five (by 0.06--0.09~pp per month), which is what one would expect if the LLMs had seen the earlier outcomes during pretraining; with Claude Haiku 4.5, it beats 1/$N$ also in the eight months after the cutoff (by 0.08~pp). \methodname stays within 0.021~pp per month of 1/$N$ with every LLM; with gpt-5-mini and Qwen3-235B, whose memory-free drafts beat 1/$N$ by the widest margins, its anchor gives up most of their advantage.

\begin{table*}[t]
  \caption{ClassAlloc with eight LLMs, all methods (\% per month). \methodname stays within 0.021~pp of 1/$N$ with every LLM; the memory-free LLM beats 1/$N$ with four of them.}
  \label{tab:cb-models}
  \footnotesize
  \setlength{\tabcolsep}{2.4pt}
  \begin{tabular}{@{}lcccccccc@{}}
    \toprule
    Method & gpt-4.1-mini & gpt-4.1-nano & gpt-5-mini & Llama-3.3-70B & Qwen3-235B & DeepSeek-V3.1 & \makecell{Gemini\\2.5 Flash} & \makecell{Claude\\Haiku 4.5} \\
    \midrule
    1/$N$ & \pmsd{0.623}{0.000} & \best{\pmsd{0.623}{0.000}} & \pmsd{0.623}{0.000} & \best{\pmsd{0.623}{0.000}} & \pmsd{0.623}{0.000} & \pmsd{0.623}{0.000} & \pmsd{0.623}{0.000} & \pmsd{0.623}{0.000} \\
    Minimum variance & \pmsd{0.230}{0.000} & \pmsd{0.230}{0.000} & \pmsd{0.230}{0.000} & \pmsd{0.230}{0.000} & \pmsd{0.230}{0.000} & \pmsd{0.230}{0.000} & \pmsd{0.230}{0.000} & \pmsd{0.230}{0.000} \\
    Memory-free & \second{\pmsd{0.657}{0.001}} & \pmsd{0.587}{0.007} & \second{\pmsd{0.713}{0.013}} & \pmsd{0.619}{0.059} & \second{\pmsd{0.688}{0.022}} & \pmsd{0.597}{0.018} & \pmsd{0.616}{0.012} & \best{\pmsd{0.668}{0.020}} \\
    Self-consistency & \best{\pmsd{0.686}{0.007}} & \pmsd{0.588}{0.014} & \best{\pmsd{0.719}{0.012}} & \pmsd{0.614}{0.017} & \best{\pmsd{0.698}{0.021}} & \pmsd{0.615}{0.024} & \pmsd{0.618}{0.003} & \second{\pmsd{0.656}{0.004}} \\
    Similarity retrieval (top-2) & \pmsd{0.649}{0.033} & \pmsd{0.557}{0.000} & \pmsd{0.701}{0.019} & \pmsd{0.602}{0.009} & \pmsd{0.599}{0.006} & \pmsd{0.603}{0.009} & \best{\pmsd{0.656}{0.026}} & \pmsd{0.640}{0.006} \\
    Similarity retrieval (top-4) & \pmsd{0.643}{0.039} & \pmsd{0.554}{0.019} & \pmsd{0.655}{0.006} & \pmsd{0.598}{0.005} & \pmsd{0.631}{0.019} & \best{\pmsd{0.625}{0.016}} & \pmsd{0.586}{0.016} & \pmsd{0.641}{0.009} \\
    FinMem\adapted & \pmsd{0.594}{0.017} & \pmsd{0.532}{0.017} & \pmsd{0.632}{0.025} & \pmsd{0.589}{0.015} & \pmsd{0.564}{0.006} & \pmsd{0.570}{0.008} & \pmsd{0.567}{0.026} & \pmsd{0.608}{0.008} \\
    MemRL\adapted & \pmsd{0.633}{0.012} & \pmsd{0.569}{0.030} & \pmsd{0.661}{0.022} & \pmsd{0.579}{0.024} & \pmsd{0.609}{0.016} & \pmsd{0.553}{0.015} & \pmsd{0.598}{0.053} & \pmsd{0.637}{0.006} \\
    Reflexion & \pmsd{0.590}{0.061} & \pmsd{0.575}{0.040} & \pmsd{0.659}{0.013} & \pmsd{0.560}{0.003} & \pmsd{0.550}{0.069} & \second{\pmsd{0.624}{0.049}} & \pmsd{0.585}{0.052} & \pmsd{0.633}{0.031} \\
    ExpeL & \pmsd{0.609}{0.006} & \pmsd{0.475}{0.025} & \pmsd{0.690}{0.013} & \pmsd{0.565}{0.021} & \pmsd{0.606}{0.012} & \pmsd{0.597}{0.007} & \pmsd{0.613}{0.010} & \pmsd{0.524}{0.017} \\
    Uplift credit\adapted & \pmsd{0.645}{0.022} & \pmsd{0.558}{0.002} & \pmsd{0.677}{0.028} & \pmsd{0.582}{0.007} & \pmsd{0.640}{0.005} & \pmsd{0.620}{0.022} & \pmsd{0.617}{0.023} & \pmsd{0.638}{0.011} \\
    Counterfactual selection & \pmsd{0.650}{0.033} & \second{\pmsd{0.617}{0.016}} & \pmsd{0.685}{0.004} & \pmsd{0.604}{0.013} & \pmsd{0.646}{0.007} & \pmsd{0.622}{0.020} & \second{\pmsd{0.640}{0.011}} & \pmsd{0.624}{0.004} \\
    Draft averaging & \pmsd{0.651}{0.008} & \pmsd{0.587}{0.003} & \pmsd{0.677}{0.005} & \pmsd{0.604}{0.011} & \pmsd{0.652}{0.009} & \pmsd{0.620}{0.011} & \pmsd{0.622}{0.003} & \pmsd{0.650}{0.002} \\
    Hedge & \pmsd{0.634}{0.008} & \pmsd{0.583}{0.002} & \pmsd{0.665}{0.011} & \pmsd{0.602}{0.010} & \pmsd{0.632}{0.009} & \pmsd{0.602}{0.007} & \pmsd{0.604}{0.012} & \pmsd{0.640}{0.008} \\
    \ours & \pmsd{0.613}{0.017} & \pmsd{0.609}{0.005} & \pmsd{0.643}{0.017} & \second{\pmsd{0.622}{0.000}} & \pmsd{0.626}{0.001} & \pmsd{0.607}{0.013} & \pmsd{0.621}{0.010} & \pmsd{0.629}{0.000} \\
    \methodname w/o trust test & \pmsd{0.612}{0.005} & \pmsd{0.616}{0.007} & \pmsd{0.642}{0.015} & \second{\pmsd{0.622}{0.013}} & \pmsd{0.640}{0.015} & \pmsd{0.604}{0.014} & \pmsd{0.625}{0.039} & \pmsd{0.618}{0.004} \\
    \bottomrule
  \end{tabular}
\end{table*}

\begin{table}[t]
  \caption{ClassAlloc before and after each LLM's knowledge cutoff (utility minus that of 1/$N$, pp per month). With five of the six LLMs whose cutoff is published, the memory-free draft trails 1/$N$ after the cutoff.}
  \label{tab:cb-cutoff-llm}
  \footnotesize
  \setlength{\tabcolsep}{2.2pt}
  \begin{tabular}{@{}lllrrrr@{}}
    \toprule
    & & & \multicolumn{2}{c}{Memory-free} & \multicolumn{2}{c}{\ours} \\
    \cmidrule(lr){4-5}\cmidrule(l){6-7}
    LLM & Cutoff & Months & before & after & before & after \\
    \midrule
    gpt-4.1-mini & Jun 2024 & 50/21 & $+0.082$ & $-0.082$ & $+0.003$ & $-0.041$ \\
    gpt-4.1-nano & Jun 2024 & 50/21 & $-0.018$ & $-0.078$ & $-0.018$ & $-0.007$ \\
    gpt-5-mini & May 2024 & 49/22 & $+0.157$ & $-0.059$ & $+0.040$ & $-0.024$ \\
    Llama-3.3-70B & Dec 2023 & 44/27 & $+0.047$ & $-0.087$ & $+0.005$ & $-0.009$ \\
    Qwen3-235B & -- &  &  &  &  &  \\
    DeepSeek-V3.1 & -- &  &  &  &  &  \\
    Gemini 2.5 Flash & Jan 2025 & 57/14 & $+0.008$ & $-0.069$ & $+0.001$ & $-0.014$ \\
    Claude Haiku 4.5 & Jul 2025 & 63/8 & $+0.040$ & $+0.079$ & $+0.005$ & $+0.013$ \\
    \bottomrule
  \end{tabular}
\end{table}

\subsection{Look-ahead checks}\label{app:lookahead}
The LLMs were trained on text that may describe the test periods, a source of look-ahead bias. We ran two checks on ClassAlloc, at a total cost of USD 3.87.

\textbf{Memorization probe.} For each of the 129 ClassAlloc decision dates (58 warm-up and 71 test months, July 2015 to March 2026), we gave each LLM only the date and asked for the return of each asset class over the 20 trading days for which a decision is held, with no market data and no memory (prompt in Appendix~\ref{app:prompts-cb}); three draws per date, at the temperature of the runs. We compare the answers with the realized returns before and after each LLM's knowledge cutoff in two ways. \emph{Absolute recall} is the correlation over months between recalled and realized returns, averaged over the four non-cash classes. \emph{Relative recall} is the correlation after removing each class's mean and each month's mean, that is, the recall of which class did better than usual in which month, which is what matters for beating 1/$N$. Table~\ref{tab:probe} gives the mean $\pm$ sd over the three draws; each LLM has 102--121 dates before its cutoff, warm-up months included, and 8--27 after it. Every LLM whose cutoff is published recalls the outcomes before its cutoff (absolute recall 0.10--0.63, relative recall 0.06--0.50) and hardly at all after it ($-0.16$ to $+0.09$ and $-0.10$ to $+0.01$). Gemini 2.5 Flash recalls the most, and Qwen3-235B and DeepSeek-V3.1, whose cutoffs are not published, have an absolute recall of 0.27 and 0.22 over all months. The advantage of the memory-free draft before the cutoff does not grow with recall: with Gemini 2.5 Flash, which recalls the most, it beats 1/$N$ by only 0.008~pp per month before the cutoff, against 0.157~pp with gpt-5-mini (Table~\ref{tab:cb-cutoff-llm}).

\begin{table}[t]
  \caption{Memorization probe: correlation of the asset-class returns that each LLM recalls from a date alone with the realized ClassAlloc returns. Every LLM recalls the outcomes before its knowledge cutoff, and hardly any after it.}
  \label{tab:probe}
  \footnotesize
  \setlength{\tabcolsep}{2.6pt}
  \begin{tabular}{@{}llcccc@{}}
    \toprule
    & & \multicolumn{2}{c}{Absolute recall} & \multicolumn{2}{c}{Relative recall} \\
    \cmidrule(lr){3-4}\cmidrule(l){5-6}
    LLM & Cutoff & before & after & before & after \\
    \midrule
    gpt-4.1-mini & Jun 2024 & \pmsd{$+0.21$}{0.02} & \pmsd{$-0.05$}{0.12} & \pmsd{$+0.19$}{0.01} & \pmsd{$-0.02$}{0.03} \\
    gpt-4.1-nano & Jun 2024 & \pmsd{$+0.10$}{0.04} & \pmsd{$0.00$}{0.04} & \pmsd{$+0.06$}{0.06} & \pmsd{$-0.10$}{0.09} \\
    gpt-5-mini & May 2024 & \pmsd{$+0.22$}{0.06} & \pmsd{$-0.04$}{0.09} & \pmsd{$+0.20$}{0.04} & \pmsd{$-0.10$}{0.13} \\
    Llama-3.3-70B & Dec 2023 & \pmsd{$+0.40$}{0.09} & \pmsd{$+0.09$}{0.15} & \pmsd{$+0.36$}{0.08} & \pmsd{$+0.01$}{0.12} \\
    Qwen3-235B & -- & \pmsd{$+0.27$}{0.03} & -- & \pmsd{$+0.21$}{0.05} & -- \\
    DeepSeek-V3.1 & -- & \pmsd{$+0.22$}{0.09} & -- & \pmsd{$+0.19$}{0.05} & -- \\
    Gemini 2.5 Flash & Jan 2025 & \pmsd{$+0.63$}{0.03} & \pmsd{$-0.07$}{0.02} & \pmsd{$+0.50$}{0.02} & \pmsd{$-0.09$}{0.06} \\
    Claude Haiku 4.5 & Jul 2025 & \pmsd{$+0.27$}{0.01} & \pmsd{$-0.16$}{0.07} & \pmsd{$+0.27$}{0.01} & \pmsd{$-0.07$}{0.10} \\
    \bottomrule
  \end{tabular}
\end{table}

\textbf{Dates hidden.} We evaluated gpt-4.1-mini and gpt-5-mini on ClassAlloc (three seeds each, with the two extra draws of Self-consistency) with no calendar date shown to the LLM: the market context omits its date line, the warm-up lessons are stored and retrieved without their dates, and the ExpeL insight prompt lists the lessons without their dates. Everything else is unchanged. The lessons and reflections are written by the LLM from the dateless context, and none of the 58 warm-up lessons of either run mentions a year or a month. Table~\ref{tab:masked} compares these runs with those that show the dates (mean $\pm$ sd over seeds, averaged over the three investors). Hiding the dates leaves the pattern of the memory-free draft intact: with gpt-4.1-mini, it beats 1/$N$ by 0.13~pp per month before the cutoff and trails it by 0.10~pp after (0.08 and 0.08 with the dates shown); with gpt-5-mini, by 0.14 and 0.07~pp (0.16 and 0.06). Over all 71 test months, it beats 1/$N$ by 0.06 and 0.07~pp (0.03 and 0.09 with the dates shown). \methodname stays within 0.02~pp per month of 1/$N$ ($+0.006$ with gpt-4.1-mini and $-0.017$ with gpt-5-mini; trust rates 0.0\% and 6.6\%) and, after the cutoff, trails it by 0.02 and 0.03~pp, against 0.10 and 0.07~pp for the memory-free draft.

\begin{table*}[t]
  \caption{ClassAlloc with every calendar date hidden from the LLM (utility minus that of 1/$N$, pp per month). Hiding the dates does not remove the memory-free draft's advantage before the knowledge cutoff.}
  \label{tab:masked}
  \footnotesize
  \setlength{\tabcolsep}{2.6pt}
  \begin{tabular}{@{}lcccccccc@{}}
    \toprule
    & \multicolumn{4}{c}{gpt-4.1-mini} & \multicolumn{4}{c}{gpt-5-mini} \\
    \cmidrule(lr){2-5}\cmidrule(l){6-9}
    & \multicolumn{2}{c}{dates shown} & \multicolumn{2}{c}{dates hidden} & \multicolumn{2}{c}{dates shown} & \multicolumn{2}{c}{dates hidden} \\
    \cmidrule(lr){2-3}\cmidrule(lr){4-5}\cmidrule(lr){6-7}\cmidrule(l){8-9}
    Method & before & after & before & after & before & after & before & after \\
    \midrule
    Memory-free & \second{\pmsd{$+0.082$}{0.008}} & \pmsd{$-0.082$}{0.017} & \best{\pmsd{$+0.127$}{0.017}} & \pmsd{$-0.099$}{0.027} & \second{\pmsd{$+0.157$}{0.020}} & \pmsd{$-0.059$}{0.025} & \second{\pmsd{$+0.136$}{0.038}} & \pmsd{$-0.066$}{0.006} \\
    Self-consistency & \best{\pmsd{$+0.118$}{0.010}} & \pmsd{$-0.066$}{0.013} & \second{\pmsd{$+0.125$}{0.030}} & \pmsd{$-0.104$}{0.012} & \best{\pmsd{$+0.160$}{0.013}} & \pmsd{$-0.048$}{0.016} & \pmsd{$+0.118$}{0.028} & \pmsd{$-0.068$}{0.015} \\
    FinMem\adapted & \pmsd{$-0.016$}{0.021} & \second{\pmsd{$-0.062$}{0.026}} & \pmsd{$-0.047$}{0.029} & \pmsd{$-0.076$}{0.003} & \pmsd{$+0.019$}{0.025} & \second{\pmsd{$-0.013$}{0.045}} & \pmsd{$-0.024$}{0.016} & \pmsd{$-0.088$}{0.035} \\
    MemRL\adapted & \pmsd{$+0.058$}{0.030} & \pmsd{$-0.104$}{0.033} & \pmsd{$+0.088$}{0.032} & \pmsd{$-0.093$}{0.050} & \pmsd{$+0.077$}{0.046} & \pmsd{$-0.048$}{0.034} & \pmsd{$+0.077$}{0.049} & \second{\pmsd{$-0.012$}{0.041}} \\
    Reflexion & \pmsd{$-0.009$}{0.072} & \pmsd{$-0.090$}{0.064} & \pmsd{$-0.001$}{0.095} & \pmsd{$-0.049$}{0.036} & \pmsd{$+0.066$}{0.026} & \pmsd{$-0.032$}{0.096} & \pmsd{$+0.051$}{0.014} & \pmsd{$-0.093$}{0.019} \\
    ExpeL & \pmsd{$+0.039$}{0.005} & \pmsd{$-0.140$}{0.012} & \pmsd{$+0.009$}{0.041} & \best{\pmsd{$-0.015$}{0.009}} & \pmsd{$+0.089$}{0.018} & \best{\pmsd{$+0.017$}{0.004}} & \pmsd{$+0.033$}{0.032} & \pmsd{$-0.134$}{0.025} \\
    Draft averaging & \pmsd{$+0.077$}{0.007} & \pmsd{$-0.091$}{0.012} & \pmsd{$+0.087$}{0.003} & \pmsd{$-0.100$}{0.006} & \pmsd{$+0.100$}{0.006} & \pmsd{$-0.049$}{0.007} & \pmsd{$+0.107$}{0.004} & \pmsd{$-0.020$}{0.014} \\
    Counterfactual selection & \pmsd{$+0.078$}{0.039} & \pmsd{$-0.096$}{0.047} & \pmsd{$+0.063$}{0.034} & \pmsd{$-0.083$}{0.016} & \pmsd{$+0.118$}{0.005} & \pmsd{$-0.064$}{0.003} & \best{\pmsd{$+0.140$}{0.014}} & \best{\pmsd{$+0.008$}{0.037}} \\
    \ours & \pmsd{$+0.003$}{0.009} & \best{\pmsd{$-0.041$}{0.050}} & \pmsd{$+0.016$}{0.000} & \second{\pmsd{$-0.017$}{0.001}} & \pmsd{$+0.040$}{0.038} & \pmsd{$-0.024$}{0.029} & \pmsd{$-0.010$}{0.047} & \pmsd{$-0.033$}{0.015} \\
    \bottomrule
  \end{tabular}
\end{table*}

\begin{figure}[t]
  \centering
  \includegraphics[width=\columnwidth]{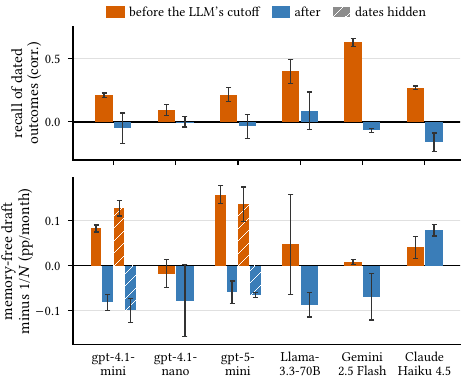}
  \caption{ClassAlloc at each LLM's knowledge cutoff. The LLMs recall dated outcomes from before their cutoffs, but hiding the dates does not remove the memory-free draft's advantage before the cutoff.}
  \Description{Two bar charts for six LLMs. Top: correlation between the asset-class returns an LLM recalls from a date alone and the realized returns, before and after its knowledge cutoff: 0.10 to 0.63 before and minus 0.16 to 0.09 after. Bottom: utility of the memory-free draft minus that of 1/N in pp per month before and after each LLM's cutoff, negative after the cutoff for five of the six LLMs; hatched bars show gpt-4.1-mini and gpt-5-mini with all dates hidden, which keep the same pattern.}
  \label{fig:cutoff}
\end{figure}

Figure~\ref{fig:cutoff} brings these checks together with Table~\ref{tab:cb-cutoff-llm}. The LLMs thus recall dated market outcomes, but the advantage of the memory-free draft before the cutoff does not depend on being told the date. It may come from recognizing a period from the returns shown, or from allocations that happened to suit the markets of 2020--2024. Either way, this effect applies to all LLM-based methods alike, and PlantedMem is free of it.

\subsection{PortBench after the knowledge cutoff}\label{app:pb-forward}
To test on months that gpt-4.1-mini, the LLM of the PortBench runs, cannot have seen, we extended the monthly PortBench stream with the decision dates that follow 2024-12-02 under the same calendar rule (the first trading day of each month; a month is skipped when its 20-day holding window would overlap the previous one). This gives twelve dates from 2025-01-02 to 2026-02-02; March 2025, December 2025 and March 2026 are skipped by the overlap rule, and April 2026 has no complete holding window before the data end on 2026-04-16. Everything else is identical to the original months: the experience pool and the FinMem and MemRL modules (which receive no feedback after July 2019), the 19 drafts per date, both configurations, gpt-4.1-mini at temperature 0.7, three seeds, and the code that builds the inputs and runs the agents, which was reused unchanged. Before any new month was run, the inputs of all 59 original months were rebuilt with the same code and compared byte by byte with the frozen inputs, including the hash of every first LLM request; they were identical. The six Reflexion chains continue from their state after 2024-12-02. The evaluation reuses the code of the main results, run forward over all 71 dates without re-tuning, and reproduces the 2019--2024 results exactly.

The publisher's price files have gaps in this period: 18 large equities (including AAPL, MSFT and NVDA) have no prices from 10 December 2024 to 24 January 2025 and after 19 February 2026, MATIC-USD ends on 24 March 2025 but remains in the agent's market view until June 2025, and INDS has no price on 24 October 2025. As for December 2024 in the original months, a month in which any execution held an asset without a price is dropped for that configuration. This removes six of the twelve months (January, April, May, June and October 2025, and February 2026) and leaves 2025-02-03, 2025-07-01, 2025-08-01, 2025-09-02, 2025-11-03 and 2026-01-02 for both configurations. The new runs cost USD 9.29 for 1,368 executions and USD 0.48 for the Reflexion chains.

Table~\ref{tab:pb-forward} reports the six new months and all eleven test months after the knowledge cutoff of gpt-4.1-mini (June 2024), averaged over the three investors. Over the eleven months, \methodname is the best LLM-based method on both configurations and stays close to 1/$N$ (0.14 and 0.01~pp per month below it), while the other LLM-based methods trail 1/$N$ by 0.36--1.57~pp; its values never passed the trust test after the cutoff. In the six new months alone, the memory-free draft of PortBench-Full beats 1/$N$ by 0.80~pp per month, and \methodname, still anchored at 1/$N$, gains only 0.06~pp; on PortBench-Raw, every LLM-based method is within 0.17~pp of 1/$N$. With six months, no difference involving \methodname is significant.

\begin{table*}[t]
  \caption{PortBench after the knowledge cutoff (\% per month). Over the eleven months after the cutoff, \methodname is the best LLM-based method and stays close to 1/$N$; in the six new months, the memory-free draft beats 1/$N$ on PortBench-Full.}
  \label{tab:pb-forward}
  \footnotesize
  \setlength{\tabcolsep}{6pt}
  \begin{tabular}{@{}lcccc@{}}
    \toprule
    & \multicolumn{2}{c}{New months (2025--2026)} & \multicolumn{2}{c}{All months after the cutoff} \\
    \cmidrule(lr){2-3}\cmidrule(lr){4-5}
    Method & Full & Raw & Full & Raw \\
    \midrule
    \multicolumn{5}{@{}l}{\textit{Rule-based portfolios}} \\
    1/$N$ & \pmsd{1.025}{0.000} & \pmsd{1.025}{0.000} & \best{\pmsd{2.075}{0.000}} & \best{\pmsd{2.075}{0.000}} \\
    Minimum variance & \pmsd{0.029}{0.000} & \pmsd{0.029}{0.000} & \pmsd{0.078}{0.000} & \pmsd{0.078}{0.000} \\
    \multicolumn{5}{@{}l}{\textit{LLM agents without experience learning}} \\
    Memory-free & \second{\pmsd{1.825}{0.228}} & \pmsd{1.055}{0.103} & \pmsd{1.290}{0.277} & \pmsd{1.363}{0.053} \\
    Self-consistency & \best{\pmsd{1.867}{0.000}} & \pmsd{1.058}{0.000} & \pmsd{1.328}{0.000} & \pmsd{1.367}{0.000} \\
    Similarity retrieval (top-2) & \pmsd{1.131}{0.351} & \pmsd{1.059}{0.141} & \pmsd{0.883}{0.299} & \pmsd{1.265}{0.229} \\
    Similarity retrieval (top-4) & \pmsd{1.350}{0.353} & \pmsd{1.002}{0.069} & \pmsd{1.007}{0.283} & \pmsd{1.211}{0.044} \\
    \multicolumn{5}{@{}l}{\textit{Experience-learning agents}} \\
    FinMem & \pmsd{1.071}{0.293} & \pmsd{0.874}{0.203} & \pmsd{0.691}{0.347} & \pmsd{1.183}{0.104} \\
    MemRL & \pmsd{1.358}{0.676} & \best{\pmsd{1.191}{0.147}} & \pmsd{0.916}{0.343} & \pmsd{1.365}{0.050} \\
    Reflexion & \pmsd{0.869}{0.092} & \second{\pmsd{1.091}{0.066}} & \pmsd{0.502}{0.185} & \pmsd{1.404}{0.153} \\
    ExpeL\adapted & \pmsd{1.197}{0.103} & \pmsd{0.920}{0.155} & \pmsd{1.033}{0.174} & \pmsd{1.270}{0.152} \\
    \multicolumn{5}{@{}l}{\textit{Credit-based and ensemble learners}} \\
    Uplift credit\adapted & \pmsd{1.034}{0.528} & \pmsd{1.051}{0.042} & \pmsd{0.598}{0.232} & \pmsd{1.425}{0.237} \\
    Counterfactual selection & \pmsd{1.794}{0.367} & \pmsd{1.024}{0.182} & \pmsd{1.217}{0.242} & \pmsd{1.542}{0.423} \\
    Draft averaging & \pmsd{1.437}{0.109} & \pmsd{1.004}{0.033} & \pmsd{1.039}{0.047} & \pmsd{1.413}{0.115} \\
    Hedge & \pmsd{1.499}{0.130} & \pmsd{1.029}{0.023} & \pmsd{1.234}{0.090} & \pmsd{1.711}{0.032} \\
    \ours & \pmsd{1.086}{0.014} & \pmsd{1.027}{0.000} & \second{\pmsd{1.939}{0.011}} & \second{\pmsd{2.064}{0.002}} \\
    \midrule
    Improv.\ (\%, $p$) & \improvneg{\pmsd{$-20.1$}{49.7}} (0.61) & \improvneg{\pmsd{$-13.8$}{12.3}} (0.64) & \improvpos{\pmsd{$+87.8$}{17.9}} (0.12) & \improvpos{\pmsd{$+47.0$}{10.9}} (0.24) \\
    \methodname $-$ 1/$N$ & \pmsd{$+0.061$}{0.014} & \pmsd{$+0.001$}{0.000} & \pmsd{$-0.135$}{0.011} & \pmsd{$-0.011$}{0.002} \\
    Trust rate of \methodname (\%) & 0.0 & 0.0 & 0.0 & 0.0 \\
    \bottomrule
  \end{tabular}
\end{table*}

\section{Case Study: Two Decisions in the 2022 Bear Market}\label{app:case}

This appendix follows \methodname through two monthly decisions on ClassAlloc (balanced investor), on 1 April and 1 September 2022. Both lie in the largest falling phase of the test period, in which the equal-weight equity index lost 21\% between January and the end of September 2022 (Appendix~\ref{app:wealth}). The retrieved lessons depend only on the date, so all seeds see the same four lessons; the drafts, contributions and learned values differ by seed. In seed~1 the trust test passed on 1 April and narrowly failed on 1 September ($p=0.058$); in seeds 0 and~2 it passed on neither date. The two dates thus show one decision that acted on the learned values and one that kept the anchored action. Table~\ref{tab:case}a lists the eight designed drafts of seed~1 on each date: the lessons each draft was given, its weights in \% (Eq.\ equities, Bd.\ bonds, Cm.\ commodities, RE real estate) and its utility in \% over the following 20 trading days, with the deployed draft in bold (on 1 September, seed~1 deployed the anchored blend) and, in the last row, the realized return of each class over these 20 days. Table~\ref{tab:case}b gives the contributions (Contr.) and learned values of the retrieved lessons in seed~1, in pp. Table~\ref{tab:case}c gives the trust test and the action in each seed: the number $n$ of forward scores, their mean $\bar\rho$, the $p$-value of the test, the action and its utility in \%, with trusted decisions in bold; the column 1/$N$ gives the utility of moving to 1/$N$ from the seed's holdings, which differs slightly across seeds because of trading costs.

\begin{table*}[t]
  \caption{Case study on ClassAlloc (balanced investor). The contributions separate the lessons although every draft lost money; in seed~1, acting on the values cost 1.66~pp in April, and keeping the anchored action gave up 2.02~pp in September.}
  \label{tab:case}
  \footnotesize
  \setlength{\tabcolsep}{2.6pt}
  \centering
  \textit{(a)}\par\smallskip
  \begin{tabular}{@{}lrrrrrr@{\hspace{12pt}}lrrrrrr@{}}
    \toprule
    \multicolumn{7}{@{}l}{1 April 2022} & \multicolumn{7}{l@{}}{1 September 2022} \\
    \cmidrule(r){1-7}\cmidrule(l){8-14}
    Lessons in the draft & Eq. & Bd. & Cm. & RE & Cash & Utility & Lessons in the draft & Eq. & Bd. & Cm. & RE & Cash & Utility \\
    \midrule
    none & 35 & 10 & 25 & 20 & 10 & $-$4.63 & none & 15 & 35 & 15 & 15 & 20 & $-$5.43 \\
    L18/04, L16/08 & 35 & 10 & 20 & 25 & 10 & $-$5.14 & L18/04, L16/08 & 15 & 30 & 10 & 15 & 30 & $-$5.05 \\
    L18/04, L17/05 & 35 & 10 & 15 & 30 & 10 & $-$5.67 & L18/04, L19/10 & 10 & 30 & 10 & 10 & 40 & $-$3.89 \\
    L16/08, L17/05 & 35 & 15 & 10 & 30 & 10 & $-$6.02 & L16/08, L19/10 & 15 & 25 & 10 & 15 & 35 & $-$4.81 \\
    \textbf{L18/04, L15/12} & \textbf{35} & \textbf{15} & \textbf{15} & \textbf{25} & \textbf{10} & \textbf{$-$5.48} & L18/04, L17/05 & 15 & 35 & 10 & 15 & 25 & $-$5.29 \\
    L16/08, L15/12 & 35 & 15 & 10 & 25 & 15 & $-$5.58 & L16/08, L17/05 & 15 & 35 & 10 & 15 & 25 & $-$5.29 \\
    L17/05, L15/12 & 35 & 10 & 15 & 30 & 10 & $-$5.67 & L19/10, L17/05 & 15 & 15 & 20 & 15 & 35 & $-$4.64 \\
    all four & 35 & 10 & 20 & 25 & 10 & $-$5.14 & all four & 15 & 25 & 10 & 15 & 35 & $-$4.81 \\
    \midrule
     & & & & & & & \textbf{anchored blend} & \textbf{19} & \textbf{21} & \textbf{19} & \textbf{19} & \textbf{21} & \textbf{$-$5.91} \\
    1/$N$ & 20 & 20 & 20 & 20 & 20 & $-$3.83 & 1/$N$ & 20 & 20 & 20 & 20 & 20 & $-$6.06 \\
    Class return (\%) & $-$8.1 & $-$4.8 & 2.2 & $-$7.5 & 0.0 & & Class return (\%) & $-$9.5 & $-$4.3 & $-$1.8 & $-$12.3 & 0.1 & \\
    \bottomrule
  \end{tabular}
  \par\bigskip
  \begin{minipage}[t]{0.33\textwidth}\centering\textit{(b)}\par\smallskip
  \begin{tabular}{@{}lrrrr@{}}
    \toprule
    & \multicolumn{2}{c}{1 Apr.} & \multicolumn{2}{c@{}}{1 Sep.} \\
    \cmidrule(lr){2-3}\cmidrule(l){4-5}
    Lesson & Contr. & Value & Contr. & Value \\
    \midrule
    L15/12 & $-$0.11 & +0.004 & -- & -- \\
    L16/08 & $-$0.11 & $-$0.015 & $-$0.18 & $-$0.016 \\
    L17/05 & $-$0.42 & $-$0.002 & $-$0.21 & $-$0.002 \\
    L18/04 & +0.12 & +0.004 & +0.28 & +0.003 \\
    L19/10 & -- & -- & +0.73 & +0.005 \\
    \bottomrule
  \end{tabular}
  \end{minipage}\hfill
  \begin{minipage}[t]{0.62\textwidth}\centering\textit{(c)}\par\smallskip
  \begin{tabular}{@{}llrrrlrr@{}}
    \toprule
    Date & Seed & $n$ & $\bar\rho$ & $p$ & Action & Utility & 1/$N$ \\
    \midrule
    1 Apr. & 0 & 18 & 0.17 & 0.123 & anchored, $q=0.835$ & $-$4.08 & $-$3.82 \\
     & 1 & \textbf{11} & \textbf{0.24} & \textbf{0.045} & \textbf{draft L18/04, L15/12} & \textbf{$-$5.48} & \textbf{$-$3.83} \\
     & 2 & 15 & 0.13 & 0.163 & anchored, $q=0.835$ & $-$4.10 & $-$3.83 \\
    \addlinespace
    1 Sep. & 0 & 22 & 0.07 & 0.286 & anchored, $q=0.885$ & $-$5.84 & $-$5.99 \\
     & 1 & 16 & 0.17 & 0.058 & anchored, $q=0.875$ & $-$5.91 & $-$6.06 \\
     & 2 & 18 & 0.03 & 0.421 & anchored, $q=0.875$ & $-$5.79 & $-$5.99 \\
    \bottomrule
  \end{tabular}
  \end{minipage}
\end{table*}

\textbf{Retrieved lessons.} Each lesson was written by the agent after one of its warm-up decisions, and its label gives the month of that decision. On 1 April 2022 the agent retrieved L18/04, L16/08, L17/05 and L15/12, and on 1 September L18/04, L16/08, L19/10 and L17/05.
{\small
\begin{description}
\item[L15/12] In December 2015, equities and real estate showed strong recent gains with moderate volatility, while commodities declined sharply. I overweighted equities and real estate, avoided commodities, and held cash. Despite this, bonds and real estate declined significantly, hurting returns. In similar scenarios, diversify more cautiously, temper bond exposure during rising volatility, and avoid complacency from recent positive trends.
\item[L16/08] In August 2016, equities and bonds showed steady gains with low volatility, while commodities declined sharply and real estate dipped despite strong prior returns. I overweighted equities and real estate, underweighted commodities and cash. The portfolio outperformed due to equities and bonds. In similar conditions, maintain exposure to stable asset classes and reduce commodities when short-term momentum weakens.
\item[L17/05] In May 2017, equities and real estate showed strong recent gains and moderate volatility, while commodities lagged with negative returns and high volatility. I overweighted equities and real estate, underweighted commodities, and outperformed the equal-weight portfolio. In similar conditions, favor assets with positive momentum and manageable volatility, reducing exposure to weak, volatile sectors like commodities.
\item[L18/04] In April 2018, equities showed recent weakness but strong 6-month gains; bonds were stable with low volatility; commodities mildly positive; real estate mixed signals. I favored equities and bonds over cash, yielding a slightly better return than equal weights. In similar conditions, maintain diversified exposure but overweight assets with solid medium-term momentum and manageable volatility for balanced growth.
\item[L19/10] In October 2019, real estate and equities showed strong recent returns with moderate volatility, while bonds declined. Allocating higher weights to real estate (35\%) and equities (30\%) capitalized on momentum, delivering a portfolio return above the equal-weight benchmark. In similar cases, emphasize assets with strong momentum and moderate risk, but maintain diversification to manage downside risks.
\end{description}}

\textbf{1 April 2022.} Equities had fallen 5.4\% over three months but risen 3.2\% in the last month, and commodities had risen 21.6\% over three months. Three of the lessons (L18/04, L16/08 and L17/05) describe months in which overweighting assets with positive momentum paid off; L15/12 describes one in which it did not and warns against complacency after recent gains. All eight drafts of seed~1, including the one without lessons, put 35\% in equities (1/$N$: 20\%) and 20--30\% in real estate (Table~\ref{tab:case}a), so this tilt comes from the LLM's reading of the market rather than from the lessons. The lessons changed the drafts only slightly, and mostly for the worse: three of the four contributions are negative, and the drafts that used L17/05 earned 0.42~pp less on average than those that did not. The 11 forward scores of seed~1 so far had a mean of 0.24 and passed the trust test ($p=0.045$), so \methodname deployed the draft with the highest total learned value (L18/04 and L15/12). Equities then fell 8.1\% and real estate 7.5\%, and the decision earned $-5.48$\%, against $-3.83$\% for 1/$N$. In seeds 0 and~2 the test did not pass ($p=0.12$ and 0.16), and the anchored action, with a weight of 0.835 on 1/$N$, earned $-4.08$\% and $-4.10$\%.

\textbf{1 September 2022.} Every risky class had lost money over the past one, three and six months, and all eight drafts of seed~1 were defensive, with 10--15\% in equities and 20--40\% in cash. The drafts that used L19/10 held 35--40\% in cash, against 20--30\% for those that did not, and L19/10 had the largest contribution ($+0.73$~pp). In seed~1 the test narrowly failed ($p=0.058$; 16 forward scores, mean 0.17), so \methodname kept the anchored action, which put a weight of 0.875 on 1/$N$ and earned $-5.91$\%, against $-6.06$\% for 1/$N$; the draft it would have deployed (L18/04 and L19/10: 40\% cash, 30\% bonds, 10\% equities) earned $-3.89$\%. In seeds 0 and~2 ($p=0.29$ and 0.42), the anchored action earned $-5.84$\% and $-5.79$\%.

\textbf{What the case shows.} First, contributions are measured on the same date, so the part of the outcome that all drafts share cancels: on both dates every draft had a utility between $-6.1$\% and $-3.8$\%, which outcome credit would charge to every lesson used, yet the contributions separate the lessons by tenths of a pp. Second, the contributions are noisy, so the learned values are small: L18/04 helped on both dates ($+0.12$ and $+0.28$~pp) and L17/05 hurt on both ($-0.42$ and $-0.21$~pp), but after shrinkage all values in seed~1 are below 0.02~pp in absolute value, so the deployed draft is decided by small differences. Third, the trust test decides whether \methodname takes on the drafts' market exposure: in April, when every draft overweighted equities before a fall, acting on the values cost 1.66~pp against 1/$N$; in September, when every draft was defensive, the test narrowly failed and the anchored action gave up the 2.02~pp by which the selected draft beat it. Over all 71 months, acting on the values did not pay on ClassAlloc: on the 93 decisions (all investors and seeds) on which the test passed, the trusted draft earned $0.05$~pp less (s.e.\ $0.07$) than the anchored blend it replaced, and the variant whose test never passes is 0.016~pp per month better (Figure~\ref{fig:ablation}). Keeping part of the anchor even on trusted decisions is a natural extension (Appendix~\ref{app:limitations}).

\section{Algorithms}\label{app:algorithms}

Algorithm~\ref{alg:loop} gives the steps of \methodname on one decision date (\S\ref{sec:loop}). Algorithm~\ref{alg:value} expands its value estimation, and Algorithm~\ref{alg:trust} its trust test and anchored action.

\begin{algorithm}[!tb]
  \caption{\methodname at decision date $t$}\label{alg:loop}
  \small\raggedright
  \begin{algorithmic}[1]
    \Require state $s_t$; retrieved experiences $R_t=(h_1,\dots,h_k)$; design $\design$ (8 subsets of $R_t$); reference $\xref$; contributions of past dates; forward scores; FTRL statistics
    \Statex \textit{Act, before the outcome}
    \For{each $S\in\design$}
      \State draft $x^S\sim\pi(\cdot\mid s_t,S)$ \Comment{8 drafts in parallel}
    \EndFor
    \State fit the content-aware and the per-experience value models on past dates\label{alg:l-fit} \Comment{Eqs.~\eqref{eq:fh}--\eqref{eq:posterior}}
    \State predict $\hat v_j$ for every $h_j\in R_t$ with both models; keep both predictions for scoring
    \State $S^*=\argmax_{S\in\design}\sum_{j\in S}\hat v_j$ for the selected model\label{alg:l-select} \Comment{Eq.~\eqref{eq:trusted}}
    \If{the trust test passes ($n\ge10$ and $p_t<\alpha$, Eq.~\eqref{eq:ttest}) and $\sum_{j\in S^*}\hat v_j>0$}
      \State deploy $x_t=x^{S^*}$ \Comment{trusted}
    \Else
      \State deploy $x_t=q_t\xref_t+(1-q_t)\bar x_t$, with $q_t$ from Eq.~\eqref{eq:ftrl}\label{alg:l-anchor} \Comment{anchored}
    \EndIf
    \Statex \textit{Learn, once the outcome is realized}
    \State $U(S)=u_t(x^S)$ for $S\in\design$; contributions $c_j$ (Eq.~\eqref{eq:contribution}); centred $y_j$ (Eq.~\eqref{eq:centre})
    \State forward score $\rho_t$ of each model (Eq.~\eqref{eq:forward}); update the FTRL statistics of Eq.~\eqref{eq:ftrl}\label{alg:l-score}
  \end{algorithmic}
\end{algorithm}

\setcounter{algorithm}{1}
\begin{algorithm}[!tb]
  \caption{Value estimation (line~\ref{alg:l-fit} of Algorithm~\ref{alg:loop})}\label{alg:value}
  \small\raggedright
  \begin{algorithmic}[1]
    \Require centred contributions $y_{s,j}$ of past dates, each with the experience it belongs to; content features $z_h$; grid $G$ of $(\sigma_\beta^2,\tau^2)$
    \For{each experience $h$ measured on $n_h\ge1$ past dates}
      \State $m_h\gets$ mean of its centred contributions
    \EndFor
    \State $\sigma^2\gets$ pooled within-experience variance of the centred contributions
    \State per-experience model: $\sigma_\beta^2=0$, $\tau^2=\argmax_G$ of the marginal likelihood \Comment{Eq.~\eqref{eq:marglik}}
    \State content-aware model: $(\sigma_\beta^2,\tau^2)=\argmax_G$ of Eq.~\eqref{eq:marglik}
    \For{each model}
      \State $\hat v\gets(\sigma_\beta^2ZZ^{\top}+\tau^2I)K^{-1}m$;\ \ $\hat\beta\gets\sigma_\beta^2Z^{\top}K^{-1}m$ \Comment{Eq.~\eqref{eq:posterior}}
    \EndFor
    \State prediction for a retrieved experience $h$: $\hat v_h$ if $n_h\ge1$, else $z_h^{\top}\hat\beta$ (0 for the per-experience model)
    \Ensure the predictions of both models
  \end{algorithmic}
\end{algorithm}

\begin{algorithm}[!tb]
  \caption{Trust test and anchored action (Algorithm~\ref{alg:loop})}\label{alg:trust}
  \small\raggedright
  \begin{algorithmic}[1]
    \Require forward scores of both models; utilities of the blends of earlier dates
    \State selected model $\gets$ per-experience model until both models have $\ge10$ forward scores, then the model with the higher mean forward score
    \State $n,\bar\rho,s_\rho\gets$ number, mean and sd of the selected model's forward scores
    \State $p\gets\Pr\big(T_{n-1}\ge\bar\rho/(s_\rho/\sqrt{n})\big)$ \Comment{Eq.~\eqref{eq:ttest}}
    \State $S^*\gets\argmax_{S\in\design}\sum_{j\in S}\hat v_j$, ties towards fewer experiences \Comment{Eq.~\eqref{eq:trusted}}
    \State trusted $\gets(n\ge10)\wedge(p<\alpha)\wedge\big(\sum_{j\in S^*}\hat v_j>0\big)$
    \If{not trusted}
      \State $q_t\gets\argmax_{q}\ \mean_{s<t}u_s\big(q\,\xref_s+(1-q)\bar x_s\big)-\lambda_n\KL(q\,\|\,\alpha_0)$ over $q\in\{0,0.005,\dots,1\}$ \Comment{Eq.~\eqref{eq:ftrl}}
    \EndIf
    \State after the outcome of date $t$: for each model, $\rho_t\gets\corr(\hat v_{t,\cdot},y_{t,\cdot})$ (Eq.~\eqref{eq:forward}), appended to its forward scores; store $u_t$ of every blend
  \end{algorithmic}
\end{algorithm}

\section{LLM Prompts}\label{app:prompts}

This appendix gives the prompts used on the three real benchmarks, copied from the logged requests. Text in angle brackets is filled in at run time, and [...] marks lines omitted here. On each benchmark, all LLM-based methods use the same prompts and differ only in the memory block that follows the market context (Table~\ref{tab:prompt-blocks}); the other LLMs of \S\ref{sec:robust} receive the same prompts unchanged. The drafts of all 16 subsets of the four retrieved lessons are logged on every date and seed, so the methods in the second row of Table~\ref{tab:prompt-blocks} differ only in which draft they deploy or how they combine the drafts. On InvestorBench and ClassAlloc, a draft is one chat completion with a system and a user message, and an invalid reply is answered once with a retry message; on PortBench, a draft is one run of the benchmark's three-stage agent, without retries.

\begin{table}[h]
  \caption{Memory block of each LLM-based method in the decision prompt on InvestorBench and ClassAlloc.}
  \label{tab:prompt-blocks}
  \footnotesize
  \setlength{\tabcolsep}{3pt}
  \begin{tabularx}{\columnwidth}{@{}>{\raggedright\arraybackslash}p{0.36\columnwidth}>{\raggedright\arraybackslash}X@{}}
    \toprule
    Method & Memory block \\
    \midrule
    Memory-free, Self-consistency & none \\
    \methodname, Similarity retrieval, Counterfactual selection, Uplift credit, Draft averaging & experiences block with a subset of the four retrieved lessons \\
    FinMem & experiences block with four lessons ranked by similarity, recency and importance \\
    MemRL & experiences block with four of the ten most similar lessons, ranked by learned value ($\varepsilon$-greedy) \\
    Reflexion & reflections block with its three most recent reflections \\
    ExpeL & insights block with at most eight insights distilled from all warm-up lessons \\
    \bottomrule
  \end{tabularx}
\end{table}

\subsection{InvestorBench}\label{app:prompts-ib}

\noindent\textbf{System message} (all calls).
\begin{promptlisting}
You are a portfolio manager. You allocate a long-only portfolio across four U.S. stocks (HON, JNJ, MSFT, UVV) and cash. Each trading day you set target weights that are held until the next trading day's close.
\end{promptlisting}
\noindent\textbf{Decision prompt} (user message), here for 25 February 2021 and the draft with all four retrieved lessons.
\begin{promptlisting}
Date: 2021-02-25

Market data (returns in %; volatility = standard deviation of daily returns over 20 days, in %):
HON: close 190.74; 1-day -2.53; 5-day +1.07; 20-day +3.17; volatility 1.48
JNJ: close 148.73; 1-day +0.10; 5-day -1.19; 20-day -2.45; volatility 1.14
MSFT: close 222.30; 1-day -2.37; 5-day -6.07; 20-day -1.45; volatility 1.53
UVV: close 41.36; 1-day -1.53; 5-day -1.16; 20-day +7.24; volatility 2.02

News today:
- JNJ: The global Cold Pain Therapy Market is projected to grow from USD 1.6 billion in 2020 to USD 2.0 billion by 2025, at a CAGR of 4.4%, driven by the increasing prevalence of arthritis and musculoskeletal disorders and growing awareness of col
- JNJ: Lucid has announced the integration of its virtual whiteboard, Lucidspark, with Microsoft Teams, enhancing real-time collaboration and innovation for Teams users. This integration allows seamless movement between Lucidchart and Lucidspark d
[... 7 more headlines: at most three per stock, each cut at 240 characters]

Experiences from your past decisions (use them only where relevant):
[1] (2020-09-14) In a mixed market with positive news for JNJ and MSFT but recent negative 5-day returns for MSFT, I overweighted JNJ and MSFT, underweighted UVV and HON, and held some cash. The portfolio outperformed slightly. In similar cases, balance recent momentum with positive news, maintain diversification, and keep modest cash to manage uncertainty.
[2] (2020-08-18) In a mixed market with modest gains and losses, overweighting MSFT and JNJ aligned with positive trends but UVV(*@\textquoteright@*)s rebound was underestimated. Diversifying more evenly and maintaining a slightly higher cash position could reduce downside risk. In similar conditions, balance growth potential with volatility by avoiding concentration in fewer stocks despite recent strong performance.
[3] (2020-07-17) In a mixed market with strong news for JNJ and MSFT but recent declines in MSFT and UVV, I overweighted HON and JNJ for stability and growth potential. Despite a dip in HON, MSFT(*@\textquoteright@*)s rebound boosted returns. In similar cases, balance stable leaders with growth stocks showing recovery signs, maintaining some cash to manage volatility.
[4] (2020-07-16) In a mixed market with strong recent gains in HON and JNJ but weakness in MSFT and UVV, overweighting HON and JNJ while underweighting MSFT and UVV led to outperformance. For similar conditions, favor stocks with positive momentum and stable news, maintain some cash for flexibility, and cautiously reduce exposure to recent laggards to improve returns.

Decide the target weights for the next trading day. Respond with a JSON object only, for example {"HON": 0.2, "JNJ": 0.2, "MSFT": 0.2, "UVV": 0.2, "CASH": 0.2, "reason": "one short sentence"}. Weights must be non-negative and sum to 1.
\end{promptlisting}
\noindent\textbf{Reply} of gpt-4.1-mini (seed~0), as parsed.
\begin{promptlisting}
{"HON": 0.25, "JNJ": 0.35, "MSFT": 0.2, "UVV": 0.1, "CASH": 0.1, "reason": "Overweight JNJ for positive news and stability; moderate HON for steady gains; reduce MSFT and UVV due to recent declines and higher volatility; keep cash for risk management."}
\end{promptlisting}
\noindent\textbf{Retry message}, sent once after an invalid reply.
\begin{promptlisting}
Your reply was not a valid JSON weight object. Reply again with the JSON object only, weights non-negative and summing to 1.
\end{promptlisting}
\noindent\textbf{Lesson prompt.} After each warm-up decision, the agent sees the outcome and writes a lesson. The prompt for 2 July 2020:
\begin{promptlisting}
<market context of the warm-up day, in the format of the decision prompt>

Your decision: HON 15%, JNJ 30%, MSFT 40%, UVV 5%, CASH 10%
Next trading day returns (%): HON +1.52, JNJ +1.43, MSFT +2.15, UVV +1.45
Your portfolio return: +1.59%; equal-weight portfolio (four stocks and cash): +1.31%.

Write one reusable lesson for future allocation decisions in at most 60 words: describe the market situation, what you did, what happened, and what you would do in a similar situation. Plain text only.
\end{promptlisting}
\noindent The written lesson, stored with its date as experience E000:
\begin{promptlisting}
(2020-07-02) In a mixed market with strong MSFT momentum and positive JNJ news, overweighting MSFT and JNJ while maintaining modest exposure to HON and UVV outperformed an equal-weight approach. Future allocations should emphasize stocks with positive catalysts and momentum, balancing risk with cash to capture upside while managing volatility.
\end{promptlisting}
\noindent\textbf{Reflection prompt} (Reflexion): the lesson prompt with its last paragraph replaced by
\begin{promptlisting}
Reflect on this decision in at most 60 words: what worked, what did not, and what you will do differently next time. Plain text only.
\end{promptlisting}
\noindent\textbf{Insight prompt} (ExpeL), run once on all 63 warm-up lessons:
\begin{promptlisting}
Below are lessons you wrote after past daily decisions, each with the outcome of that decision.

- [2020-07-02; portfolio +1.59% vs equal weight +1.31%] In a mixed market with strong MSFT momentum and positive JNJ news, overweighting MSFT and JNJ while maintaining modest exposure to HON and UVV outperformed an equal-weight approach. Future allocations should emphasize stocks with positive catalysts and momentum, balancing risk with cash to capture upside while managing volatility.
[... 62 more lessons, one per warm-up day]

Compare the successful and unsuccessful decisions and extract at most 8 general insights for future daily allocation across HON, JNJ, MSFT, UVV and cash. Return a numbered list; each insight at most 30 words.
\end{promptlisting}
\noindent\textbf{Insights and reflections blocks}, which take the place of the experiences block for ExpeL and Reflexion:
\begin{promptlisting}
Insights distilled from your past decisions:
1. Balance growth momentum with recent price trends and volatility to avoid overexposure to short-term weakness or reversals in high-volatility stocks like MSFT and UVV.
2. Emphasize stocks with strong fundamentals and positive catalysts (e.g., HON, JNJ, MSFT) while maintaining modest cash to manage downside risk and market uncertainty.
3. Maintain diversification across all stocks to capture unexpected rebounds, especially in volatile or uncertain markets, avoiding concentrated bets in recent winners.
[... at most 8 insights]

Your reflections on your most recent decisions (oldest first):
- <reflection on the third most recent decision>
- <reflection on the second most recent decision>
- <reflection on the most recent decision>
\end{promptlisting}

\subsection{ClassAlloc}\label{app:prompts-cb}

\noindent\textbf{System message} (all calls).
\begin{promptlisting}
You are a portfolio manager. You allocate a long-only portfolio across four asset classes (equities, bonds, commodities, real estate) and cash. At the start of each month you set target weights that are held for the next 20 trading days.
\end{promptlisting}
\noindent\textbf{Decision prompt}, here for 1 April 2022 (Appendix~\ref{app:case}) and the draft with all four retrieved lessons.
\begin{promptlisting}
Date: 2022-04-01

Market data (each asset class and cash is an equal-weighted index of exchange-traded assets; returns in %; volatility = standard deviation of daily returns over the last 20 trading days, in %):
equities: 1-month +3.20; 3-month -5.35; 6-month +5.93; volatility 1.62
bonds: 1-month -2.06; 3-month -5.59; 6-month -5.23; volatility 0.45
commodities: 1-month -1.35; 3-month +21.55; 6-month +22.05; volatility 2.03
real estate: 1-month +2.17; 3-month -8.63; 6-month +2.81; volatility 1.38
cash: 1-month -0.05; 3-month -0.12; 6-month -0.16; volatility 0.01

Experiences from your past decisions (use them only where relevant):
[1] (2018-04-02) In April 2018, equities showed recent weakness but strong 6-month gains; bonds were stable with low volatility; commodities mildly positive; real estate mixed signals. I favored equities and bonds over cash, yielding a slightly better return than equal weights. In similar conditions, maintain diversified exposure but overweight assets with solid medium-term momentum and manageable volatility for balanced growth.
[2]-[4] <the lessons of August 2016, May 2017 and December 2015; Appendix (*@\ref{app:case}@*)>

Decide the target weights for the next 20 trading days. Respond with a JSON object only, for example {"equities": 0.2, "bonds": 0.2, "commodities": 0.2, "real_estate": 0.2, "cash": 0.2, "reason": "one short sentence"}. Weights must be non-negative and sum to 1.
\end{promptlisting}
\noindent The retry message, the reflection prompt and the memory blocks are those of InvestorBench. In the lesson and reflection prompts, the outcome lines become
\begin{promptlisting}
Your decision: equities <x>%, bonds <x>%, commodities <x>%, real estate <x>%, cash <x>%
Returns over the next 20 trading days (%): equities <+x.xx>, bonds <+x.xx>, commodities <+x.xx>, real estate <+x.xx>, cash <+x.xx>
Your portfolio return: <+x.xx>%; equal-weight portfolio (the five asset classes): <+x.xx>%.
\end{promptlisting}
\noindent and the insight prompt (ExpeL) becomes
\begin{promptlisting}
Below are lessons you wrote after past monthly decisions, each with the outcome of that decision.

- [<date>; portfolio <+x.xx>% vs equal weight <+x.xx>%] <lesson>
[... one line per warm-up month, 58 in all]

Compare the successful and unsuccessful decisions and extract at most 8 general insights for future monthly allocation across equities, bonds, commodities, real estate and cash. Return a numbered list; each insight at most 30 words.
\end{promptlisting}
\noindent\textbf{Memorization probe} (Appendix~\ref{app:lookahead}), here for 1 April 2022; the system message is ``You answer questions about financial markets.''
\begin{promptlisting}
Date: 2022-04-01

Consider five asset classes, each an equal-weighted index of exchange-traded assets: equities, bonds, commodities, real estate and cash. From what you know about the markets, give the return in % of each asset class over the 20 trading days starting on this date. If you are not sure, give your best estimate. Respond with a JSON object only, for example {"equities": 1.5, "bonds": -0.3, "commodities": 2.0, "real_estate": 0.4, "cash": 0.1, "reason": "one short sentence"}.
\end{promptlisting}

\subsection{PortBench}\label{app:prompts-pb}

A draft is one run of the benchmark's native agent in three stages: market interpretation, signal generation and weight optimization. Each stage is one chat completion that returns a JSON object (JSON mode, at most 8,192 output tokens) and, from stage~2 on, receives the output of the previous stage; the memory enters all three stages as the benchmark's experience block, which holds the records selected by the method. The prompts below are those of PortBench-Raw for the balanced investor on 1 April 2022, for the draft with all four retrieved records; PortBench-Full extends the market-data table of stage~1 from eight assets to all eligible assets.

\noindent\textbf{System message} (all stages).
\begin{promptlisting}
You are the investment agent in a simulated research benchmark. Follow the native stage instruction and return its JSON object; historical experiences are observations, not guarantees.
\end{promptlisting}
\noindent\textbf{Stage 1: market interpretation.}
\begin{promptlisting}
You are a portfolio manager analyzing market conditions on 2022-04-01.

MARKET DATA (trailing 60 trading days):
  QQQ: price=352.83, trailing_return=-5.72%, ann_vol=31.03%
  IWM: price=197.08, trailing_return=-4.32%, ann_vol=26.82%
  VTI: price=215.83, trailing_return=-3.35%, ann_vol=22.60%
  IVV: price=429.40, trailing_return=-2.98%, ann_vol=21.48%
  EEM: price=41.63, trailing_return=-4.83%, ann_vol=28.28%
  EFA: price=65.73, trailing_return=-5.89%, ann_vol=22.76%
  VEA: price=42.58, trailing_return=-5.04%, ann_vol=21.35%
  FXI: price=30.07, trailing_return=-5.79%, ann_vol=58.06%
Macro: fed_funds_rate=0.330, cpi_yoy=0.000, unemployment=3.900, gdp_growth_qoq=0.000, t10y2y_spread=0.040

MACRO INDICATORS:
  Fed Funds Rate (%): 0.33
  Unemployment (%): 3.90
  10Y-2Y spread (pp): 0.04
  [... 12 more indicators]

[... pairwise, intra-class and inter-class return correlations]

Current market regime context: unknown

RECENT NEWS / FILINGS:
[INVESTOR PROFILE] This investor is BALANCED: allocate at most 65% to equities and cryptocurrency combined; allocate at least 20% to bonds and cash; the portfolio may accept up to a 20% maximum drawdown.

[Kaggle 2022-03-31] ["Apple now allows Netflix, Spotify and other 'reader' apps to link to their sites for payment It's part of the tech giant's settlement with the Japan Fair Trade Commission.", [... further headlines of the previous day]

TASK: Interpret the market data and provide structured asset views.

Return a sparse set of material asset views. Any visible asset omitted from
"asset_views" is deterministically interpreted as neutral (0.0). Only use
visible asset identifiers and do not invent assets. You may include an
explicit neutral 0.0 view, but do not need to list every neutral asset.

For each reported asset, assign a sentiment score in [-1.0, +1.0]:
  +1.0 = strongly bullish (expect strong outperformance)
   0.0 = neutral
  -1.0 = strongly bearish (expect significant underperformance)

Identify the overall market regime: one of "bull", "bear", "sideways", "high-volatility".

JSON SCHEMA:
{
  "asset_views": { "ASSET": <float in [-1, 1]> },
  "detected_regime": "<bull|bear|sideways|high-volatility>",
  "confidence": <float in [0, 1]>,
  "macro_summary": "<one sentence>"
}

OUTPUT FORMAT (*@\textemdash@*) STRICT:
  1. Reply with EXACTLY ONE JSON object and nothing else.
  2. The very first character of your reply MUST be `{` and the very last
     character MUST be `}`. No leading whitespace, no trailing newline text.
  3. Do NOT wrap the object in markdown code fences (no ```json, no ```).
  4. Do NOT add any explanation, preamble, or commentary before/after.
  5. Use double quotes for all keys and string values; numbers must be raw
     JSON numbers (no quotes, no `%`, no commas).
  6. Every key listed in the schema below MUST appear exactly once.

FROZEN HISTORICAL EXPERIENCE (observations, not proof of causal advantage):
{"text":"[{\"id\":\"ec8700d3d7c4...\",\"source_ids\":[...],\"text\":\"<the record below, JSON-encoded>\"}, <3 more records>]","unique_source_ids":[]}
\end{promptlisting}
\noindent\textbf{One record of the experience block}, decoded. Each record describes an earlier decision of the benchmark's agent (2015--2019): its allocation by asset class, a readout of its execution (net return $J$, fee, turnover, concentration HHI, largest weight and fraction of assets held), public market features at the time, and the realized 20-day net return.
\begin{promptlisting}
{
 "actual_target_weights_by_asset_class": {"bonds": 0.111, "cash": 0.011, "commodities": 0.1184, "cryptocurrency": 0.0074, "equities": 0.6856, "real_estate": 0.0666},
 "benchmark": "PortBench-Dynamic-99asset-native-shared-memory-raw-price-adaptation",
 "decision_date": "2016-05-02",
 "execution_readout_J_fee_turnover_HHI_maxweight_nonzero_fraction": [0.010039484803091359, 0.0010229302, 0.682, 0.029007079999999998, 0.0735, 0.9883720930232558],
 "experience_id": "portbench/ec8700d3d7c4...:portbench:2016-05-02:r0",
 "feedback_scope": "actual20sessiontradeoutcome; unexecutedalternativesunknown; notcausalsuperiority",
 "public_state_features": [0.0, 0.86, 0.00153055, 0.01327625, 0.61627907, 0.19767442],
 "realized_net_interval_return": 0.010039484803091359,
 "settled_date": "2016-05-31"
}
\end{promptlisting}
\noindent\textbf{Stage 2: signal generation.}
\begin{promptlisting}
You are a portfolio manager on 2022-04-01.

Stage 1 market interpretation produced these asset views (scale: -1=bearish, +1=bullish):
  QQQ: view=-0.400
  IWM: view=+0.000
  VTI: view=+0.000
  [... the other assets]

Detected market regime: high-volatility
Macro summary: Market shows elevated volatility with flat CPI, low Fed funds rate, narrow yield curve, and moderate unemployment, indicating uncertain economic conditions and cautious sentiment.

Recent news / filings:
[investor profile and news, as in stage 1]

TASK: Convert each asset view into an actionable trading signal.

Rules:
  - view >  0.15: consider "buy"
  - view < -0.15: consider "sell"
  - otherwise:    consider "hold"

Use your judgement to refine signals based on regime and macro context.
Signal strength should reflect conviction (0.0 = low, 1.0 = high).
Return only non-hold signals. Omitted visible assets are interpreted as hold with strength 0.5.
The signals and strengths objects must contain exactly the same selected assets.

JSON SCHEMA:
{
  "signals": {"SELECTED_ASSET": "<buy|sell>", ...},
  "strengths": {"SELECTED_ASSET": <float in [0, 1]>, ...},
  "reasoning": "<one sentence>"
}

OUTPUT FORMAT (*@\textemdash@*) STRICT:
[the six rules of stage 1]

[experience block, as in stage 1]
\end{promptlisting}
\noindent\textbf{Stage 3: weight optimization.}
\begin{promptlisting}
You are a portfolio manager on 2022-04-01.

Stage 2 signals:
  QQQ: signal=sell, strength=0.60
  IWM: signal=hold, strength=0.50
  VTI: signal=hold, strength=0.50
  [... the other assets]

Current portfolio weights: QQQ=0.0103, IWM=0.0103, VTI=0.0103, [...] (equal weights)
Portfolio NAV: $1,000,000
Market regime: unknown

[return correlations, as in stage 1]

TASK: Select a sparse portfolio allocation based on the signals above.

Constraints:
  - allocation_scores must be finite, non-negative decimal numbers, with at least one strictly positive score
  - Do not normalize allocation_scores; the deterministic environment normalizes their relative magnitudes into target weights
  - Omitted visible assets receive a target weight of 0.0; a listed 0.0 explicitly excludes that asset
  - Prefer a concise allocation, while retaining any material existing holdings
  - "sell" signals should receive reduced weight (ideally 0.0)
  - "buy" signals should receive increased weight
  - Minimize unnecessary turnover from current weights

JSON SCHEMA:
{
  "allocation_scores": {"SELECTED_ASSET": <non-negative decimal>, ...},
  "expected_return": <annualized decimal, e.g. 0.08>,
  "expected_vol": <annualized decimal, e.g. 0.12>,
  "sharpe_estimate": <decimal>
}

OUTPUT FORMAT (*@\textemdash@*) STRICT:
[the six rules of stage 1]

[experience block, as in stage 1]
\end{promptlisting}
\noindent\textbf{Replies} of gpt-4.1-mini in this draft (seed~1). The environment turns the stage-3 scores into target weights, which are then projected to the investor's mandate (Appendix~\ref{app:impl}). Here, stage~3 gives the same score to the 93 assets without a sell signal and none to the four with one.
\begin{promptlisting}
Stage 1: {"asset_views": {"FXI": -0.7, "QQQ": -0.4, "EFA": -0.5, "EEM": -0.4}, "detected_regime": "high-volatility", "confidence": 0.8, "macro_summary": "Market shows elevated volatility with flat CPI, low Fed funds rate, narrow yield curve, and moderate unemployment, indicating uncertain economic conditions and cautious sentiment."}

Stage 2: {"signals": {"QQQ": "sell", "EEM": "sell", "EFA": "sell", "FXI": "sell"}, "strengths": {"QQQ": 0.6, "EEM": 0.6, "EFA": 0.7, "FXI": 0.9}, "reasoning": "In a high-volatility regime with cautious sentiment and bearish views on QQQ, EEM, EFA, and FXI, moderate to high conviction sell signals are warranted to reduce risk exposure in these assets."}

Stage 3: {"allocation_scores": {"IWM": 0.5, "VTI": 0.5, "IVV": 0.5, "VEA": 0.5, ... (89 more assets)}, "expected_return": 0.07, "expected_vol": 0.14, "sharpe_estimate": 0.5}
\end{promptlisting}

\section{Limitations}\label{app:limitations}

\begin{itemize}
  \item \emph{Gains from memory on real markets remain to be shown.} On PortBench and InvestorBench, \methodname's advantage over the experience-learning agents comes from abstaining: its values passed the trust test on fewer than 1\% of the test decisions, and with the same anchor the baselines do as well (Appendix~\ref{app:ablation}). On ClassAlloc they passed on 15\%, and the trusted drafts did not beat the anchored blend they replaced. These results show that \methodname avoids the losses of acting on untested memory, not that memory improves real decisions; showing the latter may need more decisions, more informative experiences or tasks in which experiences matter more. When its values are trusted, \methodname deploys the selected draft in full and so takes on its market exposure (Appendix~\ref{app:case}); keeping part of the anchor even then is a natural extension.
  \item \emph{Pretraining exposure (look-ahead bias).} The knowledge cutoff of gpt-4.1-mini (June 2024) is later than the whole InvestorBench test period and than 15 of the 20 PortBench test months, so the LLM may have seen these market outcomes during pretraining; the same holds for 50 of the 71 ClassAlloc test months. On ClassAlloc, the memory-free LLM beats 1/$N$ by 0.08~pp per month before the cutoff and trails it by 0.08~pp after it (Appendix~\ref{app:classalloc}), although 21 months are too few for this difference to be significant. Every LLM recalls dated outcomes from before its cutoff, and hiding the dates on ClassAlloc did not remove the memory-free draft's advantage before the cutoff (Appendix~\ref{app:lookahead}), but the LLM may still recognize a period from the returns it is shown; we did not hide the dates on PortBench and InvestorBench. This exposure affects all LLM-based methods alike but may change how much any LLM agent gains from memory; PlantedMem, whose drafts do not depend on the date, is free of it.
  \item \emph{The anchor also holds back good drafts.} When the LLM's drafts beat the reference, as on ClassAlloc before the knowledge cutoff, the anchored action keeps \methodname close to 1/$N$ and gives up most of their advantage (Appendices~\ref{app:classalloc} and~\ref{app:cb-llms}). Its online blend moves towards better drafts, but slowly under a prior of 0.9 on the reference (Appendix~\ref{app:regime}); a lower prior adapts faster but protects less when the reference is strong.
  \item \emph{Error rate of the trust test.} The trust test is applied on every date to dependent scores of a selected learner (Appendix~\ref{app:trusttest}), so its false-pass rate has no formal bound; we measure it instead (2--3\% of PlantedMem decisions without signal). A sequential test with time-uniform error control on block or permutation scores would bound it.
  \item \emph{Cost.} \methodname needs eight drafts per decision instead of one. The drafts run in parallel, so latency does not grow, but the LLM cost does; adaptive designs that spend more drafts only on dates where contributions are uncertain could reduce it.
  \item \emph{Benchmark protocols.} PortBench scores each monthly decision as a separate episode that starts from the equal-weight portfolio, so its costs are charged on the distance from that portfolio rather than on actual turnover, and its experience pool is fixed before the first decision date, so learning happens in the credit rather than in the content of memory. Our InvestorBench task is a multi-asset adaptation of the benchmark's single-stock task, with four stocks and cash. ClassAlloc's class indices are equal-weighted and rebalanced daily, and its cryptocurrency class is left out. PlantedMem is semi-synthetic: its experiences act on the drafts through a simple tilt model (Appendix~\ref{app:impl}), which is cleaner than the way real lessons act on an LLM, and each period's outcome is known before the next decision.
\end{itemize}

\end{document}